%% file: main.tex
\documentclass{article}

\usepackage[nonatbib, main, final]{neurips_2026}

\usepackage[utf8]{inputenc} 
\usepackage[T1]{fontenc}    
\usepackage{amsfonts}       
\usepackage{nicefrac}       
\usepackage{xcolor}         
\usepackage[sort&compress, numbers]{natbib}

\usepackage{microtype}
\usepackage{graphicx}
\usepackage{subcaption}
\usepackage{booktabs} 
\usepackage{layouts}

\usepackage{hyperref}
\usepackage{xurl}
\usepackage{placeins}

\usepackage{amsmath}
\usepackage{amssymb}
\usepackage{mathtools}
\usepackage{amsthm}
\usepackage{multirow}

\usepackage{wrapfig}

\usepackage{titletoc}

\usepackage[capitalize,noabbrev]{cleveref}
\usepackage{tikz}
\usetikzlibrary{arrows.meta,positioning,calc}

\usepackage[normalem]{ulem}

\theoremstyle{plain}
\newtheorem{theorem}{Theorem}[section]

\newtheorem{lemma}[theorem]{Lemma}

\theoremstyle{definition}

\theoremstyle{remark}

\renewcommand{\v}[1]{\ensuremath{\mathbf{#1}}}
\newcommand{\tv}[1]{\ensuremath{\mathbf{\tilde{#1}}}}

\newcommand{\softmax}{\ensuremath{\operatorname{Softmax}}}

\DeclareMathOperator*{\argmax}{arg\,max}

\newcommand{\trim}[1]{}

\usepackage[most]{tcolorbox}
\tcbuselibrary{breakable,listings}

\newtcblisting{promptbox}[1]{%
  breakable,
  listing only,
  colback=black!1,
  colframe=black!40,
  boxrule=0.5pt,
  arc=2mm,
  left=2mm,right=2mm,top=1mm,bottom=1mm,
  title=#1,
  listing options={
    basicstyle=\ttfamily\small,
    breaklines=true,
    breakatwhitespace=false
  }
}

\usepackage[textsize=tiny]{todonotes}

\title{Finding Interpretable Prompt-Specific Circuits in Language Models}

\author{%
  Gabriel Franco \\
  Department of Computer Science\\
  Boston University\\
  \texttt{gvfranco@bu.edu} \\
  \And
  Lucas M. Tassis \\
  Department of Computer Science\\
  Boston University\\
  \texttt{ltassis@bu.edu} \\
  \And
  Azalea Rohr \\
  Faculty of Computing \& Data Sciences\\
  Boston University\\
  \texttt{arohr@bu.edu} \\
  \And
  Mark Crovella \\
  Department of Computer Science and\\
  Faculty of Computing \& Data Sciences\\
  Boston University\\
  \texttt{crovella@bu.edu} \\
}

\begin{document}

\maketitle

\begin{abstract}
Understanding the internal circuits that language models use to solve tasks remains a central challenge in mechanistic interpretability. A crucial part of finding circuits is understanding why each attention head attends where it does. To this end, we introduce \textbf{ACC++}, an improved circuit-tracing method based on the principle of \emph{attention-causal communication} (ACC) \cite{franco2025pinpointing}, which identifies \emph{signals}, i.e., contents of low dimensional subspaces that cause attention on a token pair. ACC++ extracts circuits from a \emph{single forward pass}, without replacement models or patching. Circuits identified by ACC++ consist of components that are causal for the model's attention decisions, together with the low-dimensional signals used to communicate between them. Here, we first detail the conceptual advances that ACC++ makes over previous work. We then show that across multiple models, a substantial portion of ACC++ signals are \emph{interpretable}: many signals admit a short natural-language description. We next present a number of new insights into model behavior obtained via ACC++. First, we use ACC++'s interpretable circuits to characterize the sensitivity of indirect object identification (IOI) circuits to prompt structure. We find that prompt-specific circuits form well-defined clusters, and across clusters, heads receive systematically different signals corresponding to distinct mechanisms for identifying the IO name. Next, in multilingual IOI, ACC++ circuits show that while model \emph{components} are reused across languages, \emph{signals} are often language-specific. In a four-language IOI case study, cross-language circuit distances are consistent with linguistic relatedness. Together, these results show that ACC++ can shed light on a broad spectrum of model behaviors.
\end{abstract}

\input{sections/intro}
\input{sections/improv_acc}

\input{sections/example_applications}

\input{sections/related_work}
\input{sections/conclusions}

\begin{ack}
\input{sections/acks}
\end{ack}

\bibliography{references}
\bibliographystyle{unsrt}



\clearpage
\appendix
\input{sections/appendix_index}
\input{sections/appendix}
\stopcontents[appendices]


\newpage
\input{checklist.tex}

\end{document}

%% file: sections/intro.tex
\section{Introduction} \label{sec:intro}

Large language models (LLMs) have become widely deployed,
yet their internal computations remain difficult to understand. This opacity is increasingly a practical problem: when models fail, behave inconsistently across prompt wordings, or exhibit unexpected generalization, we often lack a mechanistic account of \emph{why}. Mechanistic interpretability aims to fill this gap by explaining model behavior in terms of human-understandable computations implemented by collections of model components --- often called \emph{circuits} \cite{Elhage21}. In this view, a circuit is not merely a correlated set of neurons or features, but a set of causal pathways: specific components write information into the residual stream, and downstream components selectively read and transform it.

To trace such circuits, we introduce \textbf{ACC++}\footnote{Code is available at
\url{https://github.com/gaabrielfranco/finding-interpretable-circuits}.}, a circuit-tracing method that builds on \emph{attention-causal communication} (ACC) \cite{franco2025pinpointing}. ACC leverages the observation that attention heads communicate through low-dimensional subspaces that can be identified via analysis of the head's query--key (QK) matrix \cite{FrancoEtAl:ICML26,NEURIPS2024_6216515a,merullo2024talkingheadsunderstandinginterlayer,franco2025pinpointing, Ahmad:NeurIPS2025}.  
For a given head, ACC uses this principle to trace which upstream components write into those low-dimensional subspaces, called \emph{signals}, that are causal for the head's attention decision.

Here, we make two sets of contributions.  Our \textbf{first set of contributions} consists of a number of conceptual advances over prior work \cite{franco2025pinpointing}.  These have the effect of \emph{dramatically reducing the noise} present in ACC traces.  ACC++ traces contain \emph{much lower-dimensional signals}, which supports natural language interpretability, as we show below.  Further, they contain \emph{50\% to 90\% fewer edges}. We show that many components included by ACC can be omitted without violating the ACC++ attention-level counterfactual criterion, suggesting that they are spurious for the traced attention decisions. We describe these advances and the benefits they provide, including natural-language interpretability, in \S \ref{sec:improv-acc}.

A key benefit of the attention-causal communication approach is that it can be used to trace circuits on a \emph{per-prompt} basis. The availability of interpretable prompt-specific circuits opens additional avenues for understanding model behavior.

Accordingly, our \textbf{second set of contributions} is to surface previously-unobserved model behaviors that are visible only when using per-prompt tracing.  We use the Indirect Object Identification (IOI) task as our setting, studying its ACC++ circuits across GPT-2, Pythia 160M and Gemma-2 2B. We first show that \emph{distinctly different circuits are associated with different IOI prompt formats}.  For example, GPT-2 circuits are highly dependent on the order of the proper nouns;  in contrast, Pythia circuits are instead sensitive to lower-level wording changes in prompts.  We explain this result by examining and interpreting the signals involved in each trace type, showing that qualitatively different signals are used in each case.  We next study the IOI task in a multi-lingual setting, and show that while common components are used in each circuit, that \emph{signals used are language-specific}, resulting in consistent differences between per-language traces that correlate with linguistic similarity of those languages.    We describe these contributions in \S \ref{sec:uses}.

Finally, we note that ACC++ uses only a single forward model pass and does not require replacement models,  patching, or counterfactual inputs.  As a result, ACC++ makes per-prompt circuits viable units of analysis at scale, and supports the study of how model circuits vary across prompts, templates, and languages. 

%% file: sections/improv_acc.tex
\section{Refining attention causal communication} \label{sec:improv-acc}

As mentioned above, ACC++ is an improved method for circuit tracing based on the principle of \emph{attention-causal communication} (ACC) as introduced in \cite{franco2025pinpointing}. ACC++ makes a number of conceptual advances over previous methods that together enable it to construct accurate and interpretable per-prompt circuits with no tuning involved beyond a single attention threshold. \trim{Circuits found with ACC++ are useful for a broad spectrum of interpretability tasks, including those described in \S \ref{sec:uses}.}

\begin{figure}
    \centering
    \includegraphics[width=\linewidth]{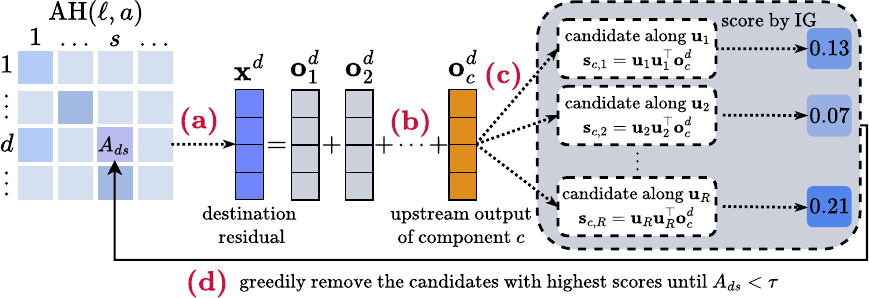}
    \caption{\textbf{ACC++ method (destination side; source side is analogous).} Given an attention head $(\ell,a)$ attending to a token pair $(d,s)$, ACC++ identifies which upstream contributions drive the attention in four steps. \textbf{(a)~Select head and token pair}, \textbf{(b)~Decompose the destination residual},
    \textbf{(c)~Construct candidate signals from singular directions}, and
    \textbf{(d)~Compute attribution by Integrated Gradients and greedily remove.}}
    \label{fig:accpp-method}
    \vspace*{-2em}
\end{figure}

\textbf{Background.} For exposition, we fix an attention head $(\ell,a)$ attending to a destination-source\footnote{We call the query token the \emph{destination} token and the key token the \emph{source} token.} token pair $(d,s)$. Let $\v{x}^j$ denote the residual stream at token position $j$ on input to head $(\ell,a)$. The pre-softmax attention score on the pair $(d,s)$ can then be written as $A'_{ds} = \v{x}^{d\top}\Omega \v{x}^s$, where $\Omega = W_Q W_K^\top$ is the head's \emph{QK matrix}. We write $A'_d$ for the row of pre-softmax scores at destination $d$ and $A_d$ for the corresponding post-softmax attention weights, so that $A_{ds}$ is the attention placed on source token $s$. By \emph{upstream components} we mean token embeddings, attention heads and feedforward (FFN) modules in layers before $\ell$. Moreover, the residual stream at any layer and token position decomposes additively into individual upstream component outputs, $\v{x}^d = \sum_i \v{o}_i^d$ (Figure~\ref{fig:accpp-method}b).

Prior work has shown that the singular vectors of QK matrices often align with meaningful features in residual representations \cite{FrancoEtAl:ICML26}, exposing the communication channels used by attention heads \cite{NEURIPS2024_6216515a,merullo2024talkingheadsunderstandinginterlayer,franco2025pinpointing, Ahmad:NeurIPS2025}. If $\Omega = \sum_{k=1}^R \sigma_k \v{u}_k \v{v}_k^\top$ is its singular value decomposition (SVD), then $A'_{ds} = \sum_{k=1}^R (\v{x}^{d\top}\v{u}_k)\sigma_k(\v{v}_k^\top \v{x}^s)$. Intuitively, the SVD yields paired query-side and key-side directions, and each direction defines a \emph{candidate signal} through which upstream components can influence attention on a token pair. 

When an attention head attends to a particular token pair, that choice is determined by features written into the token representations by upstream components \cite{Elhage21}. The attention-causal communication (ACC) problem of \cite{franco2025pinpointing} formalized this goal for a fixed destination-source token pair. To pinpoint causality, methods in \cite{franco2025pinpointing} seek a small subspace and a small set of \emph{signals}, i.e., upstream components writing into that subspace, such that removing those contributions from the residual stream would cause the head to not attend to the token pair. ACC builds this subspace from the singular vectors of $\Omega$, leveraging their alignment with meaningful features in the residual stream.

\textbf{Tracing circuits with ACC++.} ACC++ identifies the signals causal for a single attention firing: a head $(\ell,a)$ attending to a token pair $(d,s)$. It proceeds in four steps (Figure~\ref{fig:accpp-method}): (a) fix the target head $(\ell,a)$ and token pair $(d,s)$; (b) decompose the destination residual $\v{x}^d$ as a sum of upstream component outputs; (c) form candidate signals by projecting each component output $\v{o}_i^d$ onto the left singular vectors $\{\v{u}_k\}$ of $\Omega$, so that every (component, direction) pair yields one rank-1 candidate; and (d) score candidates by their contribution to $A_{ds}$ using Integrated Gradients (IG)~\cite{sundararajan2017axiomatic} and greedily remove top-ranked candidates until $A_{ds} < \tau$. The source side is analogous, using right singular vectors. A complete and detailed description of ACC++ is provided in Appendices \ref{app:notation} through \ref{app:finding-tau}; see Appendix~\ref{app:new-acc:recipe}, Figure~\ref{fig:acc-recipe-new-v3}, for a more detailed summary of the method.

We define a \emph{circuit} for a prompt as the directed graph traced by applying ACC++ recursively across the model. Its nodes are model components at specific token positions, and each edge is a signal returned by ACC++, linking an upstream component to a downstream attention firing along a specific subspace. We build the circuit by starting at the components with a direct effect on the logit direction of interest (e.g., the correct-answer direction), running ACC++ on each attention firing in the current trace, and recursing upstream until none remain.

\textbf{ACC++ advances.} 
ACC++ makes two conceptual advances and one technical advance over previous work \cite{franco2025pinpointing}. On the conceptual side, ACC++ (1) dramatically sharpens the identification of candidate signals by identifying them more precisely with subspace-component pairs; and (2) more accurately measures attention causality by defining its objective directly on the head's nonlinear softmax attention weight. In comparison, prior work (1) identified subspaces and upstream components independently, leading to higher-dimensional and potentially less interpretable signals; and (2) used a linear surrogate for attention, which introduced spurious components in the trace.  We find that these conceptual advances yield signals which frequently admit a short natural-language description, and circuits that are much less noisy without sacrificing attention causality (Table \ref{tab:acc-vs-accpp}).

The technical advance made by ACC++ addresses heterogeneity in model architecture: we show that, under mild full-rank assumptions, it is always possible to construct a single QK matrix for each attention head, such that the matrix correctly computes attention regardless of whether bias or Rotary Positional Embeddings (RoPE) \cite{su2024roformer} is used in the particular model. Among other benefits, this simplifies the use of SVD-based analysis of QK matrices in real models.

\begin{wrapfigure}{R}{0.3\columnwidth}
\centering
\vspace{-1em} 
\resizebox{\linewidth}{!}{%
\begin{tikzpicture}[ 
  >={Latex[length=1.5mm]},
  comp/.style={draw, rounded corners=1.5pt, inner sep=2.5pt,
               font=\scriptsize, align=center},
  head/.style={draw, rounded corners=1.5pt, inner sep=2.5pt,
               font=\scriptsize, align=center, fill=gray!12},
  elab/.style={font=\scriptsize, inner sep=1pt},
  destcol/.style={red!75!black},
  srccol/.style={blue!70!black}
]
  \node[head]                    (down){AH$(\ell,a)$\\$(d,s)$};
  \node[comp, below left=12mm and 1mm of down]  (mlp) {MLP$(\ell')$\\$d$};
  \node[comp, below right=12mm and 1mm of down] (ah)  {AH$(\ell'',a')$\\$s$};
  \draw[->, destcol] (mlp) -- (down)
      node[elab, midway, left=-1pt, destcol]
           {$\v{u}_{k_1}\v{u}_{k_1}^{\!\top}\v{o}_1$};
  \draw[->, srccol]  (ah)  -- (down)
      node[elab, midway, right=-1pt, srccol]
           {$\v{v}_{k_2}\v{v}_{k_2}^{\!\top}\v{o}_2$};
\end{tikzpicture}}
\caption{ACC++ edges feeding the head $(\ell,a)$ attending $(d,s)$. A \textcolor{red!75!black}{destination} edge (red) comes from an upstream component writing at the destination token $d$ and carries the signal $\v{u}_k\v{u}_k^{\!\top}\v{o}_i$, using a left singular vector $\v{u}_k$. A \textcolor{blue!70!black}{source} edge (blue) comes from an upstream component writing at the source token $s$ and carries $\v{v}_k\v{v}_k^{\!\top}\v{o}_i$, using a right singular vector $\v{v}_k$. 
}
\label{fig:accpp-edge}
\vspace{-2em}
\end{wrapfigure}

\textbf{Precise signal candidates.} 
ACC++ leverages the empirical observation that singular vectors correspond to communication channels.  The first conceptual advance in ACC++ lies in how it uses this observation to define specific candidate signals. A candidate signal corresponds to one upstream component writing in one singular vector direction. Figure~\ref{fig:accpp-edge} shows the relationship between upstream components, signals, and the downstream head.

Each candidate signal $\v{s}_i$ induces a vector $\v{c}_i$ of pre-softmax score contributions across source token positions. For example, in the destination view, consider a candidate signal $\v{s}_i$ corresponding to upstream component $m$ projected onto singular direction $k$, i.e., $\v{s}_i = \v{u}_k \v{u}_k^{\top} \v{o}_m$ (Figure~\ref{fig:accpp-method}c). 
The contribution  of signal $\v{s}_i$ to the pre-softmax attention score of source position $j$ is $[\v{c}_i]_j = \v{s}_i^\top \Omega \v{x}^j$. 
Figure~\ref{fig:accpp-edge} illustrates both edge types: a destination edge from an upstream component writing at $d$ carries the signal $\v{u}_k\v{u}_k^{\top}\v{o}_1$, while a source edge from an upstream component writing at $s$ carries $\v{v}_k\v{v}_k^{\top}\v{o}_2$. Thus $\v{c}_i$ collects the contribution of that candidate signal across all source positions. The source-side view is analogous, with the candidate contribution entering on the key side.  Note that this is a complete accounting for pre-softmax attention. Namely, summing across all candidate-induced vectors recovers the exact pre-softmax score vector for the destination token: $A'_d \;=\; \sum_{i} \v{c}_i$. 


A key strategy taken in ACC++ is to identify the set of candidate signals as the cross-product of \emph{all} outputs of upstream components with \emph{all} singular vector channels of the head. This contrasts with previous work \cite{franco2025pinpointing}, which identified components and channels separately. By considering each possible combination of an upstream component and a singular vector, ACC++ defines more precise candidate signals.

\textbf{Accurate attention accounting.} Once candidate signals have been defined, ACC++ identifies a small set whose removal, considered as a counterfactual, would cause the chosen attention weight $A_{ds}$ to fall below a threshold $\tau$ (Figure~\ref{fig:accpp-method}d). 
(We discuss the rationale for our choice of $\tau$ in Appendix~\ref{app:finding-tau}.)

There are two cases of this counterfactual: \emph{destination} signals, removed from the destination residual $\v{x}^d$ and affecting the query, and \emph{source} signals, removed from the source residuals $\{\v{x}^j\}_{j\le d}$ and affecting the competing keys (Figure~\ref{fig:accpp-edge}). Because ACC++ uses softmax to compute $A_d$, it must evaluate the nonlinear impact of signal removal using the full row $A_d$, not just the entry corresponding to source token $s$. Under softmax, a candidate is important not because it increases $A'_{ds}$ in isolation, but because it increases that score relative to the competing source tokens, thereby increasing $A_{ds}$.

Since softmax is nonlinear, exact identification of the minimal set of candidate signals to remove is challenging.  To address this efficiently, ACC++ uses a greedy strategy based on an estimate of the nonlinear impact of each signal removal.  ACC++ first ranks candidate signals by how much each affects the true attention weight $A_{ds}$, using IG.
Given these attribution scores, ACC++ solves the counterfactual by greedy removal. Starting from the full set of candidates, it repeatedly removes the highest-scoring candidate, recomputes the attention weight, and stops once the chosen weight falls below $\tau$. The selected candidates are signals in the resulting trace. 

This is a key point where ACC++ diverges from prior work \cite{franco2025pinpointing} which only operated on pre-softmax scores $A'$ via a linear surrogate.  In contrast, ACC++ evaluates the nonlinear softmax explicitly, allowing it to identify much smaller circuits that are still causal for attention, as we show below.

\begin{wraptable}[14]{R}{0.55\columnwidth}
\centering
\scriptsize
\caption{Comparison of ACC and ACC++ graph sizes across tasks and models. Entries are mean $\pm$ std. error. 
}
\label{tab:acc-vs-accpp}
\setlength{\tabcolsep}{2pt}
\renewcommand{\arraystretch}{0.95}
\resizebox{\linewidth}{!}{%
\begin{tabular}{cccccc}
\toprule
                     &             & \multicolumn{2}{c}{Nodes} & \multicolumn{2}{c}{Edges}     \\
Task                 & Model       & ACC          & ACC++      & ACC            & ACC++        \\ \hline
\multirow{3}{*}{IOI} & Gemma-2 2B  & 688 $\pm$ 12 & 87 $\pm$ 3 & 7086 $\pm$ 200 & 215 $\pm$ 11 \\
                     & GPT-2 Small & 205 $\pm$ 2  & 85 $\pm$ 1 & 639 $\pm$ 9    & 154 $\pm$ 3  \\
                     & Pythia-160M & 123 $\pm$ 2  & 76 $\pm$ 1 & 354 $\pm$ 8    & 147 $\pm$ 4  \\ \hline
\multirow{3}{*}{GP}  & Gemma-2 2B  & 121 $\pm$ 20 & 8 $\pm$ 0  & 806 $\pm$ 159  & 7 $\pm$ 1    \\
                     & GPT-2 Small & 114 $\pm$ 3  & 27 $\pm$ 1 & 263 $\pm$ 10   & 28 $\pm$ 1   \\
                     & Pythia-160M & 63 $\pm$ 3   & 33 $\pm$ 2 & 132 $\pm$ 8    & 52 $\pm$ 4   \\ \hline
\multirow{2}{*}{GT}  & GPT-2 Small & 176 $\pm$ 3  & 45 $\pm$ 1 & 594 $\pm$ 12   & 81 $\pm$ 2   \\
                     & Pythia-160M & 78 $\pm$ 4   & 43 $\pm$ 5 & 177 $\pm$ 13   & 88 $\pm$ 12  \\ 
\bottomrule
\end{tabular}
}
\end{wraptable}

\textbf{Unified QK analysis.} ACC++ also makes technical contributions for analyzing QK matrices.  Expressing an attention score $A'_{ds}$ as a bilinear form $\v{x}^{d\top} W_Q W_K^\top \v{x}^s$ overlooks details that vary across model architecture.  In particular, some models add a bias term to the attention calculation, and some models use RoPE positional encoding.  Both of these prevent using a single $\Omega = W_QW_K^\top$ matrix to describe the attention computation on a per-attention head basis.  ACC++ overcomes these challenges by folding the bias and RoPE rotations into the tokens $\v{x}^d$ and $\v{x}^s$.  In the case of bias, ACC++ adds a function of the bias to the tokens, and in the case of RoPE it performs a suitable rotation on the tokens, \emph{before computing attention}.  We describe the specifics in Appendix \ref{app:unified-bilinear}, which presents proofs that after these transformations on the tokens, the correct attention score is computed as a simple bilinear form of $\Omega$.  This vastly simplifies the analysis of $\Omega$ using SVD since there is only one $\Omega$ per head, independent of bias or token positions.

 \textbf{Evaluation.} We use ACC++ to trace circuits for Indirect Object Identification (IOI) \cite{DBLP:conf/iclr/WangVCSS23}, Greater-Than (GT) \cite{hanna2023does}, and Gender Pronoun (GP) \cite{athwin2identifying} in GPT-2 Small \cite{radford2019language}, Pythia-160M \cite{biderman2023pythia}, and Gemma-2 2B \cite{team2024gemma}. We show that ACC++ recovers circuits with high overlap (precision, recall, and F1 against reference circuits, following~\cite{franco2025pinpointing}) with ACC circuits and with patching-based methods (Path Patching~\cite{DBLP:conf/iclr/WangVCSS23}, EP~\cite{bhaskar2025finding}, ACDC~\cite{conmy2023automatedcircuitdiscoverymechanistic}, EAP~\cite{syed2024attribution}). Full results are in Appendix~\ref{app:reproducing-previous-results}.

The ACC++ signals present in these circuits are lower-dimensional than ACC's: across IOI, GP, and GT, the majority use a \emph{single} singular direction of $\Omega$ (Appendix~\ref{app:reproducing-previous-results}, Figures~\ref{fig:sparse-attn-decomposition} and~\ref{fig:distrib-n-svs}). Moreover, we ablate subsets of the signals selected by ACC++ on IOI and show that these signals also often have a measurable causal effect on the model's IOI performance while only minimally changing the residual stream, with post-ablation cosine similarity close to $0.99$ and norm ratio close to $1$; full results are reported in the Appendix~\ref{app:reproducing-previous-results}.  

Table~\ref{tab:acc-vs-accpp} shows that \textbf{ACC++ circuits are typically 2$\times$ to 10$\times$ smaller} than ACC traces while preserving the traced attention decisions under our counterfactual attention-weight criterion. Finally, we emphasize that ACC++ tracing is automatic. Apart from specifying a value for $\tau$ (for which ACC++ incorporates reasonable defaults, see Appendix \ref{app:finding-tau}) there is no tuning involved, no replacement models to train, and no counterfactual prompts to formulate.

\textbf{Automatic interpretation of ACC++ signals.}
Next, we interpret signals by constructing a short natural language description of each signal (i.e., edge in the trace).
We use the fully automated autointerpretation framework of \cite{paulo2025automatically} for SAE features and adapt it to ACC++ signals.

Starting from a signal identified by ACC++, we (i) retrieve high-activation contexts by scoring token pairs in cached activations from The Pile \cite{gao2020pile} (via \texttt{pile-10k} \cite{ThePile}), (ii) prompt an LLM to propose a short natural-language description of the signal from these contexts, and (iii) evaluate the description with the a LLM-as-a-judge protocol of \cite{paulo2025automatically} by mixing top contexts with random contexts. Full details are presented in Appendix~\ref{app:interp-signals}. We evaluate this pipeline on a large set of signals (13,406 from GPT-2 small, 15,316 from Pythia-160M, and 22,223 from Gemma-2 2B) generated by a random sample of 100 prompts from the IOI task.

\textbf{A substantial fraction of signals are interpretable.} Table~\ref{tab:interpretability-comparison} reports quantitative interpretability for the signals across the three models. The Table shows that ACC++ signals achieve autointerpretation performance that approaches SAE-feature baselines. To put these results in context, we note that SAE features are explicitly trained to separate superposed features, while ACC++ signals arise directly from causal attention interactions.

\begin{wraptable}[12]{R}{0.5\columnwidth}
\centering
\caption{Autointerpretation scores for ACC++ signals on IOI. We report medians with interquartile range (25\%--75\%); the SAE column is from \cite{paulo2025automatically}.}
\label{tab:interpretability-comparison}
\footnotesize
\setlength{\tabcolsep}{2.5pt}
\renewcommand{\arraystretch}{1.15}
\resizebox{\linewidth}{!}{%
\begin{tabular}{@{}lcccc@{}}
\toprule
Metric & SAE & GPT-2 & Pythia & Gemma \\
\midrule
Accuracy &
\shortstack{$0.76$\\{\scriptsize $(0.67\text{--}0.86)$}} &
\shortstack{$0.70$\\{\scriptsize $(0.60\text{--}0.80)$}} &
\shortstack{$0.65$\\{\scriptsize $(0.53\text{--}0.76)$}} &
\shortstack{$0.60$\\{\scriptsize $(0.53\text{--}0.70)$}} \\
Precision &
-- &
\shortstack{$0.68$\\{\scriptsize $(0.57\text{--}0.79)$}} &
\shortstack{$0.62$\\{\scriptsize $(0.52\text{--}0.75)$}} &
\shortstack{$0.58$\\{\scriptsize $(0.51\text{--}0.68)$}} \\
Recall &
-- &
\shortstack{$0.85$\\{\scriptsize $(0.70\text{--}0.95)$}} &
\shortstack{$0.85$\\{\scriptsize $(0.65\text{--}0.95)$}} &
\shortstack{$0.80$\\{\scriptsize $(0.65\text{--}0.90)$}} \\
\bottomrule
\end{tabular}
}
\end{wraptable}


We also evaluate when a signal should count as \emph{interpretable} under a hypothesis-testing criterion. Given a candidate interpretation and judge scores, we compute a one-sided $p$-value using Fisher's exact test \cite{fisher1922interpretation} to test whether the judge accepts the interpretation more often on top contexts than on random contexts. We then control the false discovery rate (FDR) at $5\%$ across signals and report the fraction of signals passing this criterion, which we label \emph{interpretable}. This measure of interpretability is meant to distinguish signal from noise, rather than to fully quantify the quality of the interpretation.

We then estimate the fraction of interpretable signals under our criterion. We find that the fraction of interpretable signals varies across models: 63\% (95\% CI: 62--64\%) in GPT-2 Small, 50\% (49--51\%) in Pythia-160M, and 31\% (30--32\%) in Gemma-2 2B. This result suggests that many ACC++ signals correspond to consistent attention interactions rather than noise. 

\begin{figure*}[h]
  \centering
    \includegraphics[angle=0, width=\textwidth]{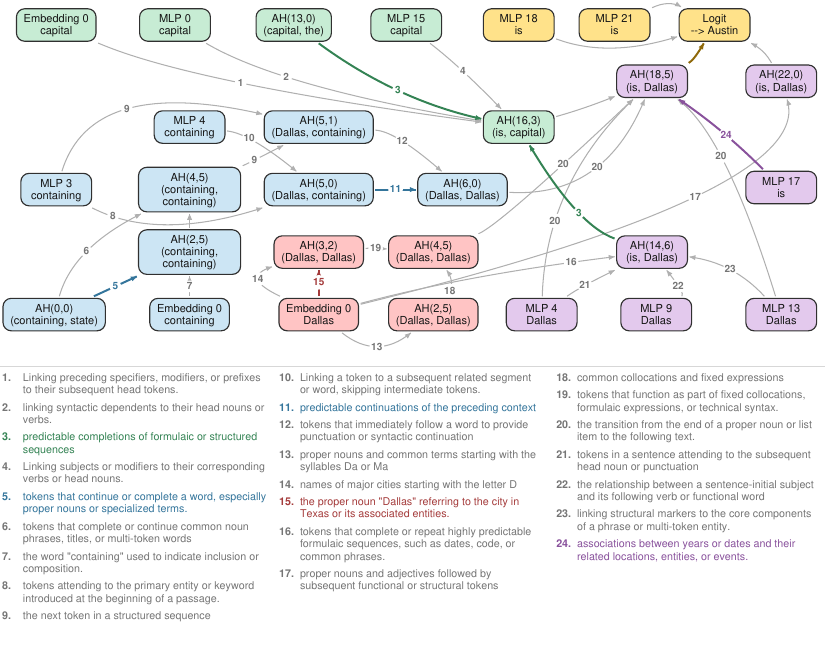}
  \caption{The ACC++ trace shows how Gemma-2 2B answers `Fact: The capital of the state containing Dallas is' with `Austin.' AH node labels are (destination token, source token); the number labels in the edges refer to automatically generated interpretations presented below the traced graph. Red: entity recognition; blue: relational parsing; green: query understanding; purple: information integration; yellow: answer generation.}
  \label{fig:gemma-facts-0}
\end{figure*}

\textbf{ACC++ outputs.} The output of ACC++ is a graph in which nodes are model components and edges are the signals causal for attention, annotated with their natural language interpretation. To illustrate an application of ACC++, we take an example from \cite{lindsey2025biology}: `Fact: The capital of the state containing Dallas is,' for which Gemma-2 2B correctly answers `Austin.' We show in Figure~\ref{fig:gemma-facts-0} the interpretable circuit for this prompt.  Each signal is annotated with its automatically generated interpretation, and key signals are in color.  The circuit provides a number of insights.

First, it shows that the model immediately recognizes  \emph{the city of Dallas,}  based on features present in the input embedding of the token `Dallas.'  Next, it shows a path (in blue) matching `state' with `containing' based on features that promote \emph{completion of a phrase particularly involving proper nouns.}  In this path, information about the notion of state is moved to `containing,' and then `containing' is matched with `Dallas,' causing `Dallas' to be annotated with the feature \emph{predictable continuations.}

The query itself is understood by integrating `capital' with `is' in head (16, 3). Here too, the features emphasize \emph{predictable completions.}  Finally and crucially, MLP 17 adds a feature to `is' causing head (18, 5) to look for \emph{geographic associations} in its source token (`Dallas').  Head (18, 5) then moves `Austin' to the end token (`is') and the `Austin' annotation is reinforced by inputs from MLPs 18 and 21, and head (22, 0).

Comparing to the methodology of \cite{lindsey2025biology}, we note that Figure~\ref{fig:gemma-facts-0} adds considerable detail to the picture of how the model computes its output; unlike \cite{lindsey2025biology}, ACC++ identifies the \emph{specific heads and MLPs} that implement the circuit, and pinpoints the precise information flow between those components to illustrate how the computation progresses.

We show more examples of automatically generated signal interpretations in Figure~\ref{fig:abba-baba-circuits-gpt}, and additional traces in Appendix~\ref{app:ioi-task} (Figures~\ref{fig:abba-baba-circuits-gpt-proper-noun} and~\ref{fig:pythia-template-circuits}) and Appendix~\ref{app:icl-translation} (Figure~\ref{fig:gemma-facts-1}). To aid in circuit inspection, ACC++ provides a user dashboard in which the interpretation of the signals in each circuit can be browsed and the evidence can be examined, as illustrated in Appendix~\ref{app:trace-examples}, Figures~\ref{fig:dallas-example1}--\ref{fig:dallas-example5}.

%% file: sections/example_applications.tex
\section{ACC++ yields mechanistic insights into model behavior}
\label{sec:uses}

\begin{wrapfigure}{R}{0.69\columnwidth}
\centering
\includegraphics[width=\linewidth]{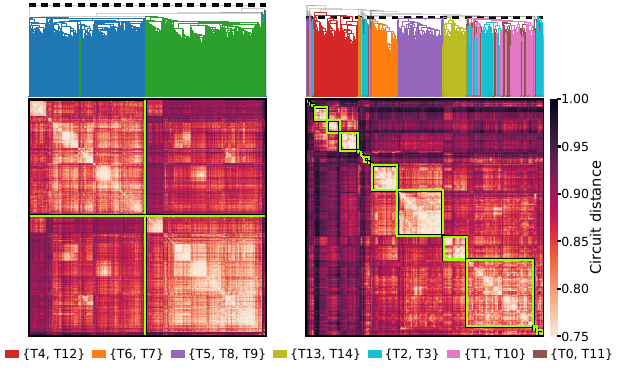}
\caption{Average linkage clustering of prompt-level traces exposes the sensitivity of IOI circuits to prompt structure. Left: GPT-2 small. Right: Pythia 160M. Branches in the dendrograms are colored by high-level template (\textcolor[HTML]{1f77b4}{ABBA} vs. \textcolor[HTML]{2ca02c}{BABA}) or by low-level template group (see legend below).}
\label{fig:clustermap-sv-as-component}
\end{wrapfigure}

ACC++ produces a circuit per prompt whose nodes are model components and whose edges carry the causal signals driving attention (\S~\ref{sec:improv-acc}). These circuits contain rich information about how the model produced its prediction, and much of it can be put into human-readable form. 
We use these interpretable signals to derive mechanistic insights in two settings: (1) the sensitivity of IOI circuits to prompt structure, including a natural-language account of how the model retrieves the indirect object (\S~\ref{sec:ioi-word-order}); (2) how mGPT solves multilingual IOI through a mix of language-independent and language-specific signals (\S~\ref{sec:multilingual-ioi}).

\subsection{Understanding IOI circuit sensitivity to prompt structure} \label{sec:ioi-word-order}

We first use ACC++ to revisit IOI, a canonical benchmark in mechanistic interpretability \cite{DBLP:conf/iclr/WangVCSS23, conmy2023automatedcircuitdiscoverymechanistic, Makelov_Lange_Nanda_2023, bhaskar2025finding, syed2024attribution, merullo2024circuit, tigges2024llm, hanna2024have, haklay2025position, meloux2025mechanistic, Ahmad:NeurIPS2025, franco2025pinpointing, garriga2024adversarial, wu2025query}. The IOI task has sentences of the form `When Mary and John went to the store, John gave the bottle to,' where the model should answer `Mary' instead of  `John.' Prior work has studied IOI using prompt templates and has highlighted the importance of positional structure in circuit analysis \cite{ haklay2025position}. Here we ask a more specific question: \emph{how much of the variation across prompts is driven by prompt structure, and how can signal-level traces explain the difference?}

We use a collection of IOI prompts balanced across two levels of templates: two high-level templates (reversing the role order of the two names, ie, ABBA vs BABA), and 15 low-level templates (changing surface wording, eg `While ...' vs `Then...').
Details are in Appendix~\ref{app:clustering-dataset}.

\begin{wrapfigure}{R}{0.4\columnwidth}
    \centering
    \includegraphics[width=\linewidth]{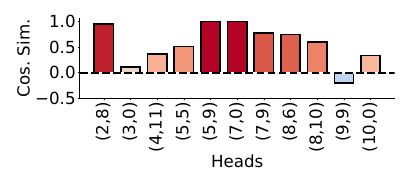}
    \caption{Signal similarity uncovers common and distinct functionality across ABBA vs. BABA circuits.}
    \label{fig:signals-comp-models}
\end{wrapfigure}

\textbf{Circuits cluster by prompt format.} For each model, we trace a circuit with ACC++ on every prompt where the model predicts the correct indirect object. To compare traces, we represent each circuit as a set of pairs of edges and singular vectors, calculate pairwise Jaccard distances between circuits from different prompts (which we refer to as \emph{circuit distance}), and group them using average-linkage hierarchical clustering. Details in Appendix~\ref{app:clustering}. 

Figure~\ref{fig:clustermap-sv-as-component} summarizes how prompt-level circuits organize. Each panel shows the matrix of pairwise circuit distances, with rows and columns reordered by the clustering. Dendrograms on top of the figure are colored by high-level template (\textcolor[HTML]{1f77b4}{ABBA} vs.\ \textcolor[HTML]{2ca02c}{BABA}) or by low-level template group, and the horizontal dashed line marks the cut height that defines the clusters outlined by green boxes on the diagonal. In GPT-2 Small, the dendrogram splits cleanly into one ABBA cluster (left) and one BABA cluster (right), with no clear substructure related to low-level template. On the other hand, Pythia-160M has a larger number of smaller clusters, each aligned with low-level template groups. This shows that IOI prompt sensitivity is strong in both models, but the relevant form of variation is model-dependent: \emph{GPT-2 is strongly sensitive to role order, whereas Pythia is more sensitive to surface wording.}

\textbf{ACC++ explains prompt-level circuit variation.} The substantial prompt-level variation of IOI circuits is highly structured: prompts form coherent clusters of circuits for solving the same task. To characterize what drives this sensitivity, \trim{without analyzing every circuit individually,} we summarize each cluster by a single \emph{representative circuit}: the traced graph whose component set has the smallest average circuit distance to the other circuits in the cluster (Appendix~\ref{app:clustering-representative}). This gives a stable unit of analysis for comparing different clusters.

To compare the signals used by each representative, we aggregate the incoming ACC++ signals of each attention head (see Figure~\ref{fig:accpp-edge}) into a single destination signal summary and source summary, and then compute cosine similarities between these signal summaries across representatives of the clusters (Appendix~\ref{app:clustering-signals}). Figure~\ref{fig:signals-comp-models} shows the destination-signal comparison for the GPT-2 ABBA and BABA representatives. \trim{Each bar represents an attention head that appears in both representatives, and shows the cosine similarity between incoming signals to that component.} The figure shows that many signals are reused across templates. However, the name-mover head $(9,9)$ receives markedly different input signals (negative cosine similarity) between ABBA and BABA, even though these signals are carried on the same token. Inspection shows that this is a critical difference that helps explain the separation of circuits for the two prompt types into distinct clusters. More details and other models comparisons are in Appendix~\ref{app:ioi-task}.

\begin{figure}[htbp]
\centering
\begin{subfigure}{0.45\columnwidth}
  \centering
  \includegraphics[width=\linewidth]{figures/interp-signals/abba_circuit_rotated_latex.pdf}
  \caption{ABBA: ``While Justin and Daniel were working at the house, Daniel gave a snack to"}
  \label{fig:abba-circuit-gpt}
\end{subfigure}
\hfill
\begin{subfigure}{0.45\columnwidth}
  \centering
  \includegraphics[width=\linewidth]{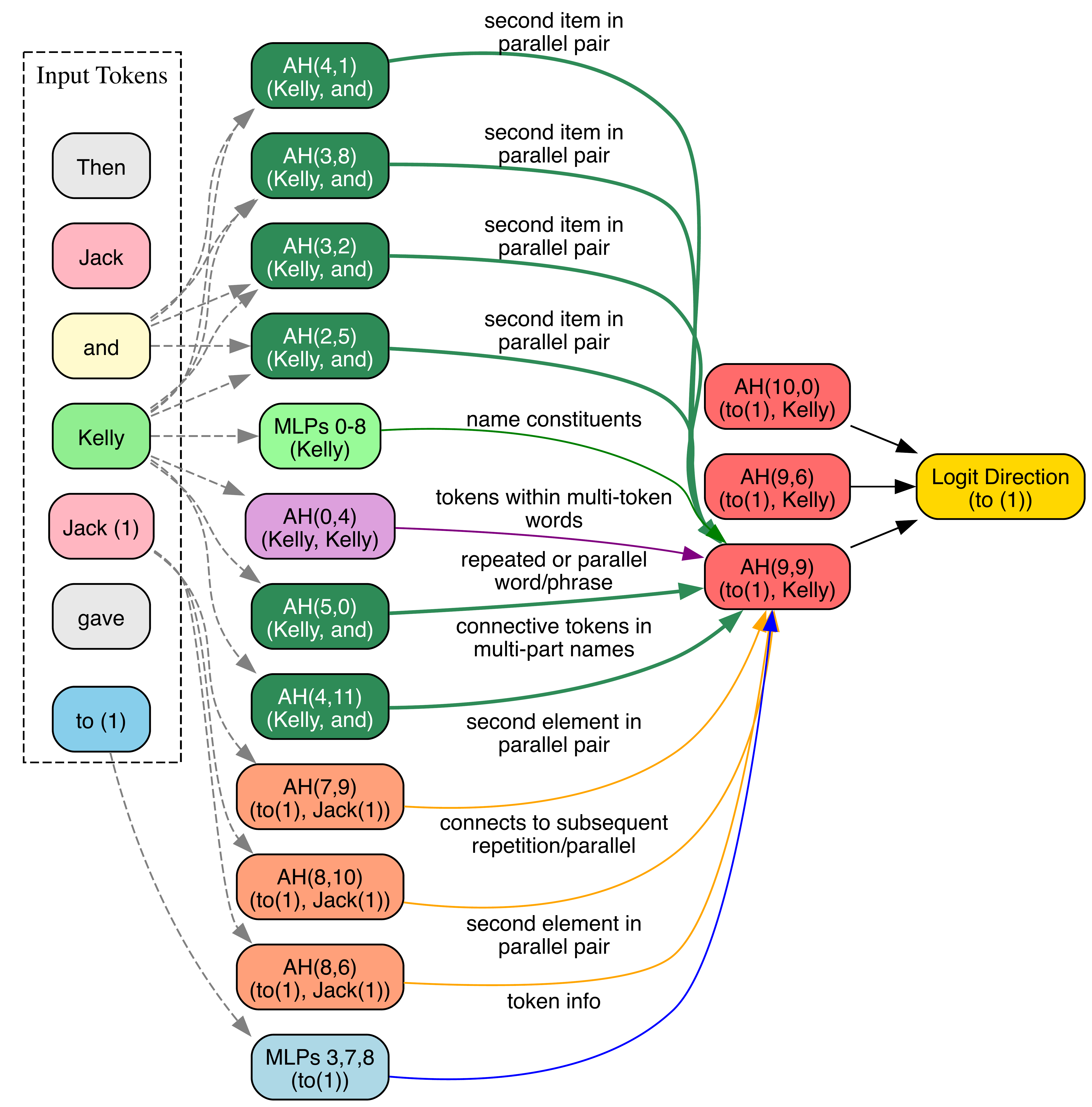}
  \caption{BABA: ``Then, Jack and Kelly went to the garden. Jack gave a basketball to"}
  \label{fig:baba-circuit-gpt}
\end{subfigure}
\caption{Traces are interpretable and expose algorithmic differences in circuits between ABBA and BABA in GPT-2. AH node labels are (destination token, source token); edge labels are automatically generated. Red: feeds logits, purple: low-level, orange: provide inhibition signal. Dark green nodes show that, in BABA only, the circuit relies on identifying the ``second item in a parallel pair'', i.e.\ ``Kelly'', as the appropriate output.}
\label{fig:abba-baba-circuits-gpt}
\end{figure}

Interpreting the signals involved (\S~\ref{sec:improv-acc}) provides further insight on why the (9,9) head processes different inputs in the two cases. \trim{Using the natural language descriptions for signals , we can obtain a mechanistic explanation for the difference between clusters. The clearest difference is the input to head $(9,9)$, so we focus on the portion of the traces upstream of that head. We then compare the GPT-2 Small ABBA and BABA representative circuits, labeling each edge with its autointerpreted signal.}
Figure~\ref{fig:abba-baba-circuits-gpt} shows the circuit components upstream of head $(9,9)$ in each template representative (details on how these qualitative trace figures were constructed from the annotated interactive traces are given in Appendix~\ref{app:trace-examples}). In the BABA case, the indirect object appears after the subject, and the model tags it with the feature ``second item in a parallel pair.'' Four different heads each contribute this feature to the ``Kelly'' token. At the same time, head $(8,6)$ suppresses the ``Jack'' token by writing the same feature into ``to''. Head $(9,9)$ then attends to the pair (to, Kelly) as a result of this feature match, moving the indirect object to the output.

In contrast, in the ABBA circuit the indirect object comes before the subject and so cannot be identified as the ``second item.'' The key features used to identify it are instead more generic structural cues, such as ``structural or sequence-completing token'' and ``tokens in parallel structures that repeat or mirror a preceding element.'' Thus the two templates reuse a canonical IOI component, but feed it with different upstream mechanisms: one centered on identifying the second item in a pair, and one centered on broader structural matching.

We note the value of ACC++ here: by producing circuits with interpretable signals, ACC++ does not just recover that GPT-2 uses different circuits for ABBA and BABA; it shows that the same head (9,9) can implement its function using different incoming signals, and lets us describe what those signals are doing.

\subsection{Multilingual IOI: shared mechanisms, language-specific signals} \label{sec:multilingual-ioi}


The sensitivity of IOI circuits to prompt structure shows that even when components are reused across task variants, the signals involved may differ in important ways.  We next explore this phenomenon in a setting where differences between prompt types are substantially greater: multilingual task variation.  We extend the IOI task by constructing a multilingual IOI dataset spanning English, Spanish, French, and Portuguese prompts (Appendix~\ref{app:mioi-task}). \trim{This new dataset allow us to find whether models reuse mechanisms across languages, and more importantly, whether those shared mechanisms carry language-specific signals.} The results in this section use the mGPT model \cite{shliazhko2024mgpt}, but a similar analysis holds for Gemma-2 2B (Appendix~\ref{app:mioi-task})

\begin{wrapfigure}{R}{0.4\columnwidth}
\centering
\includegraphics[width=\linewidth]{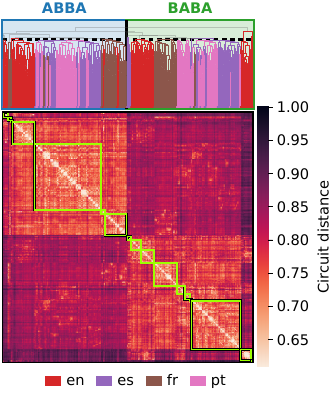}
\caption{Average linkage clustering of prompt-level traces exposes grouping by high-level template first then by language. Branches in the dendrograms are colored by language.}
\label{fig:multilingual-ioi-clustermap}
\end{wrapfigure}


\textbf{Components are reused; signals, not necessarily.} Prior work has shown that models can use the same components when processing IOI prompts and their translation into another language \cite{zhang2025the}. We first confirm that ACC++ also recovers this component reuse (Appendix~\ref{app:mioi-results}, Figure~\ref{fig:multilingual-ioi-heatmap}).

However, analyzing signals via ACC++ shows \emph{how} components are reused. Thus, we can analyze if components reuse the same signals across languages, or if these components carry language-specific signals.  Figure~\ref{fig:multilingual-ioi-clustermap} shows the clustering results using the same pipeline as \S~\ref{sec:ioi-word-order},
which summarizes how multilingual traces cluster in mGPT.  Importantly, clustering is based on signal use as well as component use, and so shows more than just component reuse.
As with the monolingual case, the primary clustering feature is template structure: prompts sharing the same high-level template group together regardless of language. 
This indicates that models carry some language-agnostic signals. However, within each high-level template cluster there is a \emph{secondary} grouping: prompts cluster by language with sub-clusters corresponding to English, Spanish, French, and Portuguese.  This secondary clustering indicates that together with shared mechanisms, circuits also incorporate language-specific signals.

\begin{wrapfigure}{R}{0.5\columnwidth}
    \centering
    \includegraphics[width=\linewidth]{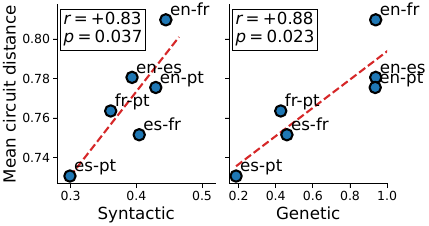}
    \caption{Mean circuit distances between mGPT language pairs strongly correlate with URIEL+ syntactic (left) and genetic (right) distances.}
    \label{fig:multilingual-ioi-uriel}
\end{wrapfigure}

\textbf{Circuit distance correlates with linguistic distance.} In Figure~\ref{fig:multilingual-ioi-clustermap}, Spanish and Portuguese mix more than the other languages, suggesting that more closely related languages share more signals. To test whether the secondary language-specific clustering is meaningful, we correlate our cross-language circuit distances with typological distance between languages. For each language pair, we compute the average Jaccard distance between paired prompt traces (matched within high- and low-level templates) and correlate it against measures of language similarity as provided by URIEL+~\cite{khan2025uriel+}. Figure~\ref{fig:multilingual-ioi-uriel} shows positive Pearson correlations with URIEL+ syntactic ($r=0.83$) and genetic ($r=0.88$) distance for mGPT, with the same pattern in Gemma-2 2B ($r=0.80$ and $r=0.84$, respectively; see Appendix~\ref{app:mioi-results} for the corresponding figure and methodology). Although the panel of six language pairs is small, the consistency of the trend across two models as well as both syntactic and genetic distances suggests that cross-language circuit distance is not arbitrary, and instead mirrors the relatedness of these languages.

%% file: sections/related_work.tex
\section{Related work}\label{sec:rel-work}

A large body of work suggests that many features are represented in low-dimensional directions \cite{Elhage22,olah2020zoom,mikolov-etal-2013-linguistic,alain2018understandingintermediatelayersusing,park2023the,DBLP:conf/iclr/GurneeT24,Gurnee_Nanda_Pauly_Harvey_Troitskii_Bertsimas_2023,Marks_Rager_Michaud_Belinkov_Bau_Mueller_2024,levy2025languagemodelsencodenumbers,engels2025languagemodelfeaturesonedimensionally,kantamneni2025languagemodelsusetrigonometry,hernandez2024linearityrelationdecodingtransformer}. Sparse Autoencoders (SAEs) are widely used to extract such features \cite{huben2024sparse,BrickenSAEs}, cross-layer transcoders extend this idea to prompt-level attribution graphs over a learned replacement model \cite{lindsey2025biology}, and Low-Rank Sparse Attention applies the same surrogate-model strategy to attention layers themselves \cite{he2026towards}; each having known drawbacks, including identifiability and stability issues \cite{leask2025sparseautoencoderscanonicalunits,gao2024scalingevaluatingsparseautoencoders,bussmanSAEs,chanin2024absorptionstudyingfeaturesplitting} and a reliance on activations alone that does not use model parameters \cite{chugtaiActivationSpaceDoomed}. \cite{arora2026language} avoids surrogates by showing that MLP neurons themselves form a sparse feature basis suitable for circuit tracing using gradient-based attribution. ACC++, similarly to \cite{ franco2025pinpointing}, considers \emph{both} activations and weights (QK matrices) in identifying signals, and operates directly on the original model rather than on a replacement model.

Several recent studies also use the singular value decomposition (SVD) of attention interactions to study how heads communicate through low-dimensional subspaces. The work in \cite{merullo2024talkingheadsunderstandinginterlayer} shows that inter-layer communication can be identified with subspaces aligned with singular vectors of attention heads, and \cite{FrancoEtAl:ICML26} shows that singular vectors themselves can align with attention-causal features. The work in \cite{Ahmad:NeurIPS2025} uses an optimization procedure to identify task-relevant directions in the SVD basis of an attention head, and shows that a single head can simultaneously implement multiple independent computations. In a different setting, \cite{NEURIPS2024_6216515a} studies token interactions in vision transformers and treats left/right singular vectors of the query--key interaction as paired interacting directions. ACC++ builds on this line by using SVD directions as \emph{candidate communication channels} and pairing them with upstream components to form precise signal candidates (\S~\ref{sec:improv-acc}).

For circuit tracing, many methods rely on counterfactual interventions such as activation patching \citep{zhang2024bestpracticesactivationpatching,goldowskydill2023localizingmodelbehaviorpath,DBLP:conf/iclr/WangVCSS23,conmy2023automatedcircuitdiscoverymechanistic,Hanna_Liu_Variengien_2023,Lieberum_Rahtz_Kramár_Nanda_Irving_Shah_Mikulik_2023,hanna2024have}. Patching is time-consuming, requiring many forward passes, often requires constructing a counterfactual dataset, and has multiple documented weaknesses \citep{Makelov_Lange_Nanda_2023,Mueller_2024,McGrath_Rahtz_Kramar_Mikulik_Legg_2023,Rushing_Nanda_2024,garriga2024adversarial}. The work in \cite{meloux2025mechanistic} reports that EAP-IG traces \cite{hanna2024have} can show high structural variance across models and tasks. Moreover, \cite{wu2025query} pursues per-prompt circuit discovery, ensembling EAP-IG over query paraphrases to find faithful query-specific circuits. ACC++ is complementary: it traces a per-prompt circuit from a single forward pass and exposes the signals that causes each attention decision, rather than only the component-to-component edges that drive overall performance.

%% file: sections/conclusions.tex
\section{Conclusions} \label{sec:conclusions}


We present ACC++, a practical tool for circuit tracing and mechanistic analysis. From a single forward model pass and without replacement models, activation patching, or hand-designed counterfactual inputs, it produces circuits whose components are causal for attention and whose edges admit a natural-language interpretation in a substantial fraction of cases. By making per-prompt circuits a viable unit of analysis at scale, ACC++ circuits uncover insights regarding the sensitivity of circuits to prompt formats, and the role of signals across languages.

\textbf{Limitations.} ACC++ has some limitations that we leave as future work. First, the method depends on setting the threshold~$\tau$, and we use a heuristic based on the distribution of the $d \cdot A_{ds}$ per model/dataset. A principled, dataset-agnostic rule for~$\tau$ is left to future work. Second, ACC++ circuits do not trace upstream components that caused FFNs to output an attention-causal signal. Third, because our signal interpretations rely on an LLM as a judge, the interpretations are suggestive but not conclusive. Finally, the multilingual analysis of \S~\ref{sec:multilingual-ioi}, including the correlation with URIEL+ distances, is based on a small number of language pairs and should also be read as suggestive rather than conclusive.

%% file: sections/acks.tex
Carson Loughridge suggested the methodological improvement for models with bias described in Appendix~\ref{app:bias-technique}. Aditya Saini made improvements to the autointerpretation pipeline. This work also benefited from early feedback from Aaron Mueller, Micah Benson, Divya Appapogu, Freya Behrens, Najoung Kim, Yukyung Lee, and Micah Adler. This research was funded by a grant from Coefficient Giving and by NSF award CNS-2312711.

%% file: sections/appendix_index.tex
\startcontents[appendices]
\begingroup
\hypersetup{hidelinks}
\small
\setlength{\parskip}{0pt}
\titlecontents{section}
  [1.6em]
  {\addvspace{4pt}\bfseries}
  {\contentslabel{1.6em}}
  {}
  {\titlerule*[0.5pc]{.}\contentspage}
\titlecontents{subsection}
  [3.7em]
  {}
  {\contentslabel{2.1em}}
  {}
  {\titlerule*[0.5pc]{.}\contentspage}
\section*{Appendix contents}
\printcontents[appendices]{}{1}{\setcounter{tocdepth}{2}}
\endgroup
\clearpage

%% file: sections/appendix.tex
\input{sections/appendix/notation}
\clearpage
\input{sections/appendix/unified_bilinear_form}
\clearpage
\input{sections/appendix/new_acc}
\clearpage
\input{sections/appendix/finding_tau}
\clearpage
\input{sections/appendix/prev_acc_results}
\clearpage
\input{sections/appendix/signals_interp_proof}
\clearpage
\input{sections/appendix/clustering}
\clearpage
\input{sections/appendix/ioi_task}
\clearpage
\input{sections/appendix/mioi_task}

\clearpage
\input{sections/appendix/icl_and_translation}
\clearpage

\section{Compute resources}
\label{app:compute}

\paragraph{ACC vs.\ ACC++ comparisons.}
The experiments in Appendix~\ref{app:reproducing-previous-results} were run on a Mac with an M4 chip and 128~GB memory RAM.

\paragraph{IOI circuit tracing.}
For GPT-2 Small, Pythia-160M, and mGPT, tracing was performed on CPU cluster nodes with 28 cores and 128~GB of RAM, taking approximately 5--10 seconds per prompt. For Gemma-2 2B, tracing was performed on nodes with NVIDIA A100 40~GB GPUs, taking approximately one minute per prompt. The balanced IOI dataset contains 3{,}000 prompts evaluated across the three models.

\paragraph{Autointerpretation.}
The pipeline in Appendix~\ref{app:interp-signals}, including activation collection, context retrieval, interpretation, and evaluation, was run on cluster nodes with four NVIDIA L40S GPUs each. Processing 13{,}406 GPT-2 Small signals, 15{,}316 Pythia-160M signals, and 22{,}223 Gemma-2 2B signals took approximately five days.

Exploratory experiments used the same hardware and required less compute than the reported experiments.

\section{Broader impacts}
\label{app:broader-impacts}

This work contributes to mechanistic interpretability of large language models. We briefly discuss its positive and negative societal implications.

\paragraph{Potential benefits.}
ACC++ helps researchers inspect how language models compute their outputs for individual prompts. This can support model auditing, debugging, and investigation of failure modes. Automatically generated signal interpretations provide natural-language descriptions that can help users inspect the traced circuits.

\paragraph{Potential risks.}
Interpretability findings could also be used to identify and modify circuits involved in model safeguards.

%% file: sections/appendix/notation.tex
\section{Notation used in the appendices}\label{app:notation}

We mostly follow the notation from \cite{franco2025pinpointing}. In the model, token embeddings are $D$-dimensional, there are $H$ attention heads in each layer, and there are $L$ layers.
We define $R = \frac{D}{H}$, which is the dimension of the spaces used for keys and queries in the attention mechanism.
We use $N$ to denote the number of tokens in a given prompt.
Superscript indices will denote (layer, head, destination token, source token); some indices will be omitted when there is no ambiguity,
and subscripts generally denote matrix entries unless otherwise stated.

\paragraph{Attention scores and weights.}
The attention mechanism operates on a set of $N$ tokens in $D$-dimensional embeddings: $X \in \mathbb{R}^{N \times D}$.
Each token $\v{x}\in \mathbb{R}^D$ is passed through linear transforms given by $\v{x}^\top W_K$, $\v{x}^\top W_Q$,
using weight matrices $W_K, W_Q \in \mathbb{R}^{D \times R}$.
Then the inner product is taken for all pairs of transformed tokens to yield attention scores
\begin{equation}
    A'_{ds} = \v{x}^{d\top} \Omega \v{x}^s,
    \label{eqn:attn-score-notation}
\end{equation}
in which $\Omega = W_Q W_K^\top$, $\v{x}^d$ is the destination token, and $\v{x}^s$ is the source token of the attention computation.
To enforce masked self-attention, $A'_{ds}$ is set to $-\infty$ for $d < s$.
Attention scores are then normalized for each destination $d$, yielding attention weights
\begin{equation}
    A_d = \softmax(A'_d/\sqrt{R}),
    \qquad
    A_{ds} = [A_d]_s.
    \label{eqn:attn-weight-notation}
\end{equation}

\paragraph{Layer/head indexing convention in this appendix.}
Throughout the appendix sections that follow, we fix an attention head $(\ell,a)$ and analyze its QK circuit at a fixed layer input.
Accordingly, we write $\v{x}^d$ to mean ``the input residual embedding at token $d$ at the input to layer $\ell$".

\paragraph{SVD of the QK matrix.}
Our methods make use of the SVD of $\Omega$.
The matrix $\Omega$ has size $D \times D$, but due to its construction it has maximum rank $R$.
We therefore work with the SVD of $\Omega = U\Sigma V^\top$ in which
$U \in \mathbb{R}^{D\times R}$, $V \in \mathbb{R}^{D\times R}$ and $\Sigma \in \mathbb{R}^{R\times R}$.
$U$ and $V$ are orthonormal matrices with $U^\top U = I$ and $V^\top V = I$, and
$\Sigma = \operatorname{diag}(\sigma_1, \sigma_2, \dots, \sigma_{R})$ with $\sigma_1 \geq \sigma_2 \geq \dots \geq \sigma_{R} \geq 0$.
Important to our work is that the SVD of $\Omega$ can equivalently be written as
\begin{equation}
    \Omega = \sum_{k=1}^{R} \v{u}^k\sigma^k\v{v}^{k\top}.
    \label{eqn:omega-svd-notation}
\end{equation}

\paragraph{Residual decomposition into upstream component outputs.}
We will use the standard view that the residual stream at the input to a layer is a linear sum of outputs of upstream components \cite{Elhage21}.
Let $\mathcal{C}(\ell)$ denote the set of all components upstream of the start of layer $\ell$ (including the input embedding).
For each token position $d$, let $\v{o}_c^{\,d}\in\mathbb{R}^D$ denote the output written into token $d$ by component $c\in\mathcal{C}(\ell)$.
Then
\begin{equation}
    \v{x}^d = \sum_{c\in\mathcal{C}(\ell)} \v{o}_c^{\,d}.
    \label{eqn:residual-sum-notation}
\end{equation}
(Here the subscript $c$ indexes components; it is not a matrix component.)

%% file: sections/appendix/unified_bilinear_form.tex
\section{A unified bilinear form for bias and rotary positional embeddings}\label{app:unified-bilinear}

As discussed in Section~\ref{sec:improv-acc}, ACC++ makes a number of methodological improvements to ACC. To deal with bias in the attention computation, ACC used homogeneous coordinates (vectors with an additional component equal to 1), and to deal with rotary positional embeddings (RoPE), ACC used position-specific $\Omega$ matrices. In ACC++, we express the attention computation in all these cases (whether involving bias, RoPE, both, or neither) using a single $\Omega$ matrix per head, which means that ACC++ only requires one SVD operation per head.

\paragraph{Core assumption.} We assume that the query and key weight matrices, $W_Q \in \mathbb{R}^{D \times R}$ and $W_K \in \mathbb{R}^{D \times R}$ (where $D > R$), are \textbf{well-conditioned} (and thus have full column rank).  For the models studied in this work, this assumption is \textbf{supported empirically:} Appendix~\ref{app:condit-number} shows that both $W_Q$ and $W_K^{\top}$ are well-conditioned, in all cases having condition numbers below $1000$, and usually much lower. For a full column rank matrix $W$ with Moore-Penrose pseudoinverse $W^\dagger$, $W^\dagger W = I_R$. When $W$ is well-conditioned, this identity is also numerically accurate. Our derivations use the following identities:
$$
W_Q^\dagger W_Q = I_R \quad \text{and} \quad W_K^\top (W_K^\top)^\dagger = I_R
$$

\subsection{Models with bias (e.g., GPT-2)} \label{app:unique-bilinear-gpt}
\label{app:bias-technique}
Let $\v{x}^d, \v{x}^s \in \mathbb{R}^{D \times 1}; W_Q, W_K \in \mathbb{R}^{D \times R}; \v{b}_Q, \v{b}_K \in \mathbb{R}^{R \times 1}$. The attention score for a model with bias is given below. To factor out $W_Q$ and $W_K$, we use the identities $W_Q^\dagger W_Q = I_R$ and $W_K^\dagger W_K = I_R$, which allow us to write $\v{b}_Q^\top = \v{b}_Q^\top W_Q^\dagger W_Q$ and $\v{b}_K^\top = \v{b}_K^\top W_K^\dagger W_K$:

\begin{align*}
A'_{ds} &= \left ( {\v{x}^d}^{\top} W_Q + \v{b}_Q^{\top} \right) \left ( {\v{x}^s}^{\top} W_K + \v{b}_K^{\top} \right)^{\top} \\
& = \left [ \left({\v{x}^d}^{\top} + \v{b}_Q^{\top} W_{Q}^{\dagger} \right) W_Q \right ] \left [ \left({\v{x}^s}^{\top} + \v{b}_K^{\top} W_{K}^{\dagger} \right) W_K \right ]^{\top} \\
& = \left({\v{x}^d}^{\top} + \v{b}_Q^{\top} W_{Q}^{\dagger} \right) W_Q W_K^{\top} \left({\v{x}^s} + {W_{K}^{\dagger}}^{\top} \v{b}_K \right) \\
& = \left({\v{x}^d}^{\top} + {\v{c}^d}^{\top} \right) \Omega \left({\v{x}^s} + \v{c}^s \right)
\end{align*}

The final expression is a bilinear form with additive bias terms ${\v{c}^d}^{\top}=\v{b}_{Q}^{\top}W_{Q}^{\dagger}$ and $\v{c}^s={W_{K}^{\dagger}}^{\top} \v{b}_K$. 

\subsection{Models with RoPE (e.g., Gemma-2)}


Given the RoPE rotation matrices $\mathcal{R}^d, {\mathcal{R}^s} \in \mathbb{R}^{R \times R}$, the attention score for token pair $(d, s)$ in RoPE models is given by: 
\begin{align*}
A'_{ds} &= \left( {\v{x}^d}^{\top} W_Q {\mathcal{R}^d} \right) \left( {\v{x}^s}^{\top} W_K \mathcal{R}^s \right)^{\top} \\
& = {\v{x}^d}^{\top} W_Q {\mathcal{R}^d} {\mathcal{R}^s}^{\top} W_K^{\top} \v{x}^s \\
& = {\v{x}^d}^{\top} W_Q \mathcal{R}^{(d-s)} W_K^{\top} \v{x}^s \\
& = {\v{x}^d}^{\top} \Omega^{(d-s)} \v{x}^s
\end{align*}

Note that $\Omega$ is now a position-dependent matrix $\Omega^{(d-s)}$. To rewrite the attention score using a fixed $\Omega$, we choose matrices $M_d, M_s \in \mathbb{R}^{D \times D}$ that transform the input vectors such that $A'_{ds} = ({\v{x}^d}^{\top} M_d) W_Q W_K^{\top} (M_s \v{x}^s)$. Let $M_d = W_Q \mathcal{R}^d W_Q^{\dagger}$ and $M_s = ({W_K^{\top}})^{\dagger} {\mathcal{R}^s}^{\top} W_K^{\top}$. Using the identities $W_Q^\dagger W_Q = I_R$ and $W_K^\top (W_K^\top)^\dagger = I_R$, we verify this formulation:

\begin{align*}
A'_{ds} &= ({\v{x}^d}^{\top} M_d)\Omega (M_s \v{x}^s) \\
&= ({\v{x}^d}^{\top} M_d) W_Q W_K^{\top} (M_s \v{x}^s) \\
& = {\v{x}^d}^{\top} W_Q \mathcal{R}^d (W_Q^{\dagger} W_Q) (W_K^{\top} ({W_K^{\top}})^{\dagger}) {\mathcal{R}^s}^{\top} W_K^{\top} \v{x}^s \\
& = {\v{x}^d}^{\top} W_Q \mathcal{R}^d {\mathcal{R}^s}^{\top} W_K^{\top} \v{x}^s \\
& = {\v{x}^d}^{\top} W_Q \mathcal{R}^{(d-s)} W_K^{\top} \v{x}^s \\
& = {\v{x}^d}^{\top} \Omega^{(d-s)} \v{x}^s
\end{align*}

Therefore, we can compute attention scores with RoPE by transforming the input vectors with $M_d$ and $M_s$, while keeping $\Omega$ fixed across token positions. This allows us to compute the SVD of $\Omega$ only once per head.

\subsection{Models with bias and RoPE (e.g., Pythia)}

We now combine the two previous derivations. First, the attention score for a token pair $(d, s)$ in a RoPE model with bias is given by:

\begin{align*}
A'_{ds} &= \left [ \left ( {\v{x}^d}^{\top} W_Q + \v{b}_Q^{\top} \right)\mathcal{R}^d \right] \left [ \left ( {\v{x}^s}^{\top} W_K + \v{b}_K^{\top} \right) {\mathcal{R}^s} \right]^{\top} \\
& = \left[ \left ( {\v{x}^d}^{\top} W_Q + \v{b}_Q^{\top} \right) \mathcal{R}^d \right] \left[ {\mathcal{R}^s}^{\top} \left ( W_K^{\top} {\v{x}^s} + \v{b}_K \right) \right] \\
& = \left ( {\v{x}^d}^{\top} W_Q \mathcal{R}^d + \v{b}_Q^{\top} \mathcal{R}^d \right) \left ( {\mathcal{R}^s}^{\top} W_K^{\top} {\v{x}^s} + {\mathcal{R}^s}^{\top} \v{b}_K \right) \\
& = \left( {\v{x}^d}^{\top} W_Q \mathcal{R}^d {\mathcal{R}^s}^{\top} W_K^{\top} \v{x}^s \right)
  + \left( {\v{x}^d}^{\top} W_Q \mathcal{R}^d {\mathcal{R}^s}^{\top} \v{b}_K \right) \\
& \quad + \left( \v{b}_Q^{\top} \mathcal{R}^d {\mathcal{R}^s}^{\top} W_K^{\top} \v{x}^s \right)
  + \left( \v{b}_Q^{\top} \mathcal{R}^d {\mathcal{R}^s}^{\top} \v{b}_K \right)
\end{align*}

Our goal is to show that this expression can be written as:
$$
A'_{ds} = \left[ \left({\v{x}^d}^{\top} + {\v{c}^d}^{\top} \right) M_d \right] W_Q W_K^{\top} \left[M_s \left({\v{x}^s} + \v{c}^s \right) \right]
$$
where $M_d$, $M_s$, $\v{c}^d$, and $\v{c}^s$ are defined as in the previous sections. We verify this by expanding its query and key terms.

First, we expand the query term using $W_Q^\dagger W_Q = I_R$:
\begin{align*}
\left[ \left({\v{x}^d}^{\top} + {\v{c}^d}^{\top} \right) M_d \right] W_Q & = \left({\v{x}^d}^{\top} + \v{b}_{Q}^{\top}W_{Q}^{\dagger} \right) (W_Q \mathcal{R}^d W_Q^{\dagger}) W_Q \\
& = \left({\v{x}^d}^{\top} + \v{b}_{Q}^{\top}W_{Q}^{\dagger} \right) W_Q \mathcal{R}^d (W_Q^{\dagger} W_Q) \\
& = \left({\v{x}^d}^{\top}W_Q + \v{b}_{Q}^{\top}W_{Q}^{\dagger} W_Q \right) \mathcal{R}^d \\
& = \left({\v{x}^d}^{\top} W_Q + \v{b}_{Q}^{\top} \right) \mathcal{R}^d \\
& = {\v{x}^d}^{\top} W_Q \mathcal{R}^d + \v{b}_{Q}^{\top} \mathcal{R}^d
\end{align*}

Second, we expand the key term using $W_K^\top (W_K^\top)^\dagger = I_R$ and $(W_K^\dagger)^\top = (W_K^\top)^\dagger$:
\begin{align*}
W_K^{\top} \left[M_s \left({\v{x}^s} + \v{c}^s \right) \right] & = W_K^{\top} M_s {\v{x}^s} + W_K^{\top} M_s \v{c}^s \\
& = W_K^{\top} ({W_K^{\top}})^{\dagger} {\mathcal{R}^s}^{\top} W_K^{\top} {\v{x}^s} + W_K^{\top} ({W_K^{\top}})^{\dagger} {\mathcal{R}^s}^{\top} W_K^{\top} {W_{K}^{\dagger}}^{\top} \v{b}_K \\
& = (I_R) {\mathcal{R}^s}^{\top} W_K^{\top} {\v{x}^s} + (I_R) {\mathcal{R}^s}^{\top} (I_R) \v{b}_K \\
& = {\mathcal{R}^s}^{\top} W_K^{\top} {\v{x}^s} + {\mathcal{R}^s}^{\top} \v{b}_K
\end{align*}

Combining the left and right parts, we have:
\begin{align*}
A'_{ds} & = \left({\v{x}^d}^{\top} W_Q \mathcal{R}^d + \v{b}_{Q}^{\top} \mathcal{R}^d \right) \left ( {\mathcal{R}^s}^{\top} W_K^{\top} {\v{x}^s} + {\mathcal{R}^s}^{\top} \v{b}_K \right) \\
& = \left( {\v{x}^d}^{\top} W_Q \mathcal{R}^d {\mathcal{R}^s}^{\top} W_K^{\top} \v{x}^s \right)
  + \left( {\v{x}^d}^{\top} W_Q \mathcal{R}^d {\mathcal{R}^s}^{\top} \v{b}_K \right) \\
& \quad + \left( \v{b}_Q^{\top} \mathcal{R}^d {\mathcal{R}^s}^{\top} W_K^{\top} \v{x}^s \right)
  + \left( \v{b}_Q^{\top} \mathcal{R}^d {\mathcal{R}^s}^{\top} \v{b}_K \right)
\end{align*}

Therefore, the attention score for a token pair $(d, s)$ in a RoPE model with bias can be written as
\[
A'_{ds} = \left[ \left({\v{x}^d}^{\top} + {\v{c}^d}^{\top} \right) M_d \right] \Omega \left[M_s \left({\v{x}^s} + \v{c}^s \right) \right],
\]
where $\Omega$ is fixed across token positions.

\subsection{Checking the condition number of the $W_Q$ and $W_K^{\top}$ matrices}\label{app:condit-number}

\begin{figure}[ht]
    \centering
    \includegraphics[width=0.45\textwidth]{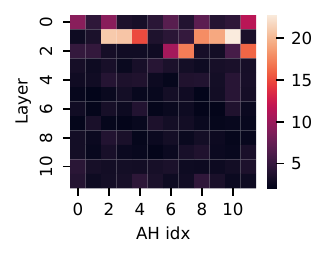}
    \includegraphics[width=0.45\textwidth]{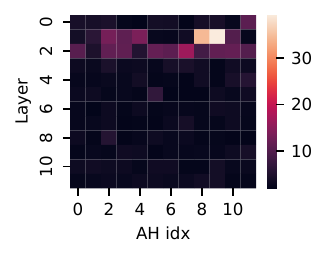}
    \caption{Condition numbers of $W_Q$ (left) and $W_K^{\top}$ (right) from GPT-2 small.}
    \label{fig:cond-number-gpt-2-small}
\end{figure}

\begin{figure}[ht]
    \centering
    \includegraphics[width=0.45\textwidth]{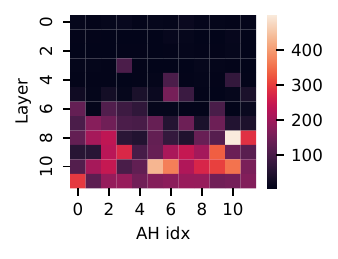}
    \includegraphics[width=0.45\textwidth]{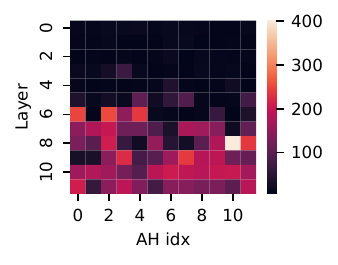}
    \caption{Condition numbers of $W_Q$ (left) and $W_K^{\top}$ (right) from Pythia-160M.}
    \label{fig:cond-number-pythia-160m}
\end{figure}

\begin{figure}[ht]
    \centering
    \includegraphics[width=0.45\textwidth]{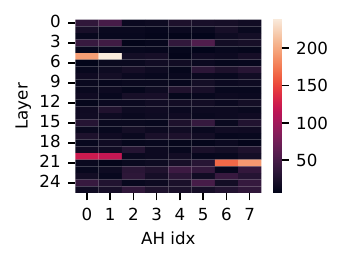}
    \includegraphics[width=0.45\textwidth]{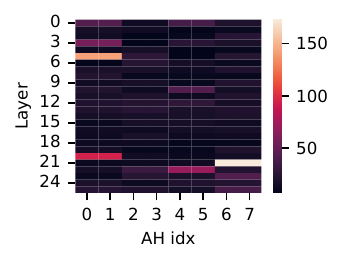}
    \caption{Condition numbers of $W_Q$ (left) and $W_K^{\top}$ (right) from Gemma-2 2B.}
    \label{fig:cond-number-gemma-2-2b}
\end{figure}

Figures~\ref{fig:cond-number-gpt-2-small}, \ref{fig:cond-number-pythia-160m}, and \ref{fig:cond-number-gemma-2-2b} report the condition numbers of $W_Q$ (left) and $W_K^{\top}$ (right) for the models used in this work. In all cases, the matrices are well-conditioned, with condition number $< 1000$, and usually much less.

%% file: sections/appendix/new_acc.tex
\section{ACC++: refined ACC as a counterfactual search} \label{app:new-acc}

This appendix provides the mathematical description of our refined attention-causal communication (ACC++) procedure.
We assume throughout that we are analyzing a fixed attention head $(\ell,a)$ so that $\v{x}^d$ denotes the residual at token $d$ at the input to layer $\ell$.

\subsection{Problem definition}\label{app:new-acc:def}

Fix a destination--source pair $(d,s)$ such that $A_{ds}$ is high. As in \citep{franco2025pinpointing}, we distinguish between \emph{destination signals} (signals in the destination token residual $\v{x}^d$) and \emph{source signals} (signals in the source token residuals $\{\v{x}^j\}_{j\leq d}$; these affect $A_{ds}$ through softmax competition).

A key improvement of ACC++ over ACC is that the counterfactual objective is defined directly in terms of the \emph{post-softmax} attention weight $A_{ds}$, rather than via a linear surrogate of attention. For simplicity, this appendix considers models without QK bias or RoPE. Appendix~\ref{app:unified-bilinear} shows how to express attention scores for models with bias and/or RoPE using transformed input vectors and a fixed $\Omega$.

We summarize the ACC++ procedure before introducing the formal definitions.

\begin{enumerate}
    \item \textbf{Set up the counterfactual.} Choose an attention head and a destination--source pair, then decide whether we are searching for \emph{destination} signals (in the destination token) or \emph{source} signals (distributed across all source tokens that compete under softmax).
    \item \textbf{Enumerate candidate signals.} Decompose the residual stream into outputs of upstream components, and project each component's contribution onto the head's singular-vector directions to form a set of candidate signals.
    \item \textbf{Build a contribution matrix.} Convert candidates into a contribution matrix whose rows correspond to candidate signals and whose columns correspond to source positions in the destination row of attention scores.
    \item \textbf{Score candidates with attribution.} Use Integrated Gradients to assign each candidate a fixed importance score for the attention weight on the chosen source token, while accounting for softmax competition across all sources.
    \item \textbf{Solve the counterfactual by greedy removal.} Starting from the full set of candidates, iteratively remove the highest-scoring candidates and recompute the attention weight until it drops below a chosen threshold; the removed candidates form the ACC++ explanation set.
\end{enumerate}

\paragraph{Destination intervention (find destination signals).}
Let $M_{\mathrm{dst}}$ denote a set of destination-signal candidates. We intervene on the destination side while keeping all source-side residuals fixed, including at position $d$:
\begin{equation}
\mathrm{do}_{\mathrm{dst}}(M_{\mathrm{dst}}):\qquad
\tv{x}^d = \v{x}^d - \Delta^d(M_{\mathrm{dst}})
\label{eqn:do-dst}
\end{equation}
and define the induced attention scores and weights by
\begin{equation}
A'_{dj}\big(\mathrm{do}_{\mathrm{dst}}(M_{\mathrm{dst}})\big)
= \tv{x}^{d\top}\Omega \v{x}^j,
\qquad
A_d\big(\mathrm{do}_{\mathrm{dst}}(M_{\mathrm{dst}})\big)
= \softmax\!\Big(A'_d\big(\mathrm{do}_{\mathrm{dst}}(M_{\mathrm{dst}})\big)/\sqrt{R}\Big).
\label{eqn:dst-induced}
\end{equation}

\noindent Here $\Delta^d(M_{\mathrm{dst}})$ is the total vector removed from token $d$; see Eq.~\eqref{eqn:Delta-dst}. The destination ACC++ objective is to find a small $M_{\mathrm{dst}}$ such that
\begin{equation}
\Big[A_d\big(\mathrm{do}_{\mathrm{dst}}(M_{\mathrm{dst}})\big)\Big]_s < \tau.
\label{eqn:acc-dst}
\end{equation}

Appendix~\ref{app:finding-tau} describes how we choose $\tau$.

\paragraph{Source intervention (find source signals).}
Let $M_{\mathrm{src}}$ denote a set of source-signal candidates. We intervene on the source-side residuals (for all $j\le d$) while keeping the residual on the destination side fixed:
\begin{equation}
\mathrm{do}_{\mathrm{src}}(M_{\mathrm{src}}):\qquad \tv{x}^j = \v{x}^j - \Delta^j(M_{\mathrm{src}})\ \text{ for all } j\le d,
\label{eqn:do-src}
\end{equation}
and define the induced attention scores and weights by
\begin{equation}
A'_{dj}\big(\mathrm{do}_{\mathrm{src}}(M_{\mathrm{src}})\big)
= \v{x}^{d\top}\Omega \tv{x}^j,
\qquad
A_d\big(\mathrm{do}_{\mathrm{src}}(M_{\mathrm{src}})\big)
= \softmax\!\Big(A'_d\big(\mathrm{do}_{\mathrm{src}}(M_{\mathrm{src}})\big)/\sqrt{R}\Big).
\label{eqn:src-induced}
\end{equation}

\noindent
Similarly, $\Delta^j(M_{\mathrm{src}})$ is the total vector removed at token $j$ (Eq.~\eqref{eqn:Delta-src}).

The source ACC++ objective is to find a small $M_{\mathrm{src}}$ such that
\begin{equation}
\Big[A_d\big(\mathrm{do}_{\mathrm{src}}(M_{\mathrm{src}})\big)\Big]_s < \tau.
\label{eqn:acc-src}
\end{equation}

Because $A_d=\softmax(A'_d/\sqrt{R})$ uses all source tokens $j\le d$, both objectives must be evaluated with respect to their effect on the full score vector $A'_d$, not only on the entry $A'_{ds}$.

\subsection{Signal candidates}\label{app:new-acc:cands}

We construct candidates by decomposing residuals into upstream component outputs and projecting onto singular vectors of $\Omega$. Recall (Appendix~\ref{app:notation}) that for each token position $j$,
\begin{equation}
\v{x}^j \;=\; \sum_{c\in\mathcal{C}(\ell)} \v{o}_c^{\,j}.
\label{eqn:residual-sum-repeat}
\end{equation}

\paragraph{Destination candidates.}
Using the left singular vectors $\{\v{u}^k\}_{k=1}^R$ of $\Omega$ (Eq.~\eqref{eqn:omega-svd-notation}),
define the rank-1 projector
\begin{equation}
P_U^k \;=\; \v{u}^k\v{u}^{k\top}.
\label{eqn:PU}
\end{equation}
A destination-side candidate for component $c$ and direction $k$ is
\begin{equation}
\v{s}_{c}^{\,dk} \;=\; P_U^k \v{o}_c^{\,d}\in\mathbb{R}^D.
\label{eqn:dest-cand}
\end{equation}

Thus, $\sum_c\sum_k\v{s}_c^{\,dk}=UU^\top\v{x}^d$. This projection preserves the attention scores because $(UU^\top)\Omega=\Omega$.

\paragraph{Source candidates.}
Using the right singular vectors $\{\v{v}^k\}_{k=1}^R$ of $\Omega$ (Eq.~\eqref{eqn:omega-svd-notation}),
define
\begin{equation}
P_V^k \;=\; \v{v}^k\v{v}^{k\top}.
\label{eqn:PV}
\end{equation}
For each source position $j\le d$, a source-side candidate is
\begin{equation}
\v{s}_{c}^{\,jk} \;=\; P_V^k \v{o}_c^{\,j}\in\mathbb{R}^D.
\label{eqn:source-cand}
\end{equation}

Thus, $\sum_c\sum_k\v{s}_c^{\,jk}=VV^\top\v{x}^j$ for each $j\le d$. This projection preserves the attention scores because $\Omega(VV^\top)=\Omega$.

\paragraph{Intervention vectors.}
Given a set of candidate indices $M\subseteq \mathcal{C}(\ell)\times\{1,\dots,R\}$, we define the intervention vector at token position $j$ as the sum of the corresponding candidate signal vectors. For the destination side,

\begin{equation}
\Delta^d(M_{\mathrm{dst}})
\;=\;
\sum_{(c,k)\in M_{\mathrm{dst}}} \v{s}_{c}^{\,dk}
\;=\;
\sum_{(c,k)\in M_{\mathrm{dst}}} P_U^k\,\v{o}_c^{\,d}.
\label{eqn:Delta-dst}
\end{equation}
For the source side, the same candidate set $M_{\mathrm{src}}$ is applied across all source positions $j\le d$:
\begin{equation}
\Delta^j(M_{\mathrm{src}})
\;=\;
\sum_{(c,k)\in M_{\mathrm{src}}} \v{s}_{c}^{\,jk}
\;=\;
\sum_{(c,k)\in M_{\mathrm{src}}} P_V^k\,\v{o}_c^{\,j},
\qquad \forall j\le d.
\label{eqn:Delta-src}
\end{equation}

\subsection{Contribution matrix (destination or source view)}\label{app:new-acc:C}

Our method constructs a contribution matrix whose rows correspond to signal candidates and whose columns correspond to source token positions $j\le d$. Let $Q =|\mathcal{C}(\ell)| \cdot R$ be the total number of signal candidates.

\paragraph{Destination view.}
Using $\sum_{c,k}\v{s}_c^{\,dk}=UU^\top\v{x}^d$ and $(UU^\top)\Omega=\Omega$, we obtain
\begin{equation}
A'_{dj} \;=\; \sum_{c}\sum_{k} \big(\v{s}_{c}^{\,dk}\big)^\top \Omega \v{x}^j.
\end{equation}
We form a matrix $C\in\mathbb{R}^{Q \times d}$ with one row per pair $(c,k)$ and define
\begin{equation}
C_{ij} \;=\; \big(\v{s}_{c}^{\,dk}\big)^\top \Omega \v{x}^j,
\qquad j\le d,
\label{eqn:C-dest}
\end{equation}
where row $i$ corresponds to candidate $(c,k)$.

\paragraph{Source view.}
Using $\sum_{c,k}\v{s}_c^{\,jk}=VV^\top\v{x}^j$ and $\Omega(VV^\top)=\Omega$, we obtain, for each $j\le d$,
\begin{equation}
A'_{dj}
\;=\;
\sum_{c}\sum_{k}\v{x}^{d\top}\Omega\,\v{s}_{c}^{\,jk}.
\label{eqn:score-decomp-source}
\end{equation}

We form a contribution matrix $C\in\mathbb{R}^{Q \times d}$ with one row per pair $(c,k)$, and define
\begin{equation}
C_{ij}
\;=\;
\v{x}^{d\top}\Omega\,\v{s}_{c}^{\,jk},
\qquad j\le d,
\label{eqn:C-source}
\end{equation}
where row $i$ corresponds to candidate $(c,k)$. Equivalently, row $i$ is the vector $\v{c}_i\in\mathbb{R}^d$ whose $j$-th entry is the contribution of $(c,k)$ to the score $A'_{dj}$. With this definition, removing the row corresponding to $(c,k)$ is the same as removing what component $c$ writes in direction $\v{v}^k$ \emph{in every source token} $j\le d$.

\paragraph{From the contribution matrix to attention scores.}
In either case, the candidate construction implies an exact decomposition of the full attention score vector:
\begin{equation}
A'_d \;=\; \sum_{i=1}^{Q}\v{c}_i,
\qquad
A_d \;=\; \softmax(A'_d/\sqrt{R}).
\label{eqn:Ad-sumrows}
\end{equation}

\subsection{Attribution via integrated gradients}\label{app:new-acc:IG}

We seek to attribute the impact of each candidate on the single attention weight $A_{ds}=[A_d]_s$, accounting for the nonlinear effects of softmax across all source positions. To do this, we define the path

\begin{equation}
\v{z}(t) \;=\; t\,A'_d/\sqrt{R},\qquad
\v{p}(t) \;=\; \softmax(\v{z}(t)),
\qquad t\in[0,1],
\label{eqn:ig-path}
\end{equation}

so that $\v{p}(1)=A_d$ and $A_{ds} = p_s(1)$.
Let $g(\v{z})=[\softmax(\v{z})]_s$ denote the scalar output of interest.
Integrated gradients \cite{sundararajan2017axiomatic} assigns candidate $i$ the attribution

\begin{equation}
\operatorname{IG}_i
\;=\;
\frac{1}{\sqrt{R}}\int_0^1
\left\langle \nabla g(\v{z}(t)),\; \v{c}_i \right\rangle dt.
\label{eqn:IG-def}
\end{equation}

Using the identity

\begin{equation}
\frac{\partial\, p_s}{\partial z_j}
\;=\;
p_s\,(\mathbf{1}\{j=s\}-p_j),
\label{eqn:softmax-grad}
\end{equation}

we obtain an explicit form that matches our implementation:

\begin{equation}
\operatorname{IG}_i
\;=\;
\frac{1}{\sqrt{R}}\int_0^1
p_s(t)\left(
C_{is} \;-\; \sum_{j\le d} C_{ij}\,p_j(t)
\right) dt
\;=\;
\frac{1}{\sqrt{R}}\int_0^1
p_s(t)\Big(C_{is}-\langle \v{c}_i,\v{p}(t)\rangle\Big)\,dt.
\label{eqn:IG-closed}
\end{equation}

This formulation makes the role of softmax competition explicit: the term $\langle \v{c}_i,\v{p}(t)\rangle$ accounts for candidate $i$'s contributions to all source positions at each point along the path. In practice, we approximate the integral in Eq.~\eqref{eqn:IG-closed} using a trapezoidal rule with $T=64$ steps.

\subsection{Greedy solver}\label{app:new-acc:greedy}

We solve the destination objective (Eq.~\eqref{eqn:acc-dst}) and the source objective (Eq.~\eqref{eqn:acc-src}) using the same greedy algorithm. We compute IG attributions $\{\operatorname{IG}_i\}_{i=1}^{Q}$ \emph{once}, using the original score vector $A'_d=\sum_{i=1}^{Q}\v{c}_i$, and treat these attributions as fixed throughout the greedy procedure. Let $S\subseteq\{1,\dots,Q\}$ denote the set of active candidates (initialized to all candidates), and define the partial score vector

\begin{equation}
A'_d(S) \;=\; \sum_{i\in S}\v{c}_i,
\qquad
A_{ds}(S) \;=\; \Big[\softmax\!\big(A'_d(S)/\sqrt{R}\big)\Big]_s.
\label{eqn:AdS}
\end{equation}

While $A_{ds}(S)\geq\tau$, remove the remaining candidate with the highest attribution according to the \emph{fixed} scores $\operatorname{IG}_i$:
\begin{equation}
i^\star \;=\; \argmax_{i\in S}\operatorname{IG}_i,
\qquad
S \leftarrow S\setminus\{i^\star\}.
\label{eqn:remove}
\end{equation}

Let $M$ denote the set of candidate pairs $(c,k)$ corresponding to the removed row indices $\{1,\dots,Q\}\setminus S$. We write $M_{\mathrm{dst}}$ or $M_{\mathrm{src}}$ depending on whether we are solving the destination or source objective.

\subsection{ACC++ high-level recipe}\label{app:new-acc:recipe}

\begin{figure}[h]
\centering
\resizebox{\textwidth}{!}{%
\begin{tikzpicture}[
  node distance=9mm,
  box/.style={draw, rounded corners, align=center, inner sep=5pt},
  arr/.style={-{Stealth[length=2.2mm]}, thick},
]
\node[box] (start) {Fix head $(\ell,a)$\\ and pair $(d,s)$};
\node[box, right=of start] (choose) {Choose view:\\ destination ($\v{x}^d$)\\ or sources ($\{\v{x}^j\}_{j\le d}$)};
\node[box, right=of choose] (svd) {Compute $\Omega$ and SVD\\ $\Omega=\sum_{k=1}^{R}\v{u}^k\sigma^k\v{v}^{k\top}$};

\node[box, right=of svd] (proj) {Form candidates via projections\\
dst: $\v{s}_{c}^{\,dk}=P_U^k\,\v{o}_c^{\,d}$\\
src: $\v{s}_{c}^{\,jk}=P_V^k\,\v{o}_c^{\,j}\ (\forall j\le d)$};
\node[box, below=of proj] (C) {Build contribution matrix $C$\\ over $j\le d$\\ (Eqs.~\eqref{eqn:C-dest},\eqref{eqn:C-source})};
\node[box, left=of C] (ig) {IG attribution: $\operatorname{IG}_i$\\
to $A_{ds}=[\softmax(A'_d/\sqrt{R})]_s$\\
along $\v{z}(t)=tA'_d/\sqrt{R}$};
\node[box, left=of ig] (greedy) {Greedily remove\\
$\argmax_{i\in S}\operatorname{IG}_i$ (fixed scores),\\
recompute $A_{ds}$ until $A_{ds}<\tau$};
\node[box, left=of greedy] (out) {Output $M_{\mathrm{dst}}$ or $M_{\mathrm{src}}$\\};

\draw[arr] (start) -- (choose);
\draw[arr] (choose) -- (svd);
\draw[arr] (svd) -- (proj);
\draw[arr] (proj) -- (C);
\draw[arr] (C) -- (ig);
\draw[arr] (ig) -- (greedy);
\draw[arr] (greedy) -- (out);
\end{tikzpicture}
}
\caption{High-level recipe for the ACC++ solver, distinguishing destination signal candidates and source signal candidates.}
\label{fig:acc-recipe-new-v3}
\end{figure}

%% file: sections/appendix/finding_tau.tex
\section{Finding $\tau$}\label{app:finding-tau}

ACC++ uses a threshold $\tau$ to define high attention weights $A_{ds}$. Because each row in the attention matrix $A$ is a probability distribution over the available source tokens, the average value of an entry depends on the row context size $d$. To make thresholds comparable across different $d$, models, and datasets, we study the scaled score $d\cdot A_{ds}$, where $d$ is the number of available source tokens, indexed by $j=1,\dots,d$.

\begin{figure}[ht]
    \centering
    \includegraphics[width=0.3\textwidth]{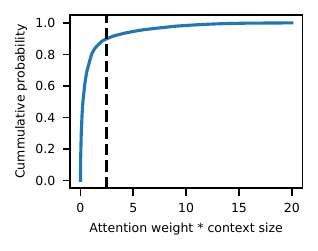}
    \includegraphics[width=0.3\textwidth]{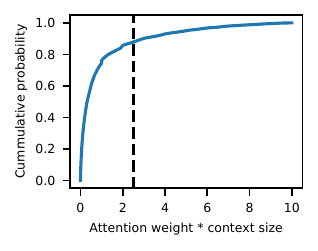}
    \includegraphics[width=0.3\textwidth]{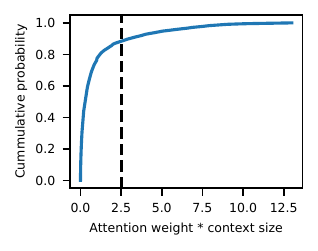}
    \caption{ECDF of the attention weight times the context size for GPT-2 small in IOI (a), GP (b), and GT (c). The vertical line marks $dA_{ds}=2.5$, corresponding to $\tau(d)=2.5/d$.}
    \label{fig:tau-gpt-2-small}
\end{figure}

\begin{figure}[ht]
    \centering
    \includegraphics[width=0.3\textwidth]{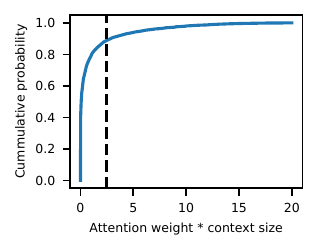}
    \includegraphics[width=0.3\textwidth]{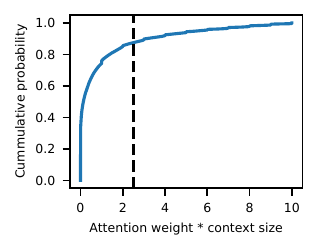}
    \includegraphics[width=0.3\textwidth]{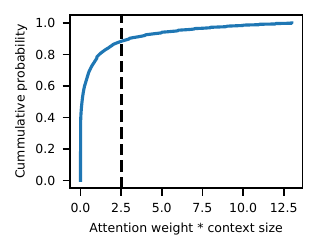}
    \caption{ECDF of the attention weight times the context size for Pythia-160M in IOI (a), GP (b), and GT (c). The vertical line marks $dA_{ds}=2.5$, corresponding to $\tau(d)=2.5/d$.}
    \label{fig:tau-pythia-160m}
\end{figure}

\begin{figure}[ht]
    \centering
    \includegraphics[width=0.3\textwidth]{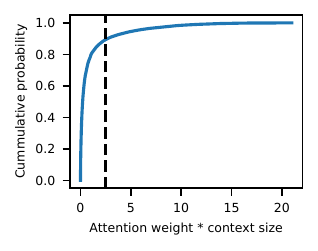}
    \includegraphics[width=0.3\textwidth]{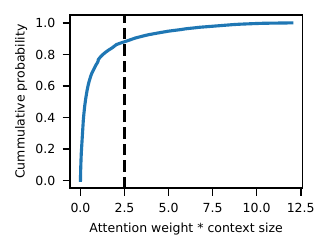}
    \caption{ECDF of the attention weight times the context size for Gemma-2 2B in IOI (a) and GP (b). The vertical line marks $dA_{ds}=2.5$, corresponding to $\tau(d)=2.5/d$.}
    \label{fig:tau-gemma-2-2b}
\end{figure}

For each model--dataset setting, we run a forward pass over the evaluation set, collect all attention weights $A_{ds}$ across layers/heads/tokens, and plot the empirical cumulative distribution function (ECDF) of $d\cdot A_{ds}$. Across GPT-2 small (Figure~\ref{fig:tau-gpt-2-small}), Pythia-160M (Figure~\ref{fig:tau-pythia-160m}), and Gemma-2 2B (Figure~\ref{fig:tau-gemma-2-2b}) on the datasets used, the ECDFs exhibit a consistent knee in the range $1.5 \leq d\cdot A_{ds} \leq 2.5$.

Based on this observation, we use a \emph{dynamic} threshold that adapts to the row context size:
\[
\tau(d)=\frac{k}{d},
\]

where $1.5 \leq k \leq 2.5$.

\begin{figure}[ht]
    \centering
    \includegraphics[width=0.3\textwidth]{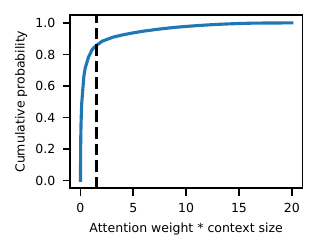}
    \includegraphics[width=0.3\textwidth]{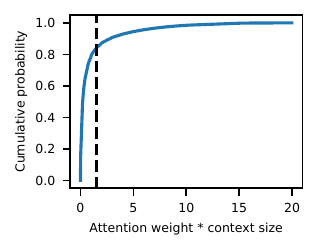}
    \caption{ECDF of the attention weight times the context size for mGPT (left) and Gemma-2 2B (right) on Multilingual IOI. The vertical line marks $dA_{ds}=1.5$, corresponding to $\tau(d)=1.5/d$.}
    \label{fig:tau-mioi}
\end{figure}

We choose a single value of $k$ for each task for simplicity. However, the impact of this strategy is unclear in other scenarios. We leave further study of better methods to set a threshold for ACC++ as future work. Lower values of $k$ produce larger graphs, while higher values produce smaller graphs. In the experiments presented in \S~\ref{sec:uses}, we use $k = 2.5$ for the IOI task, and $k = 1.5$ for the mIOI task (see Figure~\ref{fig:tau-mioi} for the ECDFs for Gemma-2 2B and mGPT).

%% file: sections/appendix/prev_acc_results.tex
\section{Reproducing prior ACC results with ACC++} \label{app:reproducing-previous-results}


We repeat the circuit tracing and intervention experiments of \cite{franco2025pinpointing} using ACC++, with the same datasets and evaluation protocol. We evaluate GPT-2 Small and Pythia-160M on IOI, GT, and GP, and Gemma-2 2B on IOI and GP. This experimental setup differs from the IOI setup used in the main body of this paper.

\begin{figure}[ht]
    \centering
    \includegraphics[scale=0.75]{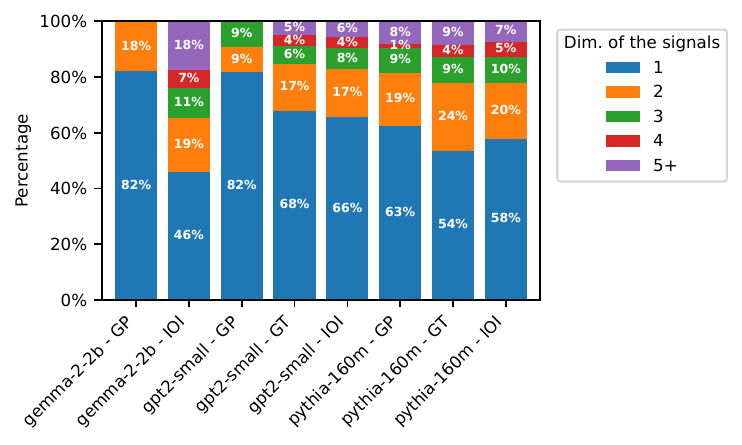}
    \caption{Distribution of ACC++ signal dimensionality, measured by the number of singular vectors used. One-dimensional signals are the most common in every model--task setting shown.}
    \label{fig:sparse-attn-decomposition}
\end{figure}

\begin{figure}[ht]
  \centering
  \begin{subfigure}[t]{0.99\linewidth}
    \centering
    \includegraphics[width=0.3\textwidth]{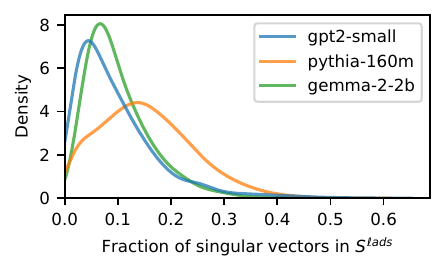}
    \includegraphics[width=0.3\textwidth]{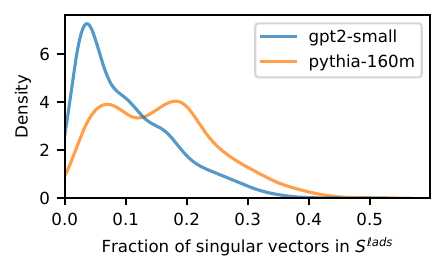}
    \includegraphics[width=0.3\textwidth]{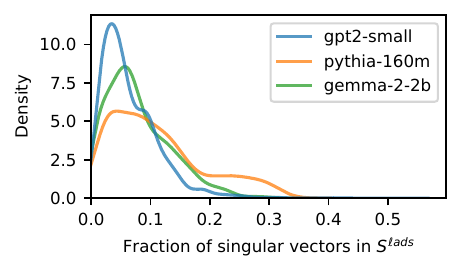}
    \caption{ACC. Source: \cite{franco2025pinpointing}.}
    \label{fig:distrib-n-svs-acc}
  \end{subfigure}
  \vspace{0.1cm}
  \begin{subfigure}[t]{0.99\linewidth}
    \centering
    \includegraphics[width=0.3\textwidth]{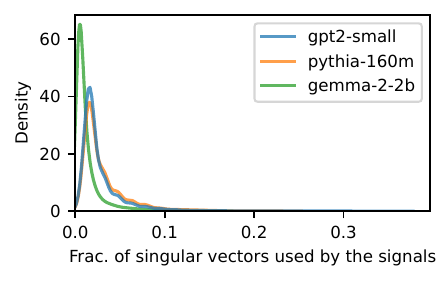}
    \includegraphics[width=0.3\textwidth]{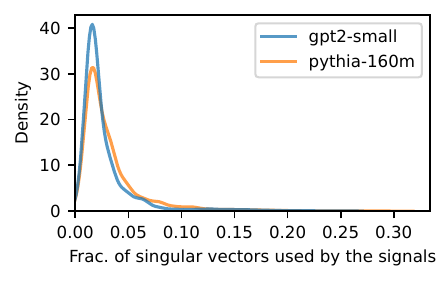}
    \includegraphics[width=0.3\textwidth]{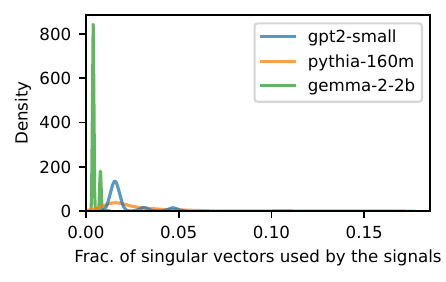}
    \caption{ACC++.}
    \label{fig:distrib-n-svs-accpp}
  \end{subfigure}
    \caption{Distributions of the fraction of singular vectors used by signals for ACC (top) and ACC++ (bottom). From left to right: IOI, GT, and GP. ACC++ selects lower-dimensional signals than ACC.}
  \label{fig:distrib-n-svs}
\end{figure}

We measure signal dimensionality by the number of singular vectors of $\Omega$ used by a signal. ACC++ selects lower-dimensional signals than ACC (Figures~\ref{fig:sparse-attn-decomposition} and~\ref{fig:distrib-n-svs}). Signals using one singular vector are the most common in every model--task setting shown.

\begin{figure}[ht]
  \centering
  \begin{subfigure}[t]{0.32\linewidth}
    \centering
    \includegraphics[width=\linewidth]{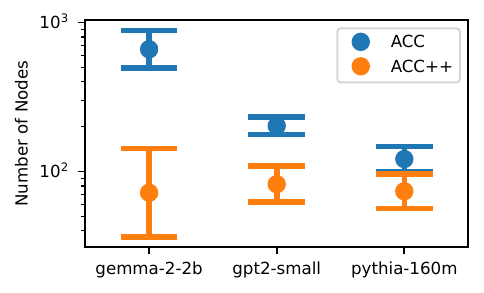}
    \caption{IOI}
    \label{fig:nodes:ioi}
  \end{subfigure}\hfill
  \begin{subfigure}[t]{0.32\linewidth}
    \centering
    \includegraphics[width=\linewidth]{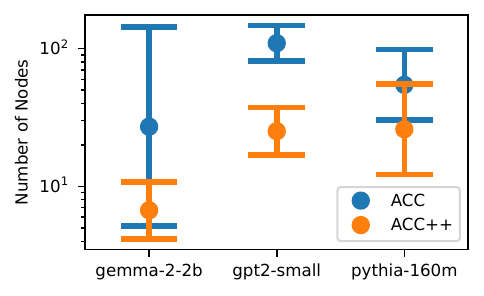}
    \caption{GP}
    \label{fig:nodes:gp}
  \end{subfigure}\hfill
  \begin{subfigure}[t]{0.32\linewidth}
    \centering
    \includegraphics[width=\linewidth]{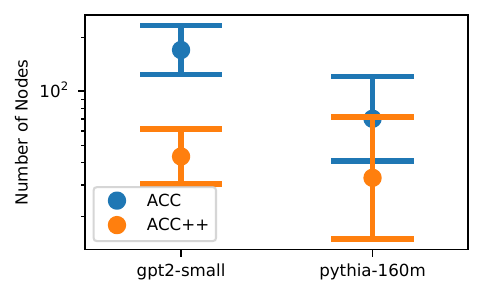}
    \caption{GT}
    \label{fig:nodes:gt}
  \end{subfigure}
\caption{Average number of nodes in ACC and ACC++ circuits for IOI, GP, and GT. Error bars show standard deviation.}
  \label{fig:n_nodes}
\end{figure}

\begin{figure}[ht]
  \centering
  \begin{subfigure}[t]{0.32\linewidth}
    \centering
    \includegraphics[width=\linewidth]{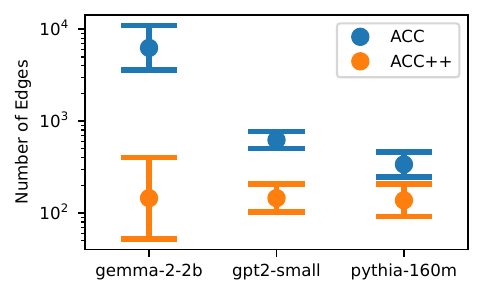}
    \caption{IOI}
    \label{fig:edges:ioi}
  \end{subfigure}\hfill
  \begin{subfigure}[t]{0.32\linewidth}
    \centering
    \includegraphics[width=\linewidth]{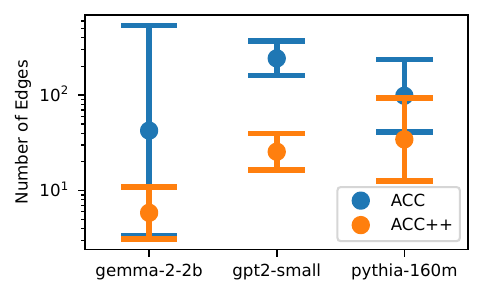}
    \caption{GP}
    \label{fig:edges:gp}
  \end{subfigure}\hfill
  \begin{subfigure}[t]{0.32\linewidth}
    \centering
    \includegraphics[width=\linewidth]{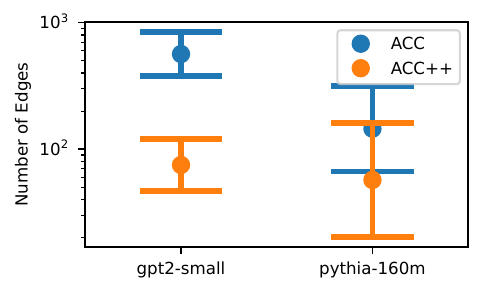}
    \caption{GT}
    \label{fig:edges:gt}
  \end{subfigure}
\caption{Average number of edges in ACC and ACC++ circuits for IOI, GP, and GT. Error bars show standard deviation.}
\label{fig:n_edges}
\end{figure}

\begin{figure}[t]
  \centering
  \begin{subfigure}[t]{0.32\linewidth}
    \centering
    \includegraphics[width=\linewidth]{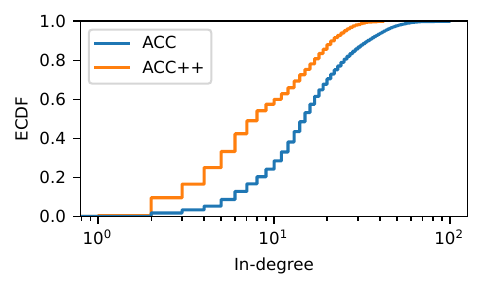}
    \caption{Gemma-2 2B}
    \label{fig:ecdf:ioi:gemma}
  \end{subfigure}\hfill
  \begin{subfigure}[t]{0.32\linewidth}
    \centering
    \includegraphics[width=\linewidth]{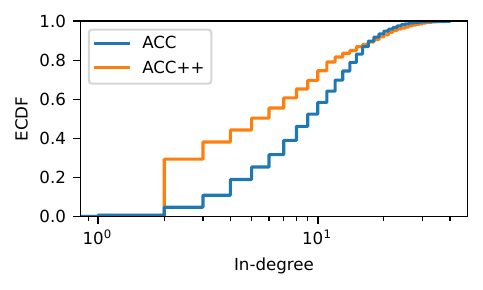}
    \caption{GPT-2 Small}
    \label{fig:ecdf:ioi:gpt2}
  \end{subfigure}\hfill
  \begin{subfigure}[t]{0.32\linewidth}
    \centering
    \includegraphics[width=\linewidth]{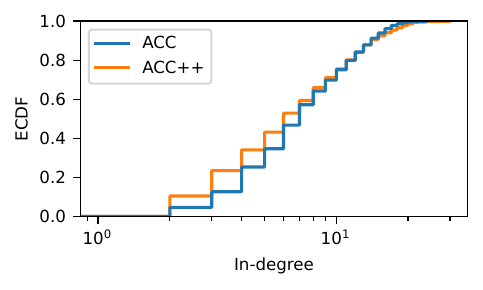}
    \caption{Pythia-160M}
    \label{fig:ecdf:ioi:pythia}
  \end{subfigure}
    \caption{ECDFs of positive node in-degrees across ACC and ACC++ circuits for IOI.}
  \label{fig:ecdf:ioi}
\end{figure}

\begin{figure}[t]
  \centering
  \begin{subfigure}[t]{0.32\linewidth}
    \centering
    \includegraphics[width=\linewidth]{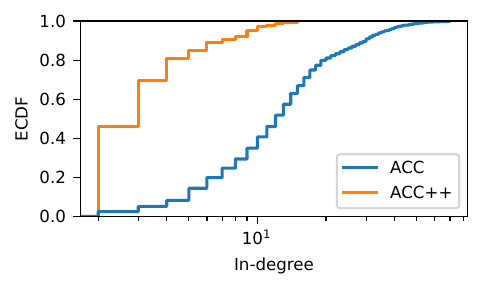}
    \caption{Gemma-2 2B}
    \label{fig:ecdf:gp:gemma}
  \end{subfigure}\hfill
  \begin{subfigure}[t]{0.32\linewidth}
    \centering
    \includegraphics[width=\linewidth]{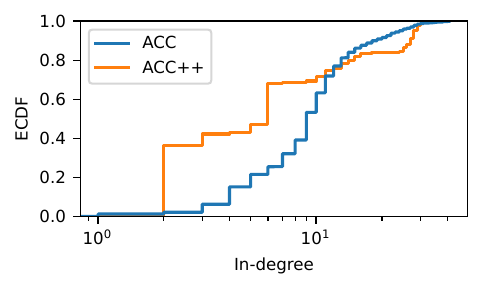}
    \caption{GPT-2 Small}
    \label{fig:ecdf:gp:gpt2}
  \end{subfigure}\hfill
  \begin{subfigure}[t]{0.32\linewidth}
    \centering
    \includegraphics[width=\linewidth]{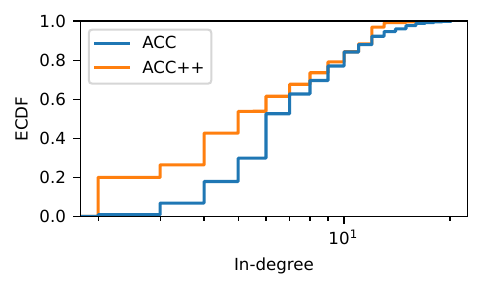}
    \caption{Pythia-160M}
    \label{fig:ecdf:gp:pythia}
  \end{subfigure}
\caption{ECDFs of positive node in-degrees across ACC and ACC++ circuits for GP.}

  \label{fig:ecdf:gp}
\end{figure}

\begin{figure}[t]
  \centering
  \begin{subfigure}[t]{0.32\linewidth}
    \centering
    \includegraphics[width=\linewidth]{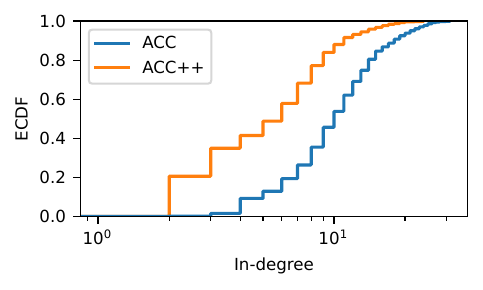}
    \caption{GPT-2 Small}
    \label{fig:ecdf:gt:gpt2}
  \end{subfigure}\hspace{0.04\linewidth}
  \begin{subfigure}[t]{0.32\linewidth}
    \centering
    \includegraphics[width=\linewidth]{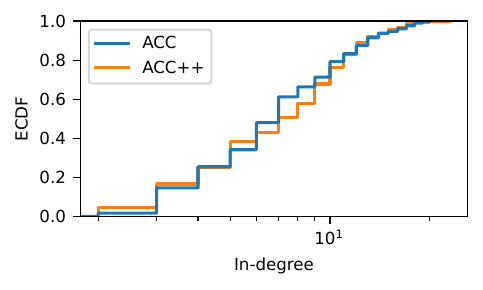}
    \caption{Pythia-160M}
    \label{fig:ecdf:gt:pythia}
  \end{subfigure}
\caption{ECDFs of positive node in-degrees across ACC and ACC++ circuits for GT.}
  \label{fig:ecdf:gt}
\end{figure}

\begin{figure}[ht]
    \centering    
    \includegraphics[width=\textwidth]{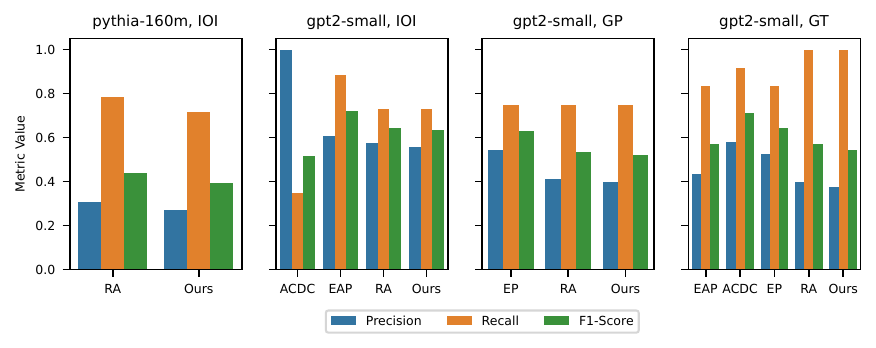}
    \caption{Precision, recall, and F1 against reference circuits under the experimental setup of \cite{franco2025pinpointing}. We compare ACC++ (Ours) with ACC (Relative Attention; RA) and the patching-based methods reported in that work (ACDC, EAP, and EP).}
    \label{fig:circuit-performance-comparison}
\end{figure}

ACC++ achieves precision, recall, and F1 scores comparable to ACC against the reference circuits (Figure~\ref{fig:circuit-performance-comparison}). ACC++ circuits also have fewer nodes and edges on average (Figures~\ref{fig:n_nodes} and~\ref{fig:n_edges}). The in-degree distributions generally show fewer incoming edges per node (Figures~\ref{fig:ecdf:ioi}, \ref{fig:ecdf:gp}, and~\ref{fig:ecdf:gt}).

\begin{figure}[ht]
\centering
\begin{subfigure}{.33\textwidth}
  \centering
  \includegraphics[width=1\linewidth]{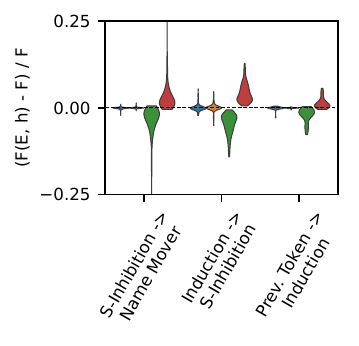}
  \caption{GPT-2 Small}
\end{subfigure}%
\begin{subfigure}{.33\textwidth}
  \centering
  \includegraphics[width=1\linewidth]{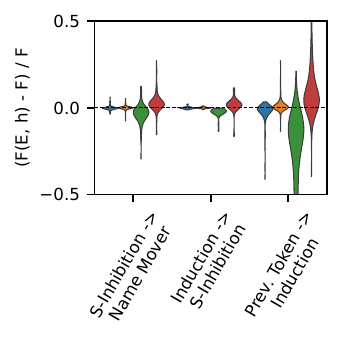}
  \caption{Pythia-160M}
\end{subfigure}
\begin{subfigure}{.33\textwidth}
  \centering
  \includegraphics[width=1\linewidth]{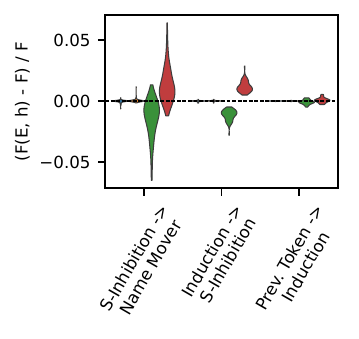}
  \caption{Gemma-2 2B}
\end{subfigure}
\caption{Effects of signal ablation and boosting on IOI performance. Green and red violins show ablation and boosting of ACC++ signals, respectively; blue and orange violins show the corresponding random-signal controls.}
\label{fig:intervention-effects}
\end{figure}
\begin{figure}[ht]
\centering
\begin{subfigure}{.48\textwidth}
  \centering
  \includegraphics[width=1\linewidth]{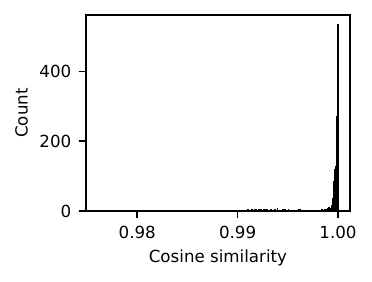}
  \caption{Cosine similarity}
\end{subfigure}%
\begin{subfigure}{.48\textwidth}
  \centering
  \includegraphics[width=1\linewidth]{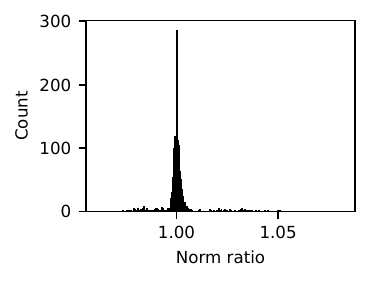}
  \caption{Norm ratio}
\end{subfigure}
\caption{Cosine similarities and norm ratios between residual vectors before and after ACC++ signal interventions. Both distributions are concentrated near 1, showing that the interventions produce small changes in the residual vectors.}
\label{fig:cosine-similarity-norm-ratio}
\end{figure}

ACC++ signals also affect model behavior: intervening on these signals changes IOI performance more than intervening on random signals (Figure~\ref{fig:intervention-effects}). These interventions produce small changes in the residual vectors, with cosine similarities and norm ratios concentrated near 1 (Figure~\ref{fig:cosine-similarity-norm-ratio}).

%% file: sections/appendix/signals_interp_proof.tex
\section{Autointerpretation pipeline details} \label{app:interp-signals}

This appendix provides implementation details for the autointerpretation results in Section~\ref{sec:improv-acc}. We first explain how we form a \emph{signal pair} by finding a corresponding source direction for each destination signal or vice versa (Appendix~\ref{app:interp-signals-find-pairs}). Because attention is a bilinear interaction between destination and source tokens, a signal alone does not specify what it matches against. Pairing provides the corresponding direction, allowing us to retrieve and interpret destination--source token contexts.

We then describe our fully automated interpretation pipeline, adapted from \cite{paulo2025automatically}, which (i) retrieves high-scoring token-pair contexts for each signal from \texttt{pile-10k}, (ii) prompts an LLM to produce a short natural-language interpretation from the top contexts, and (iii) uses an LLM judge to evaluate the interpretation on top contexts and random controls, then applies our statistical criterion to label signals as interpretable (Appendix~\ref{app:autointerp-pipeline-desc}).

\subsection{Finding signal pairs} \label{app:interp-signals-find-pairs}

We fix a subset of singular vector indices $I = \{i_1,\dots,i_k\}$ and consider a \emph{destination signal} $\v{p}$ in the subspace spanned by the corresponding left singular vectors,
\[
\v{p} \in \operatorname{Span}\{\v{u}^{i_1},\dots,\v{u}^{i_k}\}, \qquad \|\v{p}\| = 1.
\]
We define the \emph{corresponding source signal} as the unit vector $\v{q}^{*}$ given by
\[
\v{q}^{*} \;=\; \arg\max_{\|\v{q}\| = 1}\ \v{p}^\top \Omega\,\v{q}.
\]

The next lemma shows that, when $\Omega^\top \v{p}\neq\mathbf{0}$, the optimal source signal is the normalized vector $\Omega^\top \v{p}$ and lies in the subspace spanned by the right singular vectors indexed by $I$.

\begin{lemma}\label{lem:signal-pair}
Let $\Omega \in \mathbb{R}^{D \times D}$ have SVD
\[
\Omega = \sum_{m=1}^R \sigma^{m}\, \v{u}^{m} \v{v}^{m \top},
\]
and let $I = \{i_1,\dots,i_k\}$ be a subset of $\{1, \cdots, R\}$. Define
\[
U_I = [\,\v{u}^{i_1}\ \dots\ \v{u}^{i_k}\,],\qquad
V_I = [\,\v{v}^{i_1}\ \dots\ \v{v}^{i_k}\,],\qquad
\Sigma_I = \operatorname{diag}(\sigma^{i_1},\dots,\sigma^{i_k}).
\]
Let $\v{p} \in \operatorname{Span}\{\v{u}^{i_1},\dots,\v{u}^{i_k}\}$ with $\|\v{p}\| = 1$. Consider
\[
\v{q}^{*} \;=\; \arg\max_{\|\v{q}\| = 1}\ \v{p}^\top \Omega\, \v{q}.
\]
Then, if $\Omega^\top \v{p} \neq \mathbf{0}$,
\[
\v{q}^* \;=\; \frac{\Omega^\top \v{p}}{\|\Omega^\top \v{p}\|},
\]
and in particular $\v{q}^* \in \operatorname{Span}\{\v{v}^{i_1},\dots,\v{v}^{i_k}\}$. If $\Omega^\top \v{p} = \mathbf{0}$, then $\v{p}^\top \Omega \v{q} = 0$ for all $\|\v{q}\|=1$ and every unit $\v{q}$ is optimal.
\end{lemma}

\begin{proof}
First, assume $\Omega^\top\v{p}\neq\mathbf{0}$. Since $\v{p} \in \operatorname{Span}\{\v{u}^{i_1},\dots,\v{u}^{i_k}\}$ and $\|\v{p}\|=1$, there exists $\boldsymbol{\alpha} \in \mathbb{R}^k$ with
\[
\v{p} = U_I \boldsymbol{\alpha},
\qquad
\|\boldsymbol{\alpha}\| = 1,
\]
because the columns of $U_I$ are orthonormal.

We want to solve
\[
\v{q}^* \;=\; \arg\max_{\|\v{q}\| = 1}\ \v{p}^\top \Omega\, \v{q}.
\]
First, note that we can, without loss of generality, restrict $\v{q}$ to the span of the right singular vectors $\{\v{v}^{i_1},\dots,\v{v}^{i_k}\}$. Indeed, decompose any $\v{q}$ as
\[
\v{q} = V_I \boldsymbol{\beta} + \mathbf{w},
\]
where $\boldsymbol{\beta} \in \mathbb{R}^k$ and $\mathbf{w} \perp \operatorname{Span}\{\v{v}^{i_1},\dots,\v{v}^{i_k}\}$. Using the SVD,
\[
\v{p}^\top \Omega \v{q}
= \sum_{m=1}^R \sigma^{m} (\v{u}^{m \top} \v{p})(\v{v}^{m \top} \v{q}).
\]
Since $\v{p} \in \operatorname{Span}\{\v{u}^{i_1},\dots,\v{u}^{i_k}\}$, we have $\v{u}^{m \top} \v{p} = 0$ for all $m \notin I$, so
\[
\v{p}^\top \Omega \v{q}
= \sum_{m\in I} \sigma^{m} (\v{u}^{m \top} \v{p})(\v{v}^{m \top} \v{q})
= \sum_{m\in I} \sigma^{m} (\v{u}^{m \top} \v{p})(\v{v}^{m \top} V_I\boldsymbol{\beta}),
\]
because $\v{v}^{m \top} \mathbf{w} = 0$ for all $m\in I$. The component $\mathbf{w}$ contributes nothing to the objective, but contributes $\|\mathbf{w}\|^2$ to the squared norm of $\v{q}$. Since the maximum is positive when $\Omega^\top\v{p}\neq\mathbf{0}$, allocating any norm to $\mathbf{w}$ reduces the maximum possible value. We can therefore restrict to
\[
\v{q} = V_I \boldsymbol{\beta},\qquad \|\boldsymbol{\beta}\| = 1.
\]

Because $V_I$ has orthonormal columns, we have
\[
\|\v{q}\|^2 = \|V_I \boldsymbol{\beta}\|^2 
= \boldsymbol{\beta}^\top V_I^\top V_I \boldsymbol{\beta} 
= \|\boldsymbol{\beta}\|^2,
\]
so the constraint $\|\v{q}\|=1$ is equivalent to $\|\boldsymbol{\beta}\|=1$.

Substituting these expressions into the objective gives:
\[
\v{p}^\top \Omega \v{q}
= \boldsymbol{\alpha}^\top U_I^\top \Omega V_I \boldsymbol{\beta}.
\]
Since $U_I^\top \Omega V_I = \Sigma_I$, we have
\[
\v{p}^\top \Omega \v{q}
= \boldsymbol{\alpha}^\top \Sigma_I \boldsymbol{\beta}.
\]
Thus the problem reduces to
\[
\boldsymbol{\beta}^* \;=\; \arg\max_{\|\boldsymbol{\beta}\|=1}\ \boldsymbol{\alpha}^\top \Sigma_I \boldsymbol{\beta}.
\]
Define $\boldsymbol{\gamma} := \Sigma_I \boldsymbol{\alpha} \in \mathbb{R}^k$. Then
\[
\boldsymbol{\alpha}^\top \Sigma_I \boldsymbol{\beta} = \boldsymbol{\gamma}^\top \boldsymbol{\beta}.
\]


Since $\Omega^\top\v{p}=V_I\boldsymbol{\gamma}\neq\mathbf{0}$, we have $\boldsymbol{\gamma}\neq\mathbf{0}$. By the Cauchy--Schwarz inequality,
\[
\boldsymbol{\gamma}^\top \boldsymbol{\beta} \;\le\; \|\boldsymbol{\gamma}\|\,\|\boldsymbol{\beta}\|
= \|\boldsymbol{\gamma}\|,
\qquad \|\boldsymbol{\beta}\| = 1.
\]
Equality holds if and only if $\boldsymbol{\beta}$ is a positive
multiple of $\boldsymbol{\gamma}$. Thus to maximize $\boldsymbol{\gamma}^\top \boldsymbol{\beta}$ under $\|\boldsymbol{\beta}\|=1$, we must have
\[
\boldsymbol{\beta}^* = \lambda\, \boldsymbol{\gamma}
\]
for some scalar $\lambda$. Enforcing $\|\boldsymbol{\beta}^*\|=1$ gives
\[
1 = \|\boldsymbol{\beta}^*\| = |\lambda|\,\|\boldsymbol{\gamma}\|
\quad\Rightarrow\quad
|\lambda| = \frac{1}{\|\boldsymbol{\gamma}\|}.
\]
To maximize (rather than minimize) $\boldsymbol{\gamma}^\top \boldsymbol{\beta}$, we take the positive sign, so
\[
\boldsymbol{\beta}^* = \frac{\boldsymbol{\gamma}}{\|\boldsymbol{\gamma}\|}.
\]

Substituting into $\v{q}^* = V_I \boldsymbol{\beta}^*$,
\[
\v{q}^* = V_I \boldsymbol{\beta}^* = V_I \frac{\boldsymbol{\gamma}}{\|\boldsymbol{\gamma}\|}
= \frac{V_I \boldsymbol{\gamma}}{\|\boldsymbol{\gamma}\|}.
\]

We now express $\Omega^\top \v{p}$ in terms of this basis:
\[
\Omega^\top \v{p}
= \sum_{m=1}^R \sigma^{m} \v{v}^{m} \v{u}^{m \top} \v{p}
= \sum_{m\in I} \sigma^{m} \v{v}^{m} (\v{u}^{m \top} \v{p})
= V_I (\Sigma_I \boldsymbol{\alpha})
= V_I \boldsymbol{\gamma}.
\]
Therefore
\[
\|\Omega^\top \v{p}\|
= \|V_I \boldsymbol{\gamma}\| = \|\boldsymbol{\gamma}\|
\]
(because $V_I$ has orthonormal columns), and we obtain
\[
\v{q}^* 
= \frac{V_I \boldsymbol{\gamma}}{\|\boldsymbol{\gamma}\|}
= \frac{\Omega^\top \v{p}}{\|\Omega^\top \v{p}\|}.
\]

This maximizer is unique and lies in $\operatorname{Span}\{\v{v}^{i_1},\dots,\v{v}^{i_k}\}$, since $\Omega^\top\v{p}=V_I\boldsymbol{\gamma}$.

Finally, if $\Omega^\top\v{p}=\mathbf{0}$, then
$\v{p}^\top\Omega\v{q}=(\Omega^\top\v{p})^\top\v{q}=0$
for every unit vector $\v{q}$, so every such vector is optimal.

\end{proof}

\paragraph{Corollary (pairing a source signal).}
The same result holds if we fix a \emph{source} signal and optimize over the destination. Let $\v{q} \in \operatorname{Span}\{\v{v}^{i_1},\dots,\v{v}^{i_k}\}$ with $\|\v{q}\|=1$, and consider
\[
\v{p}^{*} \;=\; \arg\max_{\|\v{p}\|=1}\ \v{p}^\top \Omega\,\v{q}.
\]
Writing $\v{p}^\top \Omega\,\v{q} = \langle \v{p},\,\Omega\v{q}\rangle$ and applying the Cauchy--Schwarz inequality exactly as in Lemma~\ref{lem:signal-pair}, we obtain
\[
\v{p}^{*} \;=\; \frac{\Omega\,\v{q}}{\|\Omega\,\v{q}\|},
\]
whenever $\Omega\v{q}\neq\mathbf{0}$. In this case, $\v{p}^*$ lies in the subspace spanned by the left singular vectors indexed by $I$. If $\Omega\v{q}=\mathbf{0}$, every unit vector $\v{p}$ is optimal.

\paragraph{Pairing signals in RoPE models.}
For models with RoPE, with or without bias, Appendix~\ref{app:unified-bilinear} expresses the attention score as
\[
A'_{ds} = \tv{x}^{d\top}\Omega\tv{x}^s,
\]
where $\tv{x}^{d\top}=(\v{x}^{d\top}+{\v{c}^d}^{\top})M_d$ and $\tv{x}^s=M_s(\v{x}^s+\v{c}^s)$, with $\v{c}^d=\v{c}^s=\mathbf{0}$ for models without bias. Lemma~\ref{lem:signal-pair} and its corollary therefore apply to signal directions in this transformed space, using the same fixed matrix $\Omega=W_QW_K^\top$.

\subsection{Automated interpretation pipeline description} \label{app:autointerp-pipeline-desc}

We adopt the autointerpretation pipeline methodology from \cite{paulo2025automatically} and adapt it to ACC++ signals. The pipeline consists of three stages: identification of high-scoring examples, generation of a natural-language interpretation for each signal, and quantitative evaluation of interpretation quality. All stages of interpretation and evaluation are fully automated. No human annotations, labeling, or judgments are used at any point in the pipeline.

\paragraph{Finding high-scoring contexts.}
We use the \texttt{NeelNanda/pile-10k} split, consisting of the first 10{,}000 sequences of The Pile \cite{gao2020pile}. To find contexts where a given signal pair is active, we cache the model's activations over \texttt{pile-10k} in 32-token chunks. Given an attention head and a normalized signal pair $(\v{p},\v{q})$, let $\tv{x}^d,\tv{x}^s$ denote the input vectors after the bias and RoPE transformations, where applicable. We score each token pair by the product of its projections onto $\v{p}$ and $\v{q}$:
\[
\mathrm{score}(d,s;\v{p},\v{q})
=
(\tv{x}^{d\top}\v{p})(\v{q}^\top\tv{x}^s),
\]
which is a proxy for the contribution of the paired signals to the head's attention logit for $(d,s)$.

\paragraph{Construction of signal examples.}
Each selected context is a 32-token text span containing both the destination and source tokens. We mark the destination token with \texttt{<< >>} and the source token with \texttt{[[ ]]}. If the same token serves as both, we use \texttt{<<[[token]]>>}.

We use these examples in two stages:
\begin{enumerate}
    \item \textbf{Interpretation stage:} we prompt an LLM with the 40 highest-scoring examples for a signal to generate a short interpretation.
    \item \textbf{Evaluation stage:} we take the top-20 examples (highest-scoring among the top-40) and mix them with 20 random control examples \emph{sampled independently for that same signal}, then ask an LLM judge to decide whether the interpretation explains each example \cite{paulo2025automatically}.
\end{enumerate}

\paragraph{Interpretation stage.}
We use \texttt{neuralmagic/DeepSeek-R1-Distill-Llama-70B-FP8-dynamic} as the interpreter model, run via vLLM \cite{deepseekai2025deepseekr1incentivizingreasoningcapability}. The interpreter is shown only the top-scoring examples and is instructed to produce one concise, high-level description of those examples. The full interpretation prompt is reproduced below.

\begin{promptbox}{System Prompt for the Signal Interpretation}
You are a meticulous AI researcher conducting an important investigation into how hidden signals in language models correspond to patterns in text. Your task is to analyze the given text snippets and provide an interpretation that clearly summarizes the linguistic or semantic pattern they reveal.  

Guidelines:  
- You will be shown several short text examples.
- In each example, one or two important tokens will be surrounded by << >> and [[ ]] (e.g. "the [[cat]] sat on the << mat>>").
- These brackets indicate where the model's internal signal was most active. Signals are found in the context of attention heads, meaning that the token highlighted with << >> acts as a destination token, while the token highlighted with [[ ]] acts as a source token. If only one token is highlighted with both << >> and [[ ]], it means that this token acts as both destination and source tokens.
- Each example may include additional context words around the highlighted token(s).
- Your goal is to produce a concise, high-level description of what these highlighted tokens have in common (their meaning, grammatical role, or thematic pattern possibility).
- Focus on the underlying pattern, not on quoting or restating the examples.
- Remember to look at the position of the tokens in the text chunk. This also matters. 
- Ignore the brackets and any tokenization artifacts when describing your interpretation.
- If the examples appear random or uninformative, say so briefly.
- Do not list multiple hypotheses; choose the single best interpretation.
- Keep your answer short and clear.
- If you are not more than 90\% certain about an interpretation, say "no valid interpretation found".
- The final line of your response must provide your conclusion in this exact format:
[interpretation]: your concise description here
\end{promptbox}

The interpreter produces one interpretation per signal or returns \texttt{no valid interpretation found}. Signals without an interpretation are excluded from evaluation. We run the pipeline over 13{,}406 signals for GPT-2 Small, 15{,}316 for Pythia-160M, and 22{,}223 for Gemma-2 2B. The fractions without an interpretation are 0.850\%, 0.359\%, and 0.576\%, respectively.

\paragraph{Gemini for qualitative studies.}
For a small number of representative circuits in our qualitative analysis (Figure~\ref{fig:gemma-facts-0}, Figure~\ref{fig:abba-baba-circuits-gpt}, and the additional traced examples in Appendix~\ref{app:trace-examples}), we use Gemini~3 Flash to generate interpretations for the signals. We chose Gemini to obtain the best possible interpretations for these manually analyzed examples, which do not require running the full evaluation pipeline. We did not use it for the full automated pipeline because of its higher cost.

\paragraph{Interpretation scoring setup.}
We evaluate each proposed interpretation using a separate judge model, Gemma-3-27B-IT \cite{team2025gemma}, also run via vLLM. The judge is used to assess whether the interpretation distinguishes the highest-scoring contexts from random controls. For each signal with an interpretation, we score 40 mixed examples: the 20 highest-scoring examples and 20 random examples sampled for that signal. We evaluate these examples in independent batches of 10, each containing 5 highest-scoring examples and 5 random examples. For each prompt, the judge returns a binary label for each example, indicating whether all marked tokens are consistent with the interpretation. The full scoring prompt is included below.

\begin{promptbox}{Fuzzing Scoring Prompt}
You are an intelligent and meticulous linguistics researcher.

You will be given a specific linguistic feature of interest, such as  "male pronouns," "negative sentiment," or "surname tokens."

You will then be given several text examples that are claimed to contain this feature. Portions of the text that supposedly represent the feature have been marked using << >> and [[ ]].
These brackets indicate where the model's internal signal was most active. 
Features are found in the context of attention heads, meaning that the token highlighted with << >> acts as a destination token, while the token highlighted with [[ ]] acts as a source token. If only one token is highlighted with both << >> and [[ ]], it means that this token acts as both destination and source tokens.

Your task is to determine whether EVERY token inside each << >> and [[ ]] span is correctly labeled as an instance of the feature.

Important:
- An example is correct ONLY if every marked tokens are representative of the feature.
- There are exactly 10 examples below.

For each example in turn:
- Return 1 if ALL marked spans correctly represent the feature.
- Return 0 if ANY marked token is mislabeled.

Important Output Format Rules:
1. Return the results as a Python Dictionary.
2. The keys must be the example numbers (1 to 10), and the values must be the binary label (0 or 1).
3. Do not assume the order; explicitly check the number I assigned to each example.
4. Ignore any numbers or formatting artifacts INSIDE the text strings (e.g., if a text contains "4).", ignore it).
5. Output format:
   {
    1: 0,
    2: 1,
    ...
    10: 0
   }

Here are the examples:

<user_prompt> 
Feature interpretation: Words related to American football positions, specifically the tight end position.
Text examples:
1. Getty Images [[ Patriots]]<< tight>> end Rob Gronkowski had his boss
2. posted You should know this[[ about]] offensive line coaches: they are large, demanding<< men>>
3. Media Day 2015 LSU [[ defensive]] end Isaiah Washington (94) speaks to<< the>>
4. running [[ backs]],'' he said. .. Defensive << end>> Carroll Phillips is improving and his injury is
5. [[ line]], with the left side namely << tackle>> Byron Bell at tackle and guard Amini
<assistant_response> 
{
1: 1,
2: 0,
3: 0,
4: 1,
5: 1,
}

Now evaluate the following examples:
\end{promptbox}

\paragraph{Example top-scoring contexts.}
We show a few retrieved top-scoring contexts for a representative signal.

\noindent\textbf{Prompt:} ``Then, Jack and Kelly went to the garden. Jack gave a basketball to" (BABA representative). \\
\noindent\textbf{Signal:} AH(8,6) ( to (1), Jack (1)) $\to$ AH(9,9) ( to (1), Kelly). \\
\noindent\textbf{Interpretation:} the second element in a parallel pair or sequence. \\
\noindent\textbf{Model-generated explanation:} The highlighted tokens consistently identify the second element in a parallel pair, list, or sequence (such as ``y'' following ``x'', ``b'' following ``a'', or an item labeled ``2'' following one labeled ``1''). The signal typically links this second element either to its own identifier or to the corresponding first element in the same or a preceding parallel structure.

\begin{promptbox}{Ten top-scoring contexts (Interpretation: second element in a parallel pair or sequence)}
- boost::arg<I> (*) () ): storage3<A1, A[[2]], A3>( a1,<< a>>2, a3 ) {}
- the function that initiates Ball(float x, float y, int Width, color Color) { ballX= x; ball[[Y]]<<=>> y; ballWidth
- speedX; ballY+=speedY; } void move(int X, int Y) { ballX=X; ball[[Y]]<<=>> Y; speedY
- ). \nfunction cartesianJoin(arr1, arr[[2]], col1,<< col>>2) {\n  var output = [];\n  for (var
- that initiates Block(float x, float y, float Width, float Height, color Color) { blockX= x; block[[Y]]<<=>> y; block
- $, where $C_1$ and $C_2$ are cycles of lengths $d_1$ and $d<<[[_]]>>2$ respectively, $d
- abnormal Ser phosphorylation of IRS-1 and IRS-2, along with degradation of IRS-1 and IRS<<[[-]]>>2. Moreover, these alterations were atten
- the function that initiates Ball(float x, float[[ y]], int Width, color Color) { ballX= x; ballY<<=>> y; ballWidth
- $\\{1,2,\\cdots,\n2m\\}$ into two element blocks $\\{(\\pi_1,\\pi<<[[_]]>>2),
- m)}(x_1,\\dots,x_m;\\tilde{k}_1,\\tilde{k}<<[[_]]>>2
\end{promptbox}

\begin{figure}[ht]
\centering
\begin{subfigure}{.33\textwidth}
  \centering
  \includegraphics[width=1\linewidth]{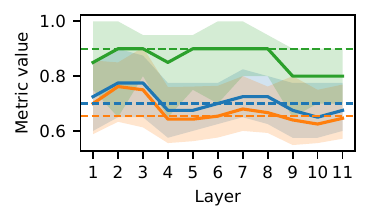}
  \caption{GPT-2 Small}
\end{subfigure}%
\begin{subfigure}{.33\textwidth}
  \centering
  \includegraphics[width=1\linewidth]{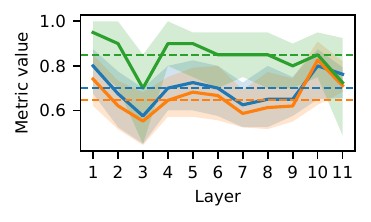}
  \caption{Pythia-160M}
\end{subfigure}
\begin{subfigure}{.33\textwidth}
  \centering
  \includegraphics[width=1\linewidth]{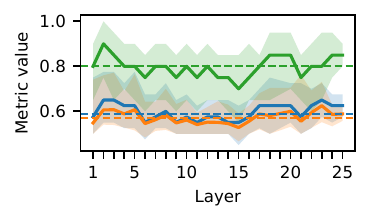}
  \caption{Gemma-2 2B}
\end{subfigure}
\caption{Interpretation evaluation metrics by layer for three models. Solid lines show median accuracy (blue), precision (orange), and recall (green) within each layer; shaded bands show the interquartile ranges across signals. Dashed horizontal lines show the corresponding medians across all signals, using the same colors as the corresponding solid lines.}
\label{fig:median-metrics-by-layer}
\end{figure}

\paragraph{Quantitative evaluation of interpretation quality.}
For each signal, we compare the judge's binary predictions with the labels identifying top-scoring and random examples. We compute accuracy, precision, and recall for each signal and report their medians and interquartile ranges across signals in Table~\ref{tab:interpretability-comparison}. Figure~\ref{fig:median-metrics-by-layer} shows these metrics by layer, with horizontal lines indicating the medians across all signals.

Along with these classification metrics, we compute a one-sided $p$-value using Fisher's exact test \cite{fisher1922interpretation} to test whether the judge accepts the interpretation more often on top-scoring contexts than on random contexts. We apply a false discovery rate (FDR) correction within each layer at the 5\% level \cite{benjamini1995controlling}. Signals passing the corrected test are labeled \emph{interpretable}, and we report their fraction for each model. This criterion tests whether an interpretation distinguishes top-scoring contexts from random controls under our evaluation protocol.

\input{sections/appendix/trace-examples}

%% file: sections/appendix/trace-examples.tex
\subsection{Interactive circuit visualizer}
\label{app:trace-examples}

For each qualitative trace shown in Figures~\ref{fig:gemma-facts-0}, \ref{fig:abba-baba-circuits-gpt}, \ref{fig:abba-baba-circuits-gpt-proper-noun}, \ref{fig:pythia-template-circuits}, and~\ref{fig:gemma-facts-1}, we provide an interactive HTML visualization in the supplementary material. The viewer displays the ACC++ circuit together with its automatically generated interpretations. Users can filter edges by a weight threshold, hide nodes that do not connect to the output, and inspect node and edge attributes, interpretations, and top-scoring contexts. The viewer supports any ACC++ trace with autointerpretation outputs. The examples below use the Gemma-2 2B trace for ``Fact: The capital of the state containing Dallas is'' (predicting ``Austin'').

We construct the static qualitative figures by inspecting these visualizations and selecting the main signals and components. Claude assisted this process by using the traced graph and automatically generated edge labels to produce a simplified view of the main paths in the graph. The interactive visualizations provide the underlying circuits for inspection.

\begin{figure}[!htbp]
    \centering
    \includegraphics[width=\textwidth, trim=10 10 10 0, clip]{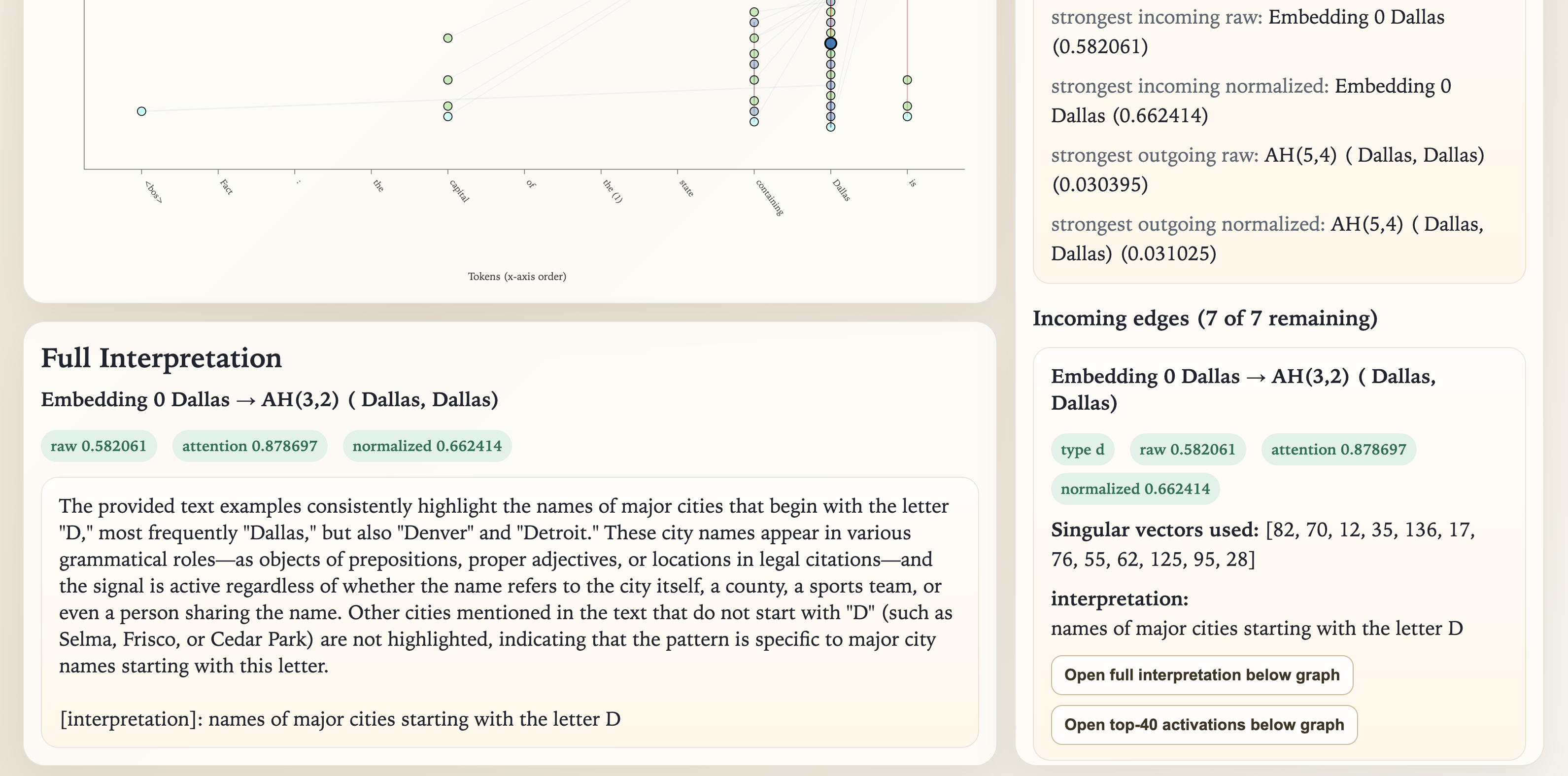}

    \vspace{3pt}

    \includegraphics[width=\textwidth, trim=7 0 7 5, clip]{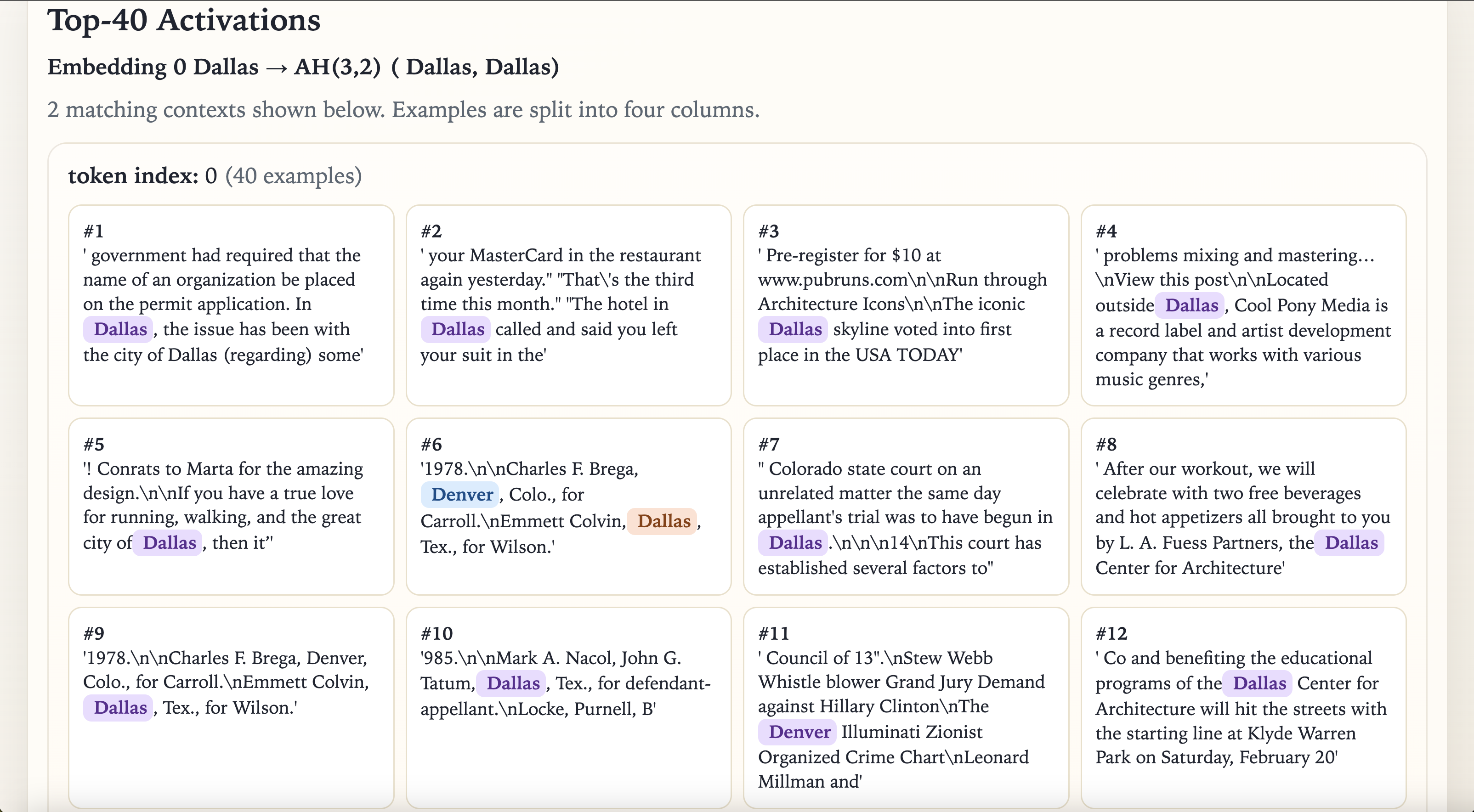}

    \caption{Autointerpretation and 40 highest-scoring contexts for the destination signal from the ``Dallas'' embedding, interpreted as names of major cities beginning with ``D''.}
    \label{fig:dallas-example1}
\end{figure}

\begin{figure}[!htbp]
    \centering

    \includegraphics[width=\textwidth, trim=0 0 0 2, clip]{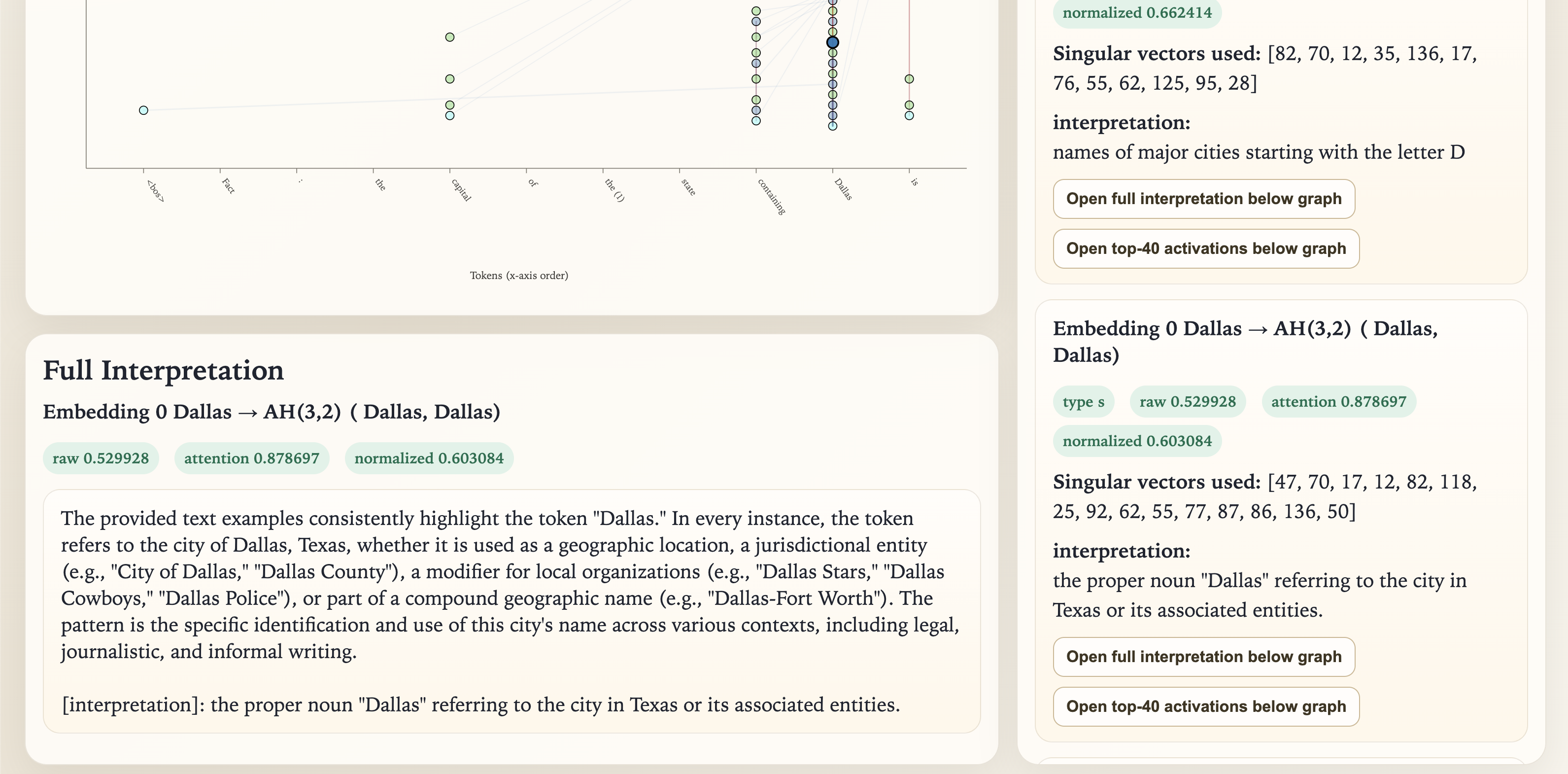}

    \vspace{3pt}

    \includegraphics[width=\textwidth, trim=0 0 0 0, clip]{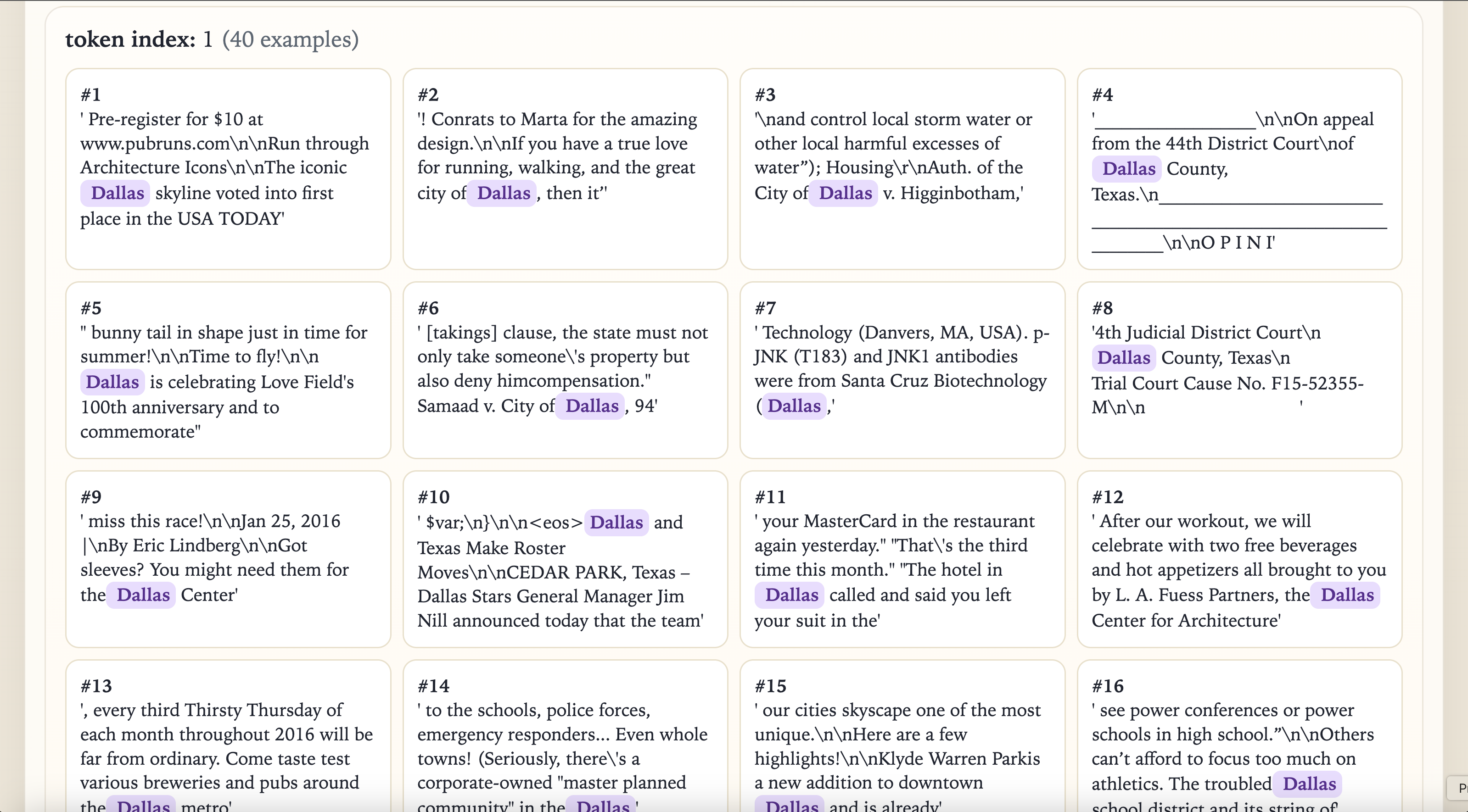}

    \caption{Autointerpretation and 40 highest-scoring contexts for the source signal from the same ``Dallas'' embedding, interpreted as references to Dallas, Texas, or its associated entities.}
    \label{fig:dallas-example2}
\end{figure}

\begin{figure}[!htbp]
    \centering
    \includegraphics[width=\textwidth, trim=0 0 0 0, clip]{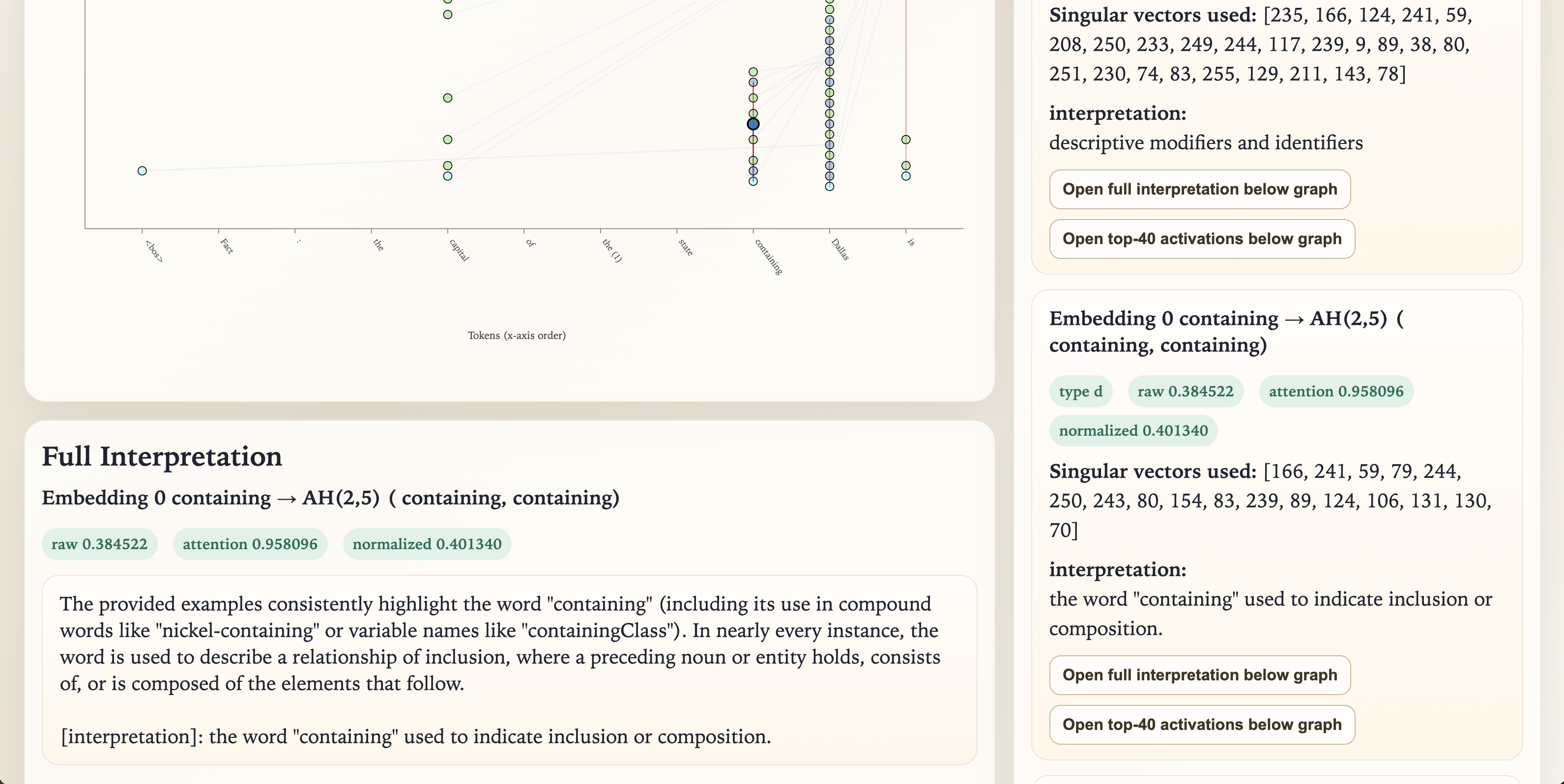}

    \vspace{3pt}

    \includegraphics[width=\textwidth, trim=10 0 10 0, clip]{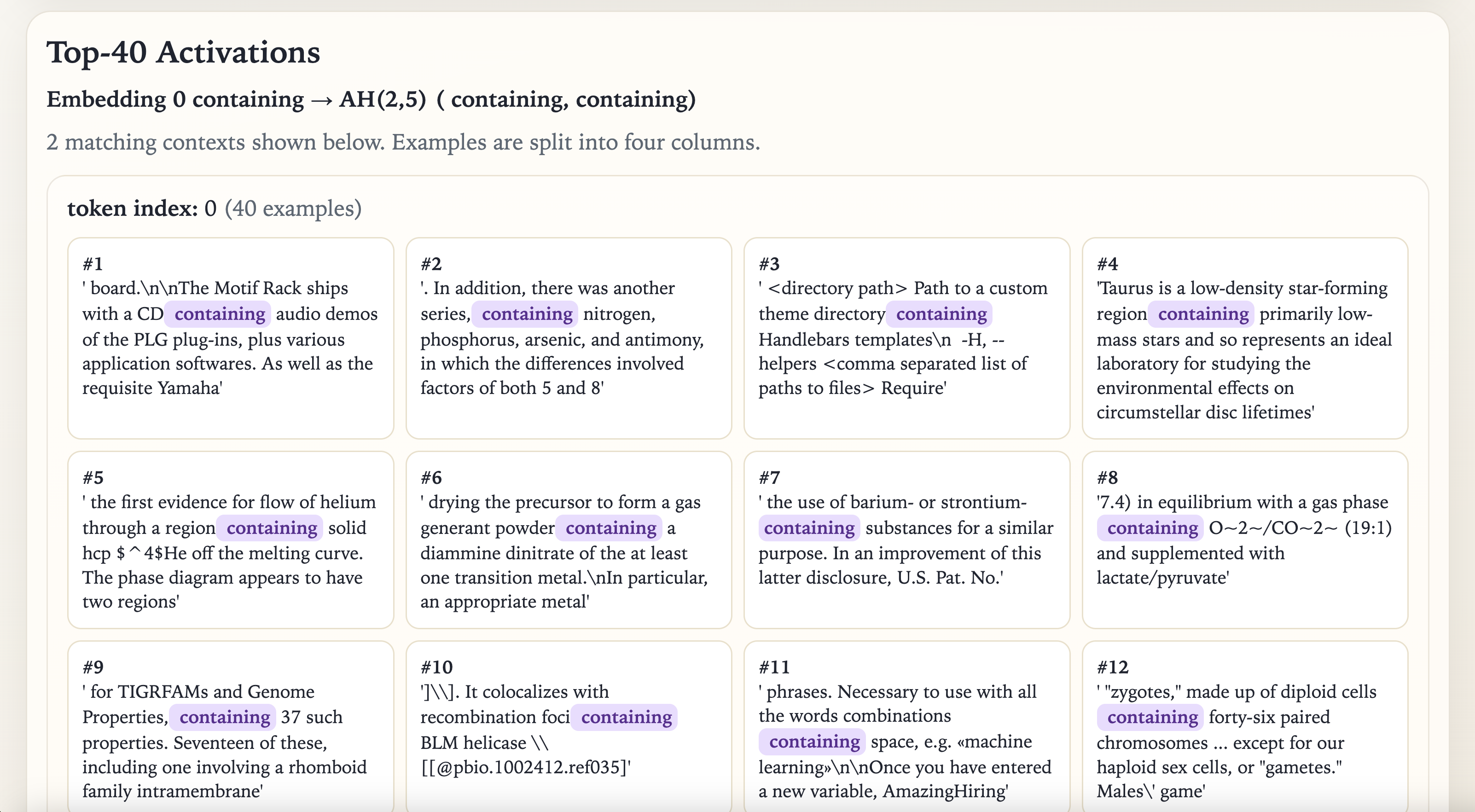}

    \caption{Autointerpretation and 40 highest-scoring contexts for the ``containing'' signal: a relational signal that captures inclusion patterns of the form ``X containing Y''.}
    \label{fig:dallas-example3}
\end{figure}

\begin{figure}[!htbp]
    \centering
    \includegraphics[width=\textwidth, trim=0 0 0 0, clip]{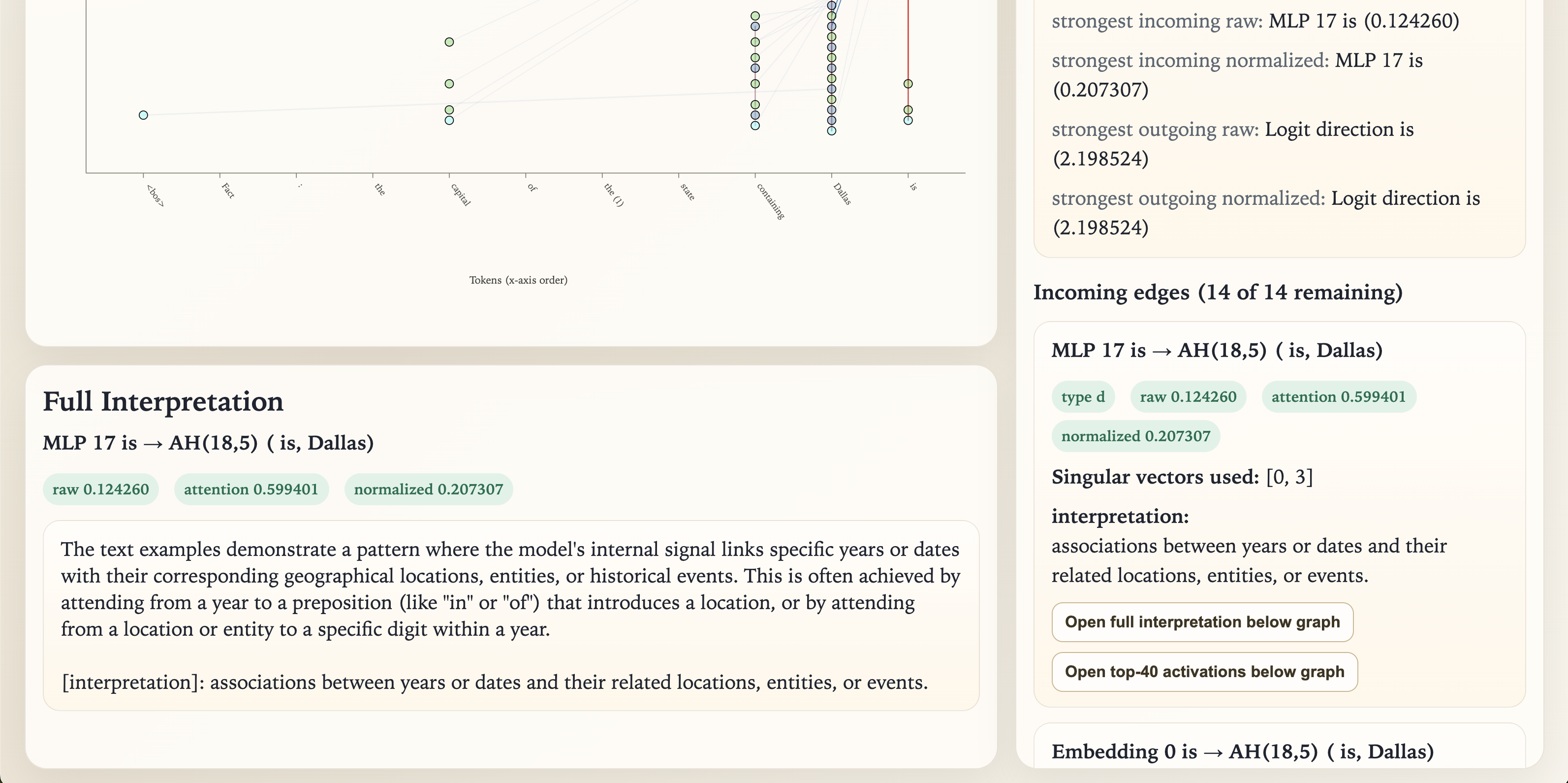}

    \vspace{3pt}

    \includegraphics[width=\textwidth, trim=0 0 0 0, clip]{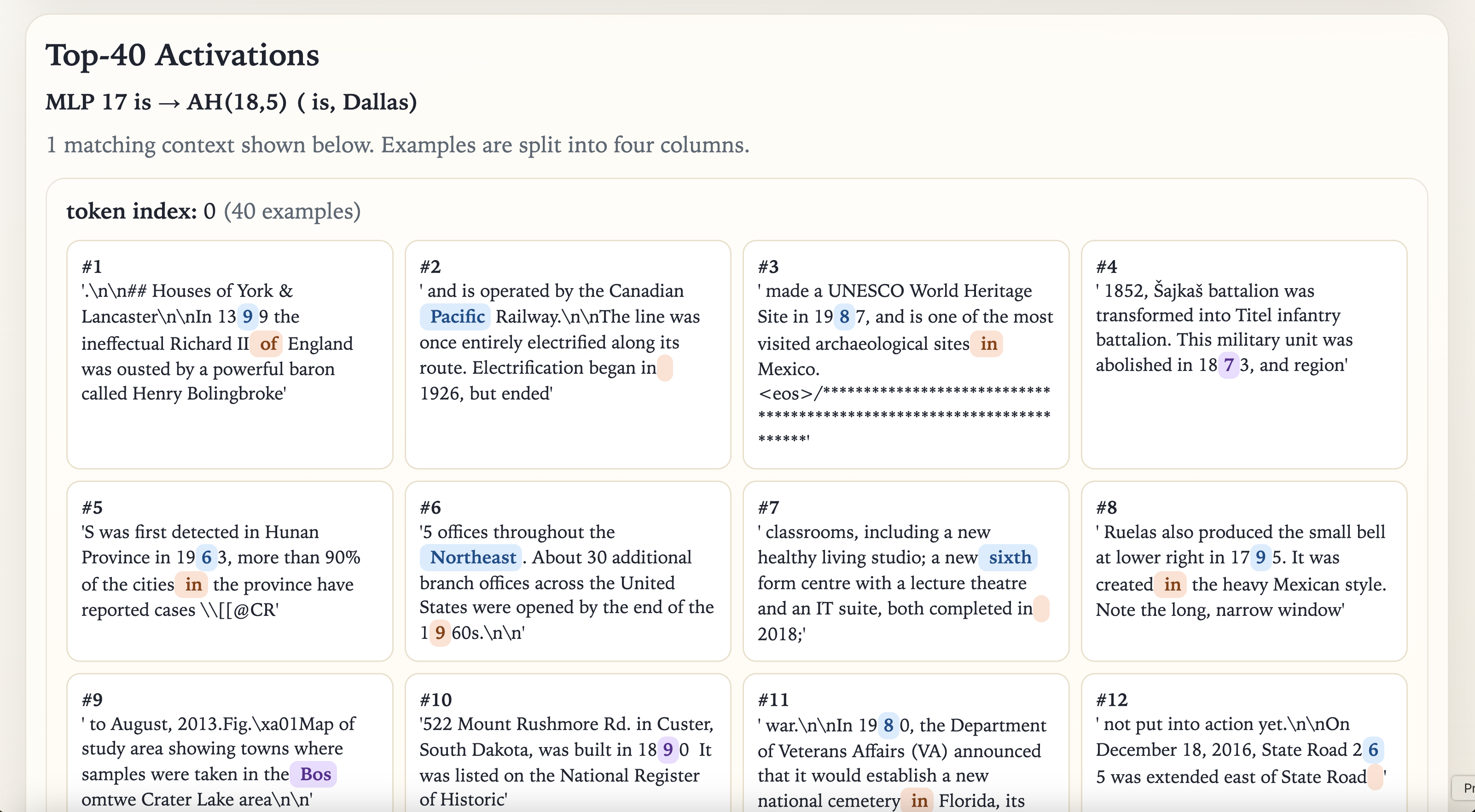}

    \caption{Autointerpretation and 40 highest-scoring contexts for a signal interpreted as associations between years or dates and related locations, entities, or events.}
    \label{fig:dallas-example4}
\end{figure}

\begin{figure}[!htbp]
    \centering
    \includegraphics[width=\textwidth, trim=0 0 0 0, clip]{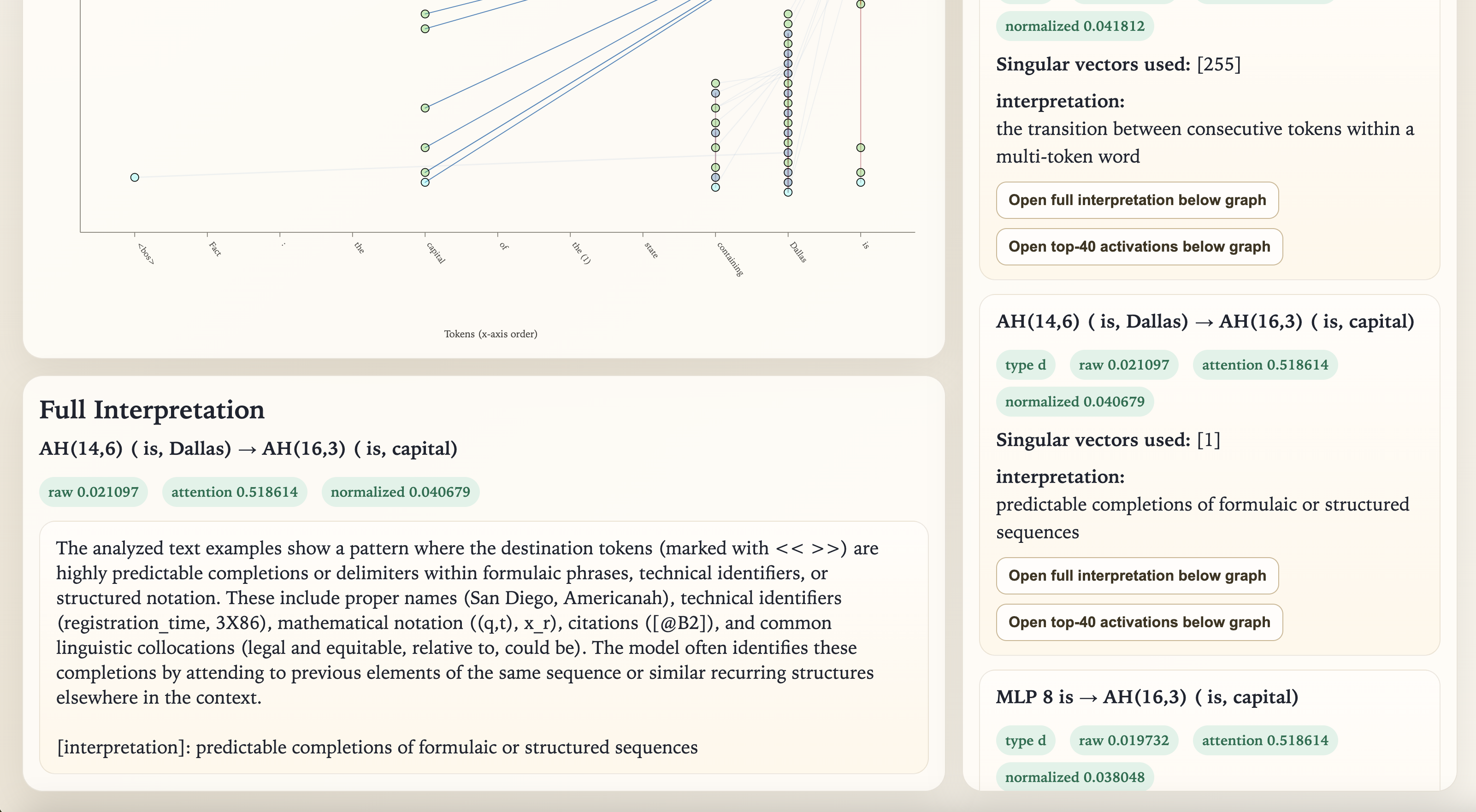}

    \vspace{3pt}

    \includegraphics[width=\textwidth, trim=0 0 0 0, clip]{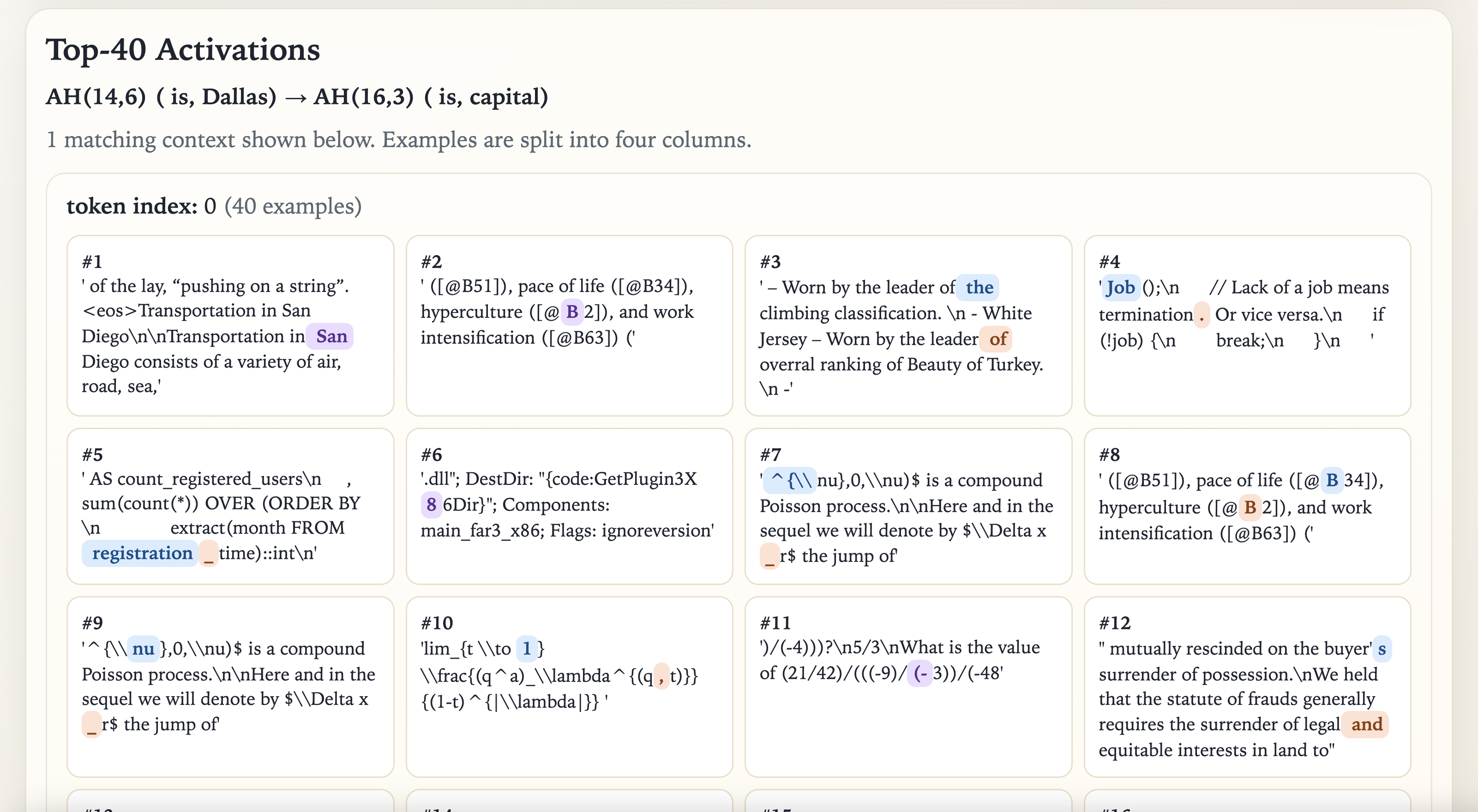}

    \caption{Autointerpretation and 40 highest-scoring contexts for a signal interpreted as predictable completions of formulaic or structured sequences.}
    \label{fig:dallas-example5}
\end{figure}

Figures~\ref{fig:dallas-example1}--\ref{fig:dallas-example5} show example panels for signals involving ``Dallas'', inclusion patterns such as ``X containing Y'', time and location associations, and predictable phrase completions. Each panel shows an edge's automatically generated interpretation and its 40 highest-scoring contexts.

\FloatBarrier

\FloatBarrier

\FloatBarrier

\FloatBarrier

\FloatBarrier

%% file: sections/appendix/clustering.tex
\section{Clustering pipeline} \label{app:clustering}
We now describe our pipeline for per-prompt circuit analysis. The pipeline takes a dataset of prompts as input, traces a circuit for each prompt using ACC++, and produces a clustering of these circuits into families. Finally, we provide a way of summarizing a family with a representative circuit. 

\subsection{Components} \label{app:clustering-components}
Given a circuit $G$, we can define the set of components present in $G$ in three different levels of granularity: nodes, edges, and edge--singular-vector pairs.

\paragraph{Nodes.} We represent the set of components in a circuit $G$ as a binary vector $\mathbf{c} \in \{0, 1\}^M$, where $\mathbf{c}_i = 1$ if the $i$-th node is included in $G$, and $0$ otherwise. The possible nodes consist of Attention Heads and MLP blocks. For example, for the GPT-2 Small architecture, this results in a vector of dimension $M = 156$ ($12 \text{ layers} \times 12 \text{ heads} + 12 \text{ MLPs}$).

\paragraph{Edges.} We represent the set of components in a circuit $G$ as a binary vector $\mathbf{c} \in \{0, 1\}^M$, where $\mathbf{c}_i = 1$ if the $i$-th edge is included in $G$, and $0$ otherwise. For example, for the GPT-2 Small architecture, we have a total of 10,296 possible edges\footnote{$(L \times (H + 1) \times (L - 1) \times H) / 2 = (12 \times 13 \times 11 \times 12) / 2  = 10,296$}. Since the circuits only use a few of the possible edges, we restrict $M$ to the number of unique edges that appear across the dataset. This reduces the size of the binary vector and speeds up computation.

\paragraph{Edge--singular-vector pairs.} We represent the set of components in a circuit $G$ as a binary vector $\mathbf{c} \in \{0, 1\}^M$, where $\mathbf{c}_i = 1$ if the $i$-th edge--singular-vector pair is included in $G$, and $0$ otherwise. For example, for the GPT-2 Small architecture, we have a total of 658,944 possible pairs of edges and singular vectors ($10,296 \times 64$). Similarly to the edges as components, since the circuits only use a few of the possible pairs, we restrict $M$ to the number of unique pairs that appear across the dataset.

\begin{figure}[h!]
    \centering
    \includegraphics[width=\linewidth]{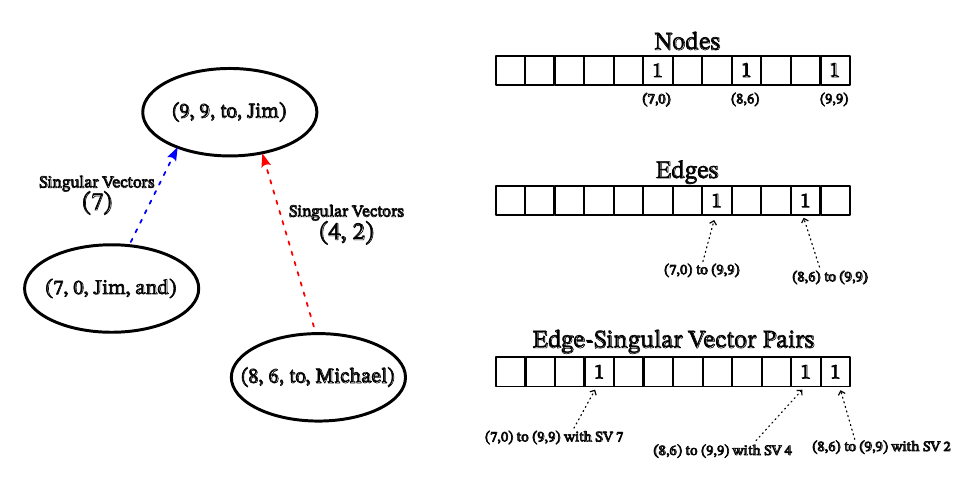}
    \caption{Illustration of circuit components at three levels of granularity: nodes, edges, and edge--singular-vector pairs.}
    \label{fig:app-components-def}
\end{figure}

Figure \ref{fig:app-components-def} provides an illustration of these three component definitions using a simplified subgraph. Notice that we increase the granularity of information when we go from nodes to edges to edge–singular-vector pairs: nodes represents which components of the language model are active, edges record the connectivity between components, and edge–singular-vector pairs not only record the connectivity, but the ``channels'' used for communication inside the model. We notice that the more fine-grained information is crucial for better clustering. 

\subsection{Clustering} \label{app:clustering-clustering}

Having established how to represent a circuit as a set of components, we now outline the pipeline used for clustering the circuits. Given a dataset $P = \{p_1, \dots, p_K\}$ of $K$ circuits, we construct a binary component matrix $C \in \{0, 1\}^{K\times M}$, where the $i$-th row of $C$ corresponds to the binary vector of circuit $p_i$. We compute pairwise distances between all circuits using the Jaccard distance, defined for two binary vectors $\mathbf{a}$ and $\mathbf{b}$ as: 

$$d_J(\mathbf{a}, \mathbf{b}) = 1 - \frac{|\mathbf{a} \cap \mathbf{b}|}{|\mathbf{a} \cup \mathbf{b}|}$$

We then group the circuits via average linkage hierarchical clustering, which iteratively merges the pair of clusters with the smallest average inter-cluster distance.

\subsection{Representative} \label{app:clustering-representative}
Assuming a set of objects $C = \{c_1, \dots, c_K\}$, and some distance function $d : C \times C \to \mathbb{R}^+$. We define the sample representative of $C$ with respect to $d$ as the point $\hat{c} \in C$ such that:

\begin{equation}
    \hat{c}  = \underset{c \in C}{\arg\min} \sum_{i=1}^{K}d(c_i, c)
\end{equation}

For a given cluster, the representative is the object whose binary component vector minimizes the average Jaccard distance to all other members of the cluster. We consider the notion of sample representative as the representative circuit of a circuit family. This concept summarizes each cluster with a single object, which can be used for downstream comparison and interpretation. 

\subsection{Signals} \label{app:clustering-signals}
Given a circuit $G$, we define the set of attention-head nodes that have incoming signals as $\mathcal{A} = \{a_1, \dots, a_M\}$. For an attention-head node $a \in \mathcal{A}$, we define the \textit{destination signal summary} $\mathbf{s}_\text{dst}^{a}$ as the normalized sum of all destination signals targeting $a$. Similarly, we define the \textit{source signal summary} $\mathbf{s}_\text{src}^{a}$ as the normalized sum of all source signals targeting $a$.

For a circuit $G$, we stack the summaries of all nodes $a \in \mathcal{A}$ with incoming signals into two signal summary matrices:

\begin{itemize}
    \item $S_{\text{dst}} \in \mathbb{R}^{M \times D}$, where the $i$-th row corresponds to the destination signal summary $\mathbf{s}_\text{dst}^{a_i}$.
    \item $S_{\text{src}} \in \mathbb{R}^{M\times D}$, where the $i$-th row corresponds to the source signal summary $\mathbf{s}_\text{src}^{a_i}$.
\end{itemize}

Given two circuits $G_1$ and $G_2$, we compute the similarity of their signals using the matrix product:
\[ 
\text{Sim}_{\text{dst}} = S_{\text{dst}}^{G_1} \Big(S_{\text{dst}}^{G_2}\Big)^T \quad \text{and} \quad \text{Sim}_{\text{src}} = S_{\text{src}}^{G_1} \Big(S_{\text{src}}^{G_2}\Big)^T 
\]

The similarity matrices allow us to analyze whether nodes in distinct graphs use similar information.

%% file: sections/appendix/ioi_task.tex
\section{Indirect object identification (IOI)} \label{app:ioi-task}

\subsection{Dataset} \label{app:clustering-dataset}
We construct an IOI dataset balanced across high-level templates and low-level templates. The high-level template represents the order in which the indirect object (A) and subject (B) appear in the initial clause. We distinguish between two configurations: ABBA and BABA.

For example, consider the sentence: ``Then, Michael and Jim went to the office. Michael gave a computer to Jim''. Here, the initial clause introduces the subject ``Michael'' before the indirect object ``Jim'', thus it constitutes a BABA example. Reversing the order of the IO and subject in the initial clause would result in an ABBA example.

Low-level templates are variations on the wording of IOI sentences (e.g., different introductory clauses and connectives). There are 15 low-level templates described in Table \ref{tab:app-low-level-temp}. To construct the final dataset we sampled 100 examples for each unique combination of low-level template and high-level template, totaling 3000 examples ($100 \times 15 \times 2$). Finally, we used the ACC++ tracing to generate the circuits used in the following experiments. 

\begin{table}[h!]
\centering
\caption{Low-level templates considering BABA as high-level template. The ABBA low-level templates are the same, but reversing A and B in the initial clause of the prompt.}
\resizebox{\linewidth}{!}{
\begin{tabular}{ccc}
\hline
ID & Color & Sentence \\ \hline
\addlinespace[0.1cm]
1 & \textcolor[rgb]{0.839, 0.000, 0.000}{\rule{2ex}{2ex}} & Then, [B] and [A] went to the [PLACE]. [B] gave a [OBJECT] to [A] \\
2 & \textcolor[rgb]{0.549, 0.231, 1.000}{\rule{2ex}{2ex}} & Then, [B] and [A] had a lot of fun at the [PLACE]. [B] gave a [OBJECT] to [A] \\
3 & \textcolor[rgb]{0.004, 0.529, 0.000}{\rule{2ex}{2ex}} & Then, [B] and [A] were working at the [PLACE]. [B] decided to give a [OBJECT] to [A] \\
4 & \textcolor[rgb]{0.000, 0.675, 0.776}{\rule{2ex}{2ex}} & Then, [B] and [A] were thinking about going to the [PLACE]. [B] wanted to give a [OBJECT] to [A] \\
5 & \textcolor[rgb]{0.592, 1.000, 0.000}{\rule{2ex}{2ex}} & Then, [B] and [A] had a long argument, and afterwards [B] said to [A] \\
6 & \textcolor[rgb]{1.000, 0.494, 0.820}{\rule{2ex}{2ex}} & After [B] and [A] went to the [PLACE], [B] gave a [OBJECT] to [A] \\
7 & \textcolor[rgb]{0.420, 0.000, 0.310}{\rule{2ex}{2ex}} & When [B] and [A] got a [OBJECT] at the [PLACE], [B] decided to give it to [A] \\
8 & \textcolor[rgb]{1.000, 0.647, 0.184}{\rule{2ex}{2ex}} & When [B] and [A] got a [OBJECT] at the [PLACE], [B] decided to give the [OBJECT] to [A] \\
9 & \textcolor[rgb]{0.341, 0.231, 0.000}{\rule{2ex}{2ex}} & While [B] and [A] were working at the [PLACE], [B] gave a [OBJECT] to [A] \\
10 & \textcolor[rgb]{0.000, 0.337, 0.349}{\rule{2ex}{2ex}} & While [B] and [A] were commuting to the [PLACE], [B] gave a [OBJECT] to [A] \\
11 & \textcolor[rgb]{0.000, 0.000, 0.867}{\rule{2ex}{2ex}} & After the lunch, [B] and [A] went to the [PLACE]. [B] gave a [OBJECT] to [A] \\
12 & \textcolor[rgb]{0.000, 0.992, 0.812}{\rule{2ex}{2ex}} & Afterwards, [B] and [A] went to the [PLACE]. [B] gave a [OBJECT] to [A] \\
13 & \textcolor[rgb]{0.631, 0.459, 0.412}{\rule{2ex}{2ex}} & Then, [B] and [A] had a long argument. Afterwards [B] said to [A] \\
14 & \textcolor[rgb]{0.737, 0.714, 1.000}{\rule{2ex}{2ex}} & The [PLACE] [B] and [A] went to had a [OBJECT]. [B] gave it to [A] \\
15 & \textcolor[rgb]{0.584, 0.710, 0.467}{\rule{2ex}{2ex}} & Friends [B] and [A] found a [OBJECT] at the [PLACE]. [B] gave it to [A] \\ \hline
\end{tabular}
}
\label{tab:app-low-level-temp}
\end{table}

\subsection{Additional clustering results} \label{app:ioi-clustering-results}

\begin{figure}[ht]
\centering
\includegraphics[width=1\linewidth]{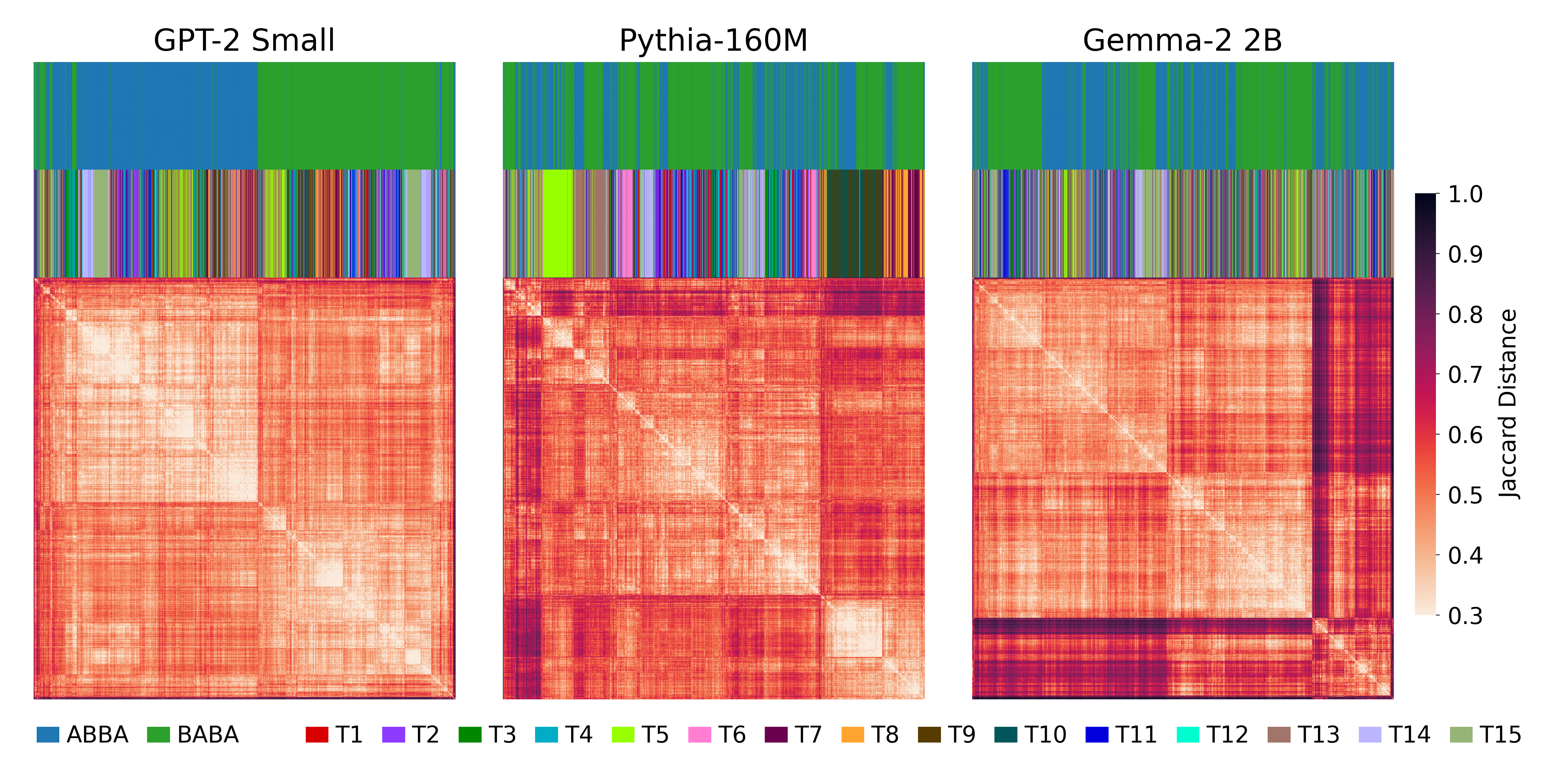}
\caption{Average linkage clustering of prompt-level traces exposes distinct circuit families rather than a single universal IOI circuit. The top annotation bar indicates high-level templates (\textcolor[HTML]{1f77b4}{ABBA} vs. \textcolor[HTML]{2ca02c}{BABA}), while the bottom bar indicates low-level templates (see Appendix~\ref{app:clustering-dataset} for color code). Circuits are represented by their attention heads and MLPs.}
\label{fig:app-clustermap-head-as-component}
\end{figure}

\begin{figure}[ht]
\centering
\includegraphics[width=1\linewidth]{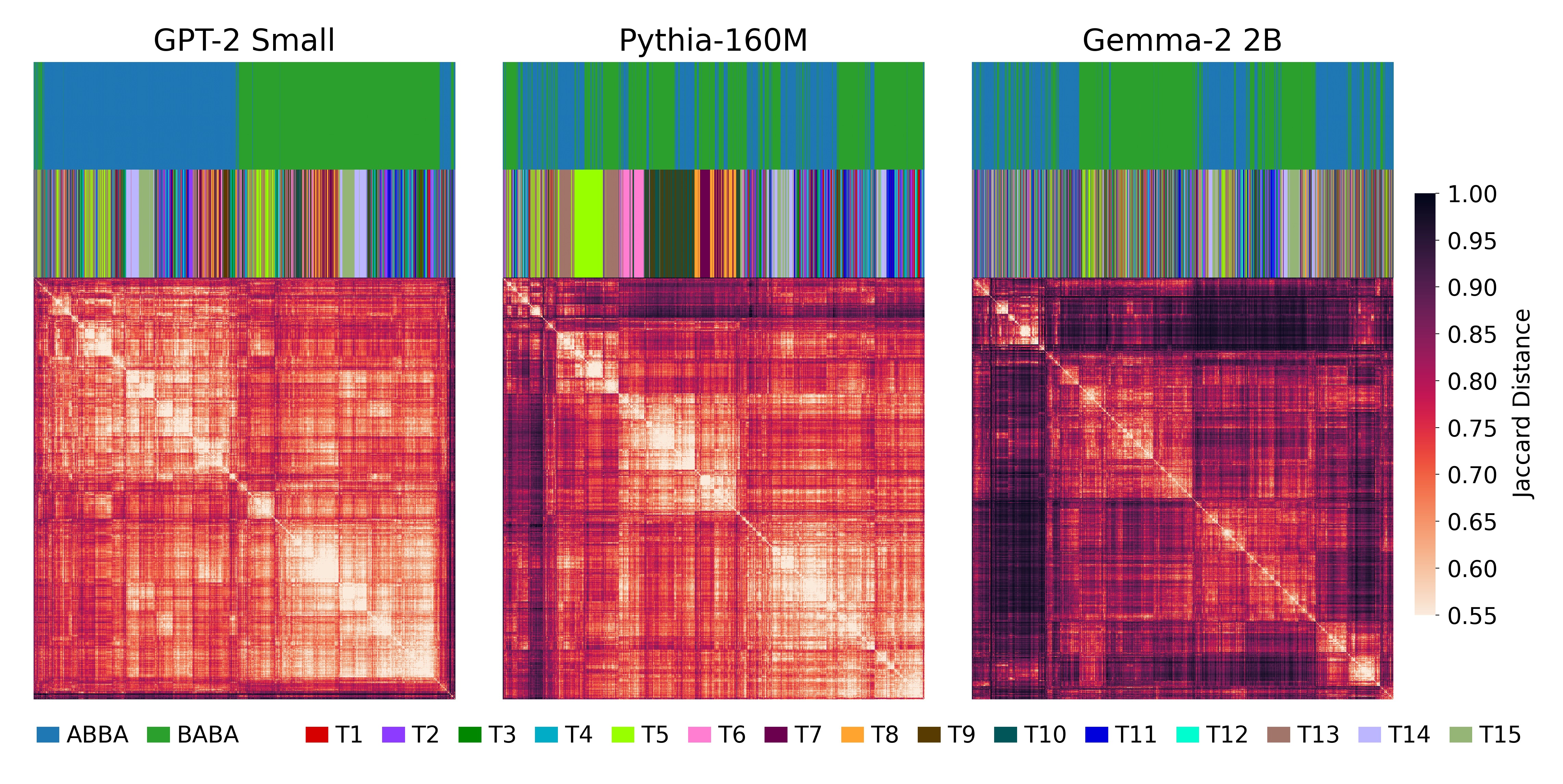}
\caption{Average linkage clustering of prompt-level traces exposes distinct circuit families rather than a single universal IOI circuit. The top annotation bar indicates high-level templates (\textcolor[HTML]{1f77b4}{ABBA} vs. \textcolor[HTML]{2ca02c}{BABA}), while the bottom bar indicates low-level templates (see Appendix~\ref{app:clustering-dataset} for color code). Circuits are represented by their edges.}
\label{fig:app-clustermap-edge-as-component}
\end{figure}

\begin{figure}[ht]
\centering
\includegraphics[width=1\linewidth]{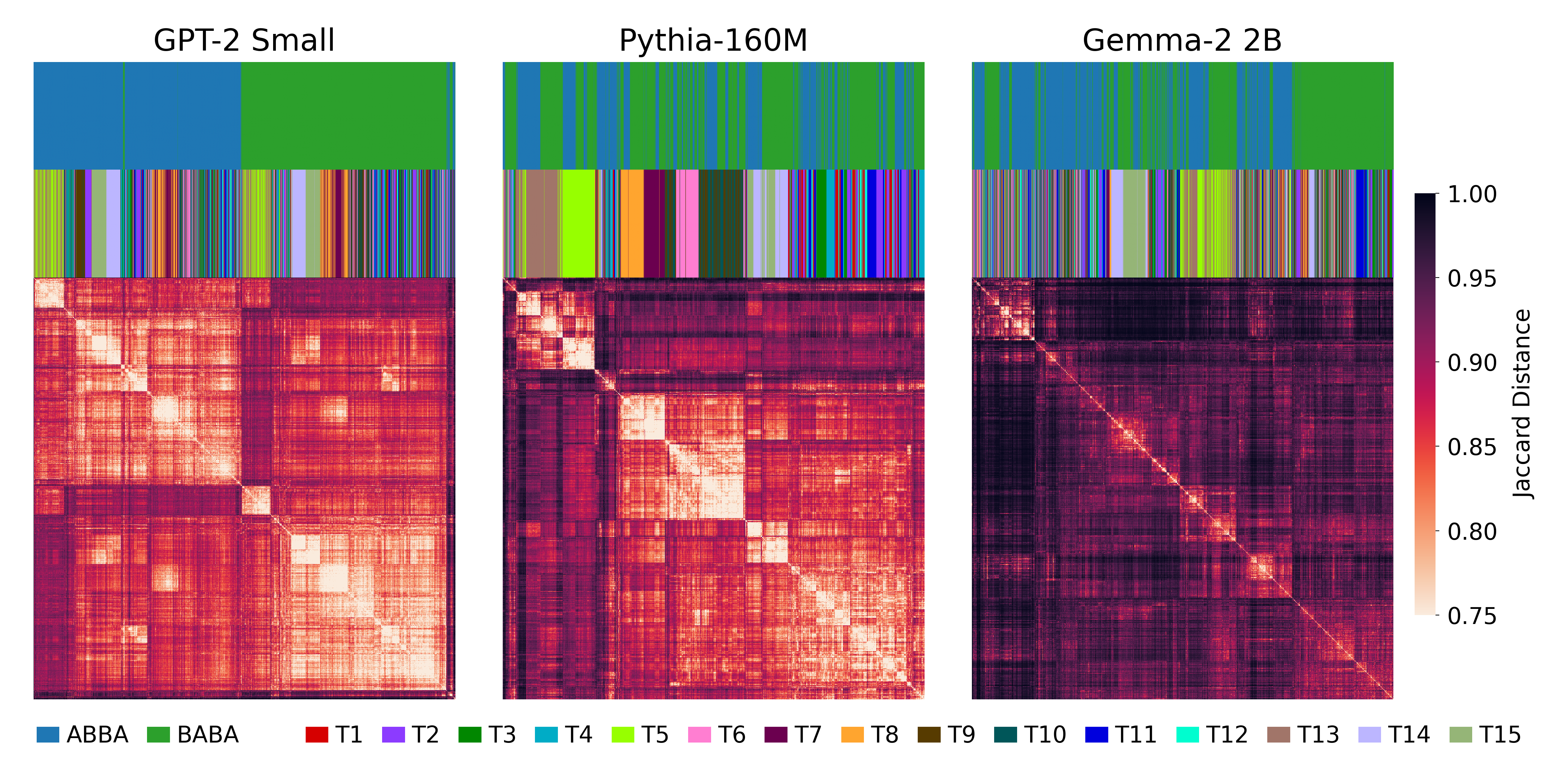}
\caption{Average linkage clustering of prompt-level traces exposes distinct circuit families rather than a single universal IOI circuit. The top annotation bar indicates high-level templates (\textcolor[HTML]{1f77b4}{ABBA} vs. \textcolor[HTML]{2ca02c}{BABA}), while the bottom bar indicates low-level templates (see Appendix~\ref{app:clustering-dataset} for color code). Circuits are represented by pairs of edges and singular vectors.}
\label{fig:app-clustermap-sv-as-component}
\end{figure}

We expand the results from Section~\ref{sec:ioi-word-order} across all models and at three levels of component granularity (Appendix~\ref{app:clustering-components}): nodes (Figure~\ref{fig:app-clustermap-head-as-component}), edges (Figure~\ref{fig:app-clustermap-edge-as-component}), and edge--singular-vector pairs (Figure~\ref{fig:app-clustermap-sv-as-component}). The clustermaps presented here can be read similarly to the ones in the main text: the two top bars annotate the high-level and low-level template of each leaf in the dendrogram, so contiguous colored blocks indicate prompts with the same template clustering together.

 
\paragraph{Clustering granularity.} As we increase the granularity of the component definition: from nodes to edges to edge--singular-vector pairs, we observe a trend where the separability of high-level and/or low-level templates improve. 

\paragraph{Model-dependent clustering.} The relationship between circuit clusters and prompt templates differs across models. For GPT-2 Small, the dominant split is between high-level templates: ABBA and BABA form two clearly separated clusters. In contrast, Pythia-160M clusters primarily by low-level template, with little evidence of a consistent ABBA/BABA partition. 
For Gemma-2 2B, we find neither the clean high-level split of GPT-2 Small nor the dominant low-level grouping of Pythia-160M. While there is a tendency for separation between ABBA and BABA (as indicated by contiguous regions in the top annotation), the boundary is not sharp, and several clusters cut across both high- and low-level labels. The results suggest that Gemma does not rely on a single, consistent template-specific circuit, and is instead more sensitive to finer-grained prompt details beyond the two template types tracked here

\subsection{Additional signal similarity results} \label{app:ioi-signals-result}

\begin{wrapfigure}{R}{0.5\columnwidth}
    \centering
    \includegraphics[width=\linewidth]{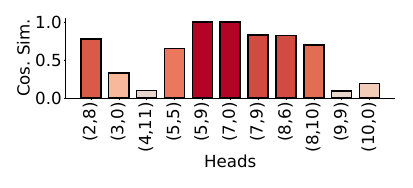}
    \caption{Comparison of source signals similarity for attention heads shared by the GPT-2 Small ABBA and BABA representative IOI circuits.}
    \label{fig:app-signals-comp-models-src}
\end{wrapfigure}

We provide the source-side signal similarity in Figure~\ref{fig:app-signals-comp-models-src}, complementing the destination-side analysis in Figure~\ref{fig:signals-comp-models} of the main text. Head $(9,9)$ shows source-side cosine similarity near zero between the two representatives. This differs from the negative cosine similarity observed on the destination side, but it also shows that signals at $(9,9)$ are different between the ABBA and BABA representatives.

\begin{figure*}[ht]
    \centering
    \includegraphics[width=\linewidth]{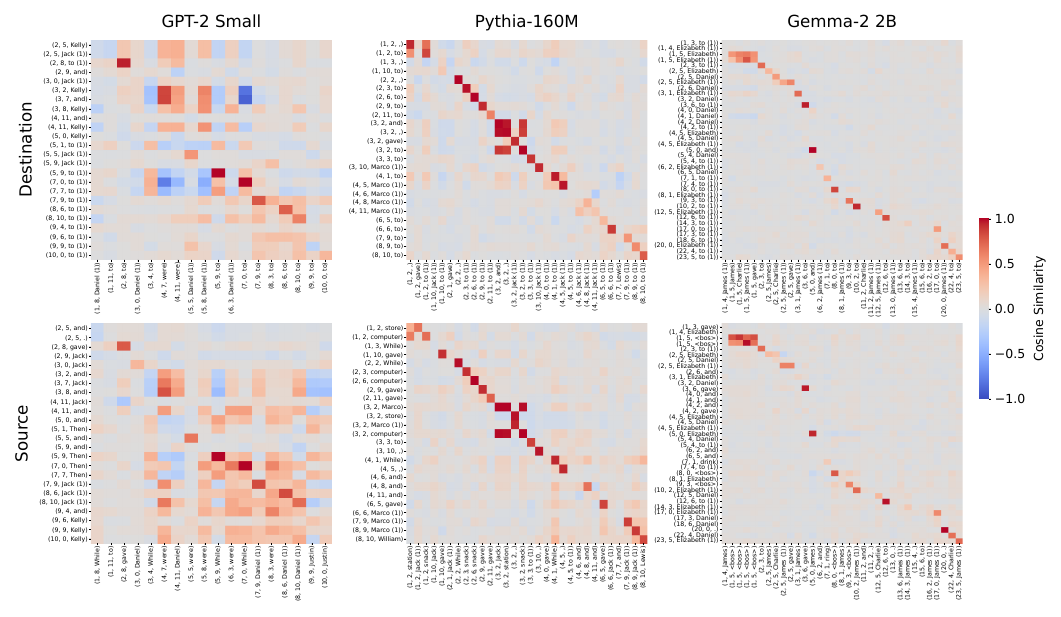}
    \caption{Signal similarity uncovers common and distinct functionality across prompts. Comparison of signal similarity matrices between representative circuits across three different models, organized by column: GPT-2 Small (left), Pythia-160M (middle), and Gemma-2 2B (right). Representatives for GPT-2 Small are from \textcolor[HTML]{1f77b4}{ABBA} (x-axis) and \textcolor[HTML]{2ca02c}{BABA} (y-axis). Representatives for Pythia-160M are from Template 10 (\textcolor[rgb]{0.000, 0.337, 0.349}{\rule{1.5ex}{1.5ex}}, x-axis) and Template 9 (\textcolor[rgb]{0.341, 0.231, 0.000}{\rule{1.5ex}{1.5ex}}, y-axis). Representatives for Gemma-2 2B are from Template 15 (\textcolor[rgb]{0.584, 0.710, 0.467}{\rule{1.5ex}{1.5ex}}, x-axis) and Template 14 (\textcolor[rgb]{0.737, 0.714, 1.000}{\rule{1.5ex}{1.5ex}}, y-axis). Numbers in parentheses (e.g. (1), (2)...) indicate the $n$-th occurrence of the token in the prompt.}
    \label{fig:app-signals-comp-models}
\end{figure*}

Next, we compare all attention-head nodes in one representative with all attention-head nodes in the other (Appendix~\ref{app:clustering-signals}). Figure~\ref{fig:app-signals-comp-models} shows these comparisons for the GPT-2 Small ABBA and BABA representatives, the Pythia-160M Templates 9 and 10 representatives, and the Gemma-2 2B Templates 14 and 15 representatives. The representatives can contain different sets of nodes, so strong matches can appear off-diagonal.

\paragraph{GPT-2 Small.}  For GPT-2 Small (Figure~\ref{fig:app-signals-comp-models}; left-side), we observe three patterns. First, some attention heads are highly stable across templates: for example, head $(5,9)$ shows an almost one-to-one match in both destination and source signals, suggesting an invariant computation that is reused in both ABBA and BABA. Second, some nodes align strongly despite being different heads, which suggests functional redundancy: for instance, the destination signals of $(3,2)$ in the BABA representative closely match those of $(4,7)$ in the ABBA representative, indicating that similar information can be routed by different heads in different templates. Third, some canonical IOI components are present in both representatives but carry different signals: notably, the name-mover head $(9,9)$ shows different destination signals, with negative cosine similarity, between ABBA and BABA, even though its high-level role is the same. This indicates that template changes can preserve \emph{which} component is used while changing \emph{how} that component implements its role, motivating a signal-level analysis of representatives as a way to study mechanistic variation across clusters.

\paragraph{Pythia-160M.} For Pythia-160M (Figure~\ref{fig:app-signals-comp-models}; middle), we compare the representatives for Template~9 (the clause ``\ldots working \emph{at} the [PLACE] \ldots'') and Template~10 (``\ldots commuting \emph{to} the [PLACE] \ldots''); see Appendix~\ref{app:clustering-dataset}, Table~\ref{tab:app-low-level-temp} for the full template definitions. Here, the node labels in the heatmaps (layer/head plus the associated token) let us distinguish two kinds of alignment: (i) the \emph{same} attention head carrying the same role-level signal but with different associated tokens, and (ii) different heads carrying similar signals. Both destination and source heatmaps show strong near one-to-one matches: most high-similarity pairs preserve head identity while swapping the expected tokens, e.g.\ $(3,2,\texttt{Marco})$ aligns with $(3,2,\texttt{Jack})$ and $(2,6,\texttt{computer})$ aligns with $(2,6,\texttt{snack})$ in the source view, while destination-side matches concentrate on the same heads attending to structural tokens such as \texttt{to} and \texttt{and}. This suggests that, despite clustering by low-level template, many Pythia representatives reuse the same head-level routing mechanism, independent of the specific tokens being processed.

\paragraph{Gemma-2 2B.} For Gemma-2 2B, representative signal comparisons are consistent with the weaker template conditioning suggested by Figure~\ref{fig:app-clustermap-sv-as-component}. Even for two low-level templates whose clusters overlap in the clustermap (Template~14 vs.\ Template~15), the representatives share only a \emph{partial} signal backbone. In Figure~\ref{fig:app-signals-comp-models} (right-side), several heads align strongly on structural routing (e.g., head $(5,0)$ matches almost perfectly on the destination token \texttt{and}, and also aligns on the corresponding name token in the source view). However, other heads that appear in both representatives have weak or near-zero alignment to their counterpart, including heads $(1,4)$ and $(8,1)$, indicating that Gemma can change head-level mechanisms even when templates are close.

\subsection{Additional interpretation results} \label{app:ioi-interp-results}

\paragraph{GPT-2: tracing proper-noun features.}

Here we show that the model knows that the IO token is a proper noun, and how circuits use these features in computing model output.   

Figure~\ref{fig:abba-circuit-gpt-proper-noun} shows a portion of the GPT-2 circuit for the ABBA representative.  Proper noun features are shown in dark green. The figure shows that the ``Justin" token is annotated with a ``proper noun" feature by 5 MLPs (MLPs 0, 1, 2, 7, and 8) and by attention head (0, 8).  It also shows that the ``to" token is annotated with the same feature, indicating that it can match with a proper noun.  This matching of proper noun features increases attention from ``to'' to ``Justin'' in heads (10, 0) and (9, 9), helping these heads move information about ``Justin'' to the output position.

Figure \ref{fig:baba-circuit-gpt-proper-noun} shows that the ``proper noun" feature is used in a similar fashion in the BABA representative, but that the ``second item" pathway also makes use of the ``proper noun" feature.  Here, the same MLPs (0, 1, 2, 7, and 8) provide the ``proper noun" annotation as before. MLP 0's annotation is also used by heads producing the ``second item in a parallel pair'' signal.  Thus, the ``proper noun'' feature of ``Kelly'' helps determine where the heads producing the ``second item'' signal attend.

\begin{figure}[ht]
\begin{subfigure}{0.98\textwidth}
\centering
  \includegraphics[width=\linewidth]{figures/interp-signals/abba_circuit_v2.pdf}
  \caption{ABBA}
  \label{fig:abba-circuit-gpt-proper-noun}
\end{subfigure}%
\\
\begin{subfigure}{0.98\textwidth}
  \centering
  \includegraphics[width=\linewidth]{figures/interp-signals/baba_circuit_v2.pdf}
  \caption{BABA}
  \label{fig:baba-circuit-gpt-proper-noun}
\end{subfigure}
\caption{Traces show how GPT-2 uses proper noun features of the IO token. Dark green indicates these features. The prompts are the same as in Figure~\ref{fig:abba-baba-circuits-gpt}.}
\label{fig:abba-baba-circuits-gpt-proper-noun}
\end{figure}

\begin{figure}[ht]
\begin{subfigure}{\textwidth}
\centering
  \includegraphics[width=0.6\textwidth]{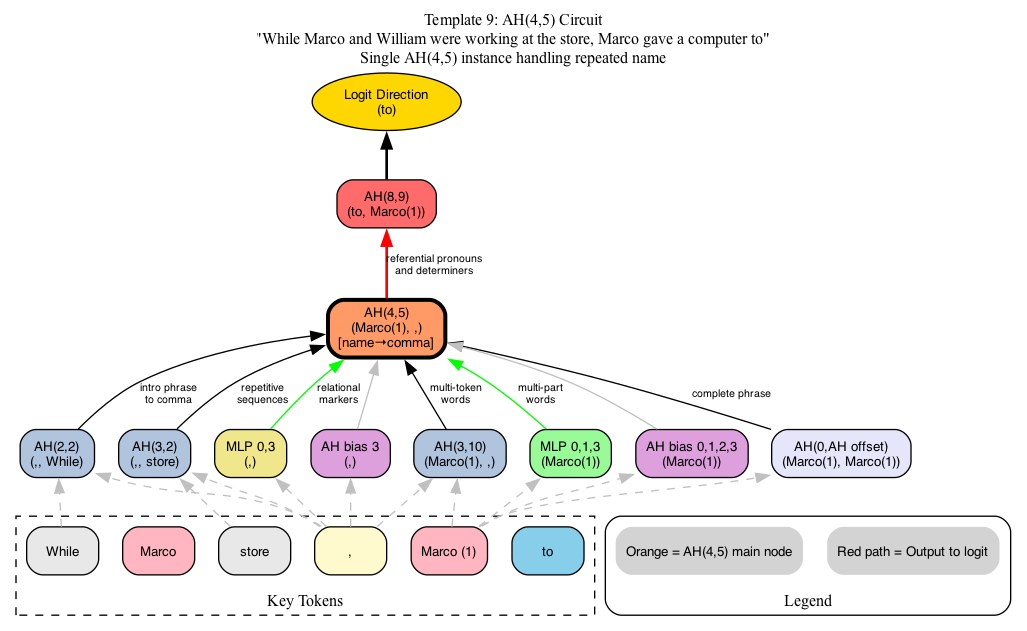}
  \caption{Template 9}
  \label{fig:pythia-t9-circuit}
\end{subfigure}%
\\
\begin{subfigure}{\textwidth}
  \centering
  \includegraphics[width=\textwidth]{figures/interp-signals/t10_45_circuit.pdf}
  \caption{Template 10}
  \label{fig:pythia-t10-circuit}
\end{subfigure}
\caption{Both Pythia circuits include head $(4,5)$ acting on the repeated name. In Template~10, head $(4,5)$ also acts on the second \texttt{to} token, with a path through head $(6,5)$ to later heads and the output.}
\label{fig:pythia-template-circuits}
\end{figure}

\paragraph{Pythia: comparing Templates 9 and 10.}
For Pythia, we compare circuits from representatives of Templates 9 and 10 (see Figure~\ref{fig:app-signals-comp-models}, middle). The Template~9 representative is ``While Marco and William were working at the store, Marco gave a computer to'', while the Template~10 representative is ``While Jack and Lewis were commuting to the station, Jack gave a snack to''. The circuits are highly similar, but show some differences. One difference is an additional firing of head (4,5) in Template~10, attending to the token pair (to(1), to(1)). To understand the reason for this difference, we show the two circuits and their signal interpretations in Figure~\ref{fig:pythia-template-circuits}. Comparing these two circuits, we see that the difference may arise because Template~10 contains two instances of the word ``to'', which may require additional disambiguation by the model. We see that head (4,5) receives a signal annotating the second ``to'' as the predecessor of a noun phrase. This suggests that the head helps identify the second ``to'' as introducing the indirect object and supports moving information about the IO to that position.

%% file: sections/appendix/mioi_task.tex
\section{Multilingual indirect object identification (MIOI)} \label{app:mioi-task}

\subsection{Dataset}

We construct a multilingual version of the IOI dataset (Appendix \ref{app:ioi-task}) spanning English, Spanish, French, and Portuguese. Each language uses the same set of template structures, with names adapted to be natural in each language. This produces a balanced dataset across both template types and language. In total there are 800 prompts across the four languages, four low-level templates and the two high-level templates (ABBA/BABA). The low-level templates are described in Table \ref{app:tab-mioi}. 

\begin{table}[h]
\caption{Language-specific low-level templates considering BABA as high-level template. The ABBA low-level templates are the same, but reversing A and B in the initial clause of the prompt.}
\begin{tabular}{ccc}
\hline
ID                 & Language & Sentence                                                                                     \\ \hline
\multirow{4}{*}{0} & English       & When [B] and [A] went to the restaurant, [B] gave a drink to [A].                 \\
                   & Portuguese       & Quando [B] e [A] foram ao restaurante, [B] deu uma bebida para [A].               \\
                   & French       & Quand [B] et [A] sont allés au restaurant, [B] a donné une boisson à [A].         \\
                   & Spanish       & Cuando [B] y [A] fueron al restaurante, [B] le dio una bebida a [A].              \\ \hline
\multirow{4}{*}{1} & English       & After [B] and [A] arrived at the library, [B] handed a book to [A].               \\
                   & Portuguese       & Depois que [B] e [A] chegaram à biblioteca, [B] entregou um livro para [A].       \\
                   & French       & Après que [B] et [A] sont arrivés à la bibliothèque, [B] a remis un livre à [A].  \\
                   & Spanish       & Después de que [B] y [A] llegaron a la biblioteca, [B] le entregó un libro a [A]. \\ \hline
\multirow{4}{*}{2} & English       & While [B] and [A] were at the school, [B] handed a pen to [A].                    \\
                   & Portuguese       & Enquanto [B] e [A] estavam na escola, [B] entregou uma caneta para [A].           \\
                   & French       & Pendant que [B] et [A] étaient à l'école, [B] a remis un stylo à [A].             \\
                   & Spanish       & Mientras [B] y [A] estaban en la escuela, [B] le entregó un bolígrafo a [A].      \\ \hline
\multirow{4}{*}{3} & English       & When [B] and [A] met at the station, [B] handed a ticket to [A].                  \\
                   & Portuguese       & Quando [B] e [A] se encontraram na estação, [B] entregou um bilhete para [A].     \\
                   & French       & Quand [B] et [A] se sont retrouvés à la gare, [B] a remis un billet à [A].        \\
                   & Spanish       & Cuando [B] y [A] se encontraron en la estación, [B] le entregó un boleto a [A].   \\ \hline
\end{tabular}
\label{app:tab-mioi}
\end{table}

\subsection{Additional results}\label{app:mioi-results}

We apply the same tracing and clustering pipeline (Appendix~\ref{app:clustering}) to mGPT and Gemma-2 2B: we trace each prompt with ACC++, represent each circuit as a set of discrete components, compute pairwise Jaccard distances between prompt-level traces, and cluster them via average linkage hierarchical clustering. Here, we report the results for Gemma-2 2B (similar to \S~\ref{sec:multilingual-ioi} for mGPT). We used $\tau=\frac{1.5}{d}$ in these experiments (Appendix~\ref{app:finding-tau}).

\begin{figure}[t]
 \begin{minipage}[t]{0.45\linewidth}
    \centering
    \includegraphics[width=\linewidth]{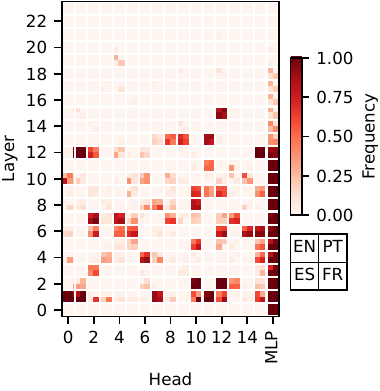}
    \caption{Frequency of components across traces for mGPT shows that multilingual models use similar set of components across languages.}
    \label{fig:multilingual-ioi-heatmap}
  \end{minipage}
  \begin{minipage}[t]{0.45\linewidth}
    \centering
    \includegraphics[width=\linewidth]{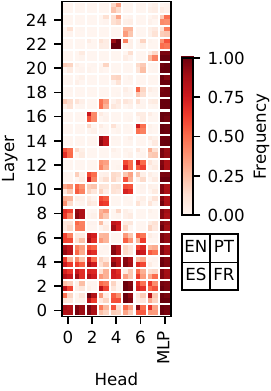}
    \caption{Frequency of components across traces for Gemma-2 2B shows that multilingual models use similar set of components across languages.}
    \label{fig:app-mioi-gemma-heatmap}
  \end{minipage}\rule{1em}{0pt}%
\end{figure}

\paragraph{Component reuse.} Figure~\ref{fig:multilingual-ioi-heatmap} and Figure~\ref{fig:app-mioi-gemma-heatmap} presents the frequency of components for mGPT and Gemma-2 2B, respectively. We notice that the majority of the same heads and MLPs activate for all languages.

\begin{figure}[t]
    \centering
    \includegraphics[scale=0.95]{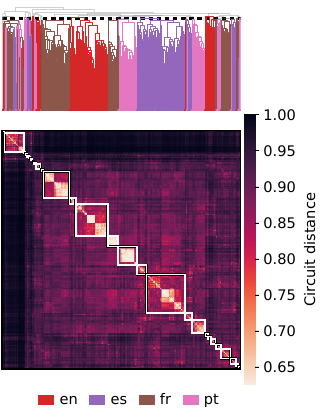}
    \caption{Average linkage clustering of prompt-level traces exposes grouping by language in Gemma-2 2B. Circuits are represented as sets of edge--singular-vector pairs.}
    \label{fig:app-mioi-gemma-clustermap}
\end{figure}

\paragraph{Language-specific signals.} Figure~\ref{fig:app-mioi-gemma-clustermap} shows that traces cluster by language in Gemma-2 2B. This result suggests the presence of language-specific signals being routed through the same components. Differently from mGPT, there is no high-level template clustering. This also mirrors the result we found in English-only IOI (\S~\ref{sec:ioi-word-order}, Appendix~\ref{app:ioi-task}), where Gemma-2 2B did not cluster by high-level or low-level templates.
  
\paragraph{Cross-language typological distance via URIEL+.}
We compare the cross-language structure of our circuits to typological distances between languages, drawn from URIEL+~\cite{khan2025uriel+}. URIEL+ represents each language as a vector of binary and numeric features (e.g.\ word order, agreement, phoneme inventory) sourced from linguistic resources such as WALS and Glottolog, organized into six feature families: \textit{syntactic}, \textit{phonological}, \textit{inventory}, \textit{genetic}, \textit{geographic}, and \textit{featural} (a concatenation of the first three). The pairwise distance between two languages within a family is the cosine distance between their feature vectors in that family; we fill missing entries with the URIEL+ KNN imputer ($k = 10$) before computing distances. On the circuit side, we restrict to prompt pairs aligned by high-level template, low-level template, and position within bucket, giving the closest one-to-one matches available in the dataset (paired sample size 142--193 per language pair), and summarize each language pair by the mean Jaccard distance over its paired prompts. For each URIEL+ family we then correlate the six cross-language pair distances against the six URIEL+ distances using Pearson $r$, and report two-sided $p$-values from an exhaustive permutation test.

\begin{figure}
    \centering
    \includegraphics[width=0.5\linewidth]{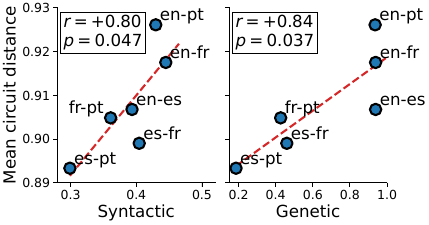}
    \caption{Mean circuit distances between Gemma-2 2B language pairs correlate with URIEL+ syntactic (left) and genetic (right) distances.}
    \label{fig:app-mioi-gemma-uriel}
\end{figure}

\paragraph{Circuit distance and typological distance (Gemma-2 2B).} Figure~\ref{fig:app-mioi-gemma-uriel} shows that the mean circuit distances strongly correlate to URIEL+ syntactic and genetic typological distances. No significant correlation was observed for the remaining feature families (\textit{phonological}, \textit{inventory}, \textit{geographic}, and \textit{featural}).

%% file: sections/appendix/icl_and_translation.tex
\section{In-context learning and translation examples}\label{app:icl-translation}

\paragraph{Gemma: in-context learning and factual recall.} We turn to an example illustrating the use of ACC++ to elucidate a circuit that performs in-context learning (ICL) as well as factual recall.  The prompt is ``What sport Jordan played? A: Basketball.  What sport Tom Brady played? A:" and the model correctly answers ``football."  Figure~\ref{fig:gemma-facts-1} shows the interpretable circuit for this prompt.  Nodes are arranged vertically according to layer in the model.  

The figure shows broadly how ICL is used in this setting by the model.  The path in blue analyzes the one-shot phrase at the start of the prompt, moving information from ``Jordan" into ``basketball" in layer 1, and ``sport" into ``basketball" in layer 10.  The path in red starts by recognizing the feature ``Tom" that is present in the input embedding of ``Brady," and head (3, 2) attends to ``Brady" due to the feature ``mentions of the primary entity or subject."  The paths unite in layers 13-20, with a key step being head (16,5) attending from the end token (``:'') to ``Brady'', guided by a signal interpreted as ``completion of words or phrases.'' Attention heads in layers 16, 17, 20, and 24 then move information about ``football'' to the end token, and this is reinforced by MLPs 21 and 25.

\begin{figure*}[p]
  \centering
  \includegraphics[width=\textwidth]{figures/interp-signals/gemma_facts_1_reordered_fullpage.pdf}
  \caption{``What sport Jordan played? A: Basketball.  What sport Tom Brady played? A:" (football).}
  \label{fig:gemma-facts-1}
\end{figure*}

\paragraph{Gemma: translation.}
We also apply ACC++ to a bilingual in-context translation prompt, \texttt{"Mary gave me a book and a card.\ Marie m'a donn\'e un livre et une"}, for which Gemma-2 2B's top-2 next-token predictions are \texttt{carte} ($P = 0.9895$) and \texttt{lettre} ($P = 0.0041$). Removing the English demonstration (i.e.\ tracing the bare French prefix \texttt{"Marie m'a donn\'e un livre et une"}) reduces the probability assigned to \texttt{carte} to ${\approx}0.077$, a factor of ${\sim}13\times$.

\begin{figure}[h]
\centering
\includegraphics[width=0.95\linewidth]{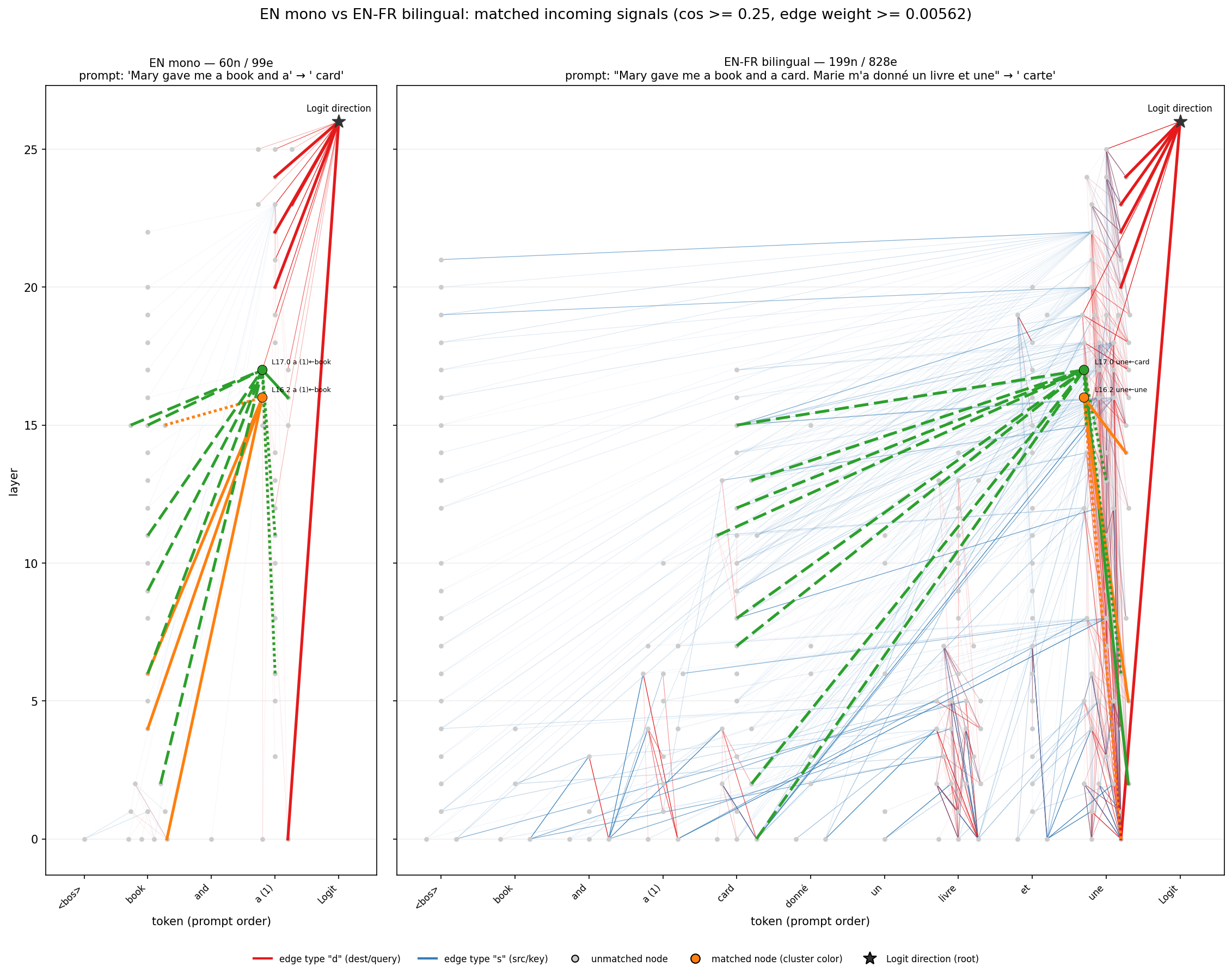}
\caption{Side-by-side circuits for the English-only prompt \texttt{"Mary gave me a book and a"} (left, predicting \texttt{card}) and the bilingual prompt \texttt{"Mary gave me a book and a card.\ Marie m'a donn\'e un livre et une"} (right, predicting \texttt{carte}). A node in one circuit is matched to its counterpart in the other when they share the same attention head (or MLP) and the cosine similarity between their aggregated incoming-signal vectors is at least $0.25$; matched nodes and their incoming edges are highlighted.}
\label{fig:bilingual-icl-sidebyside}
\end{figure}

Table~\ref{tab:bilingual-icl-decomp} reports the full set of root-incoming edges produced by ACC++ for both answer tokens; weights are signed logit-unit contributions. The two cross-lingual heads AH(18, 6) and AH(19, 1), which attend from the final French determiner \texttt{une} to the English demonstration noun \texttt{card}, appear only in the \texttt{carte} trace (i.e., tracing from the \texttt{carte} logit direction); the \texttt{lettre} trace contains only MLP 24 and AH(25, 7). Figure~\ref{fig:bilingual-icl-sidebyside} compares the bilingual \texttt{carte} trace (right) to the trace of the English-only prompt \texttt{"Mary gave me a book and a"} (left, predicting \texttt{card}). Two nodes, one in each circuit, are taken to be matched when they share the same attention head (or MLP) and the cosine similarity between their aggregated incoming-signal vectors is at least $0.25$; in that case both the nodes and their incoming edges are highlighted in the figure. A large fraction of the bilingual trace's incoming signals match signals already present in the monolingual circuit, with AH(18, 6) and AH(19, 1) appearing as bilingual-only cross-lingual additions from \texttt{une} to \texttt{card}.

\begin{table}[h]
\centering
\caption{Logit-edge trace of the bilingual prompt for the top-1 prediction \texttt{carte} and the runner-up \texttt{lettre}. Weights are signed contributions to the answer token's logit at the final position.}
\label{tab:bilingual-icl-decomp}
\small
\begin{tabular}{llrrrr}
\toprule
& & \multicolumn{2}{c}{\texttt{carte}} & \multicolumn{2}{c}{\texttt{lettre}} \\
\cmidrule(lr){3-4} \cmidrule(lr){5-6}
Source & Tokens & Weight & \% & Weight & \% \\
\midrule
Embedding                   & une\,$\leftarrow$\,une  & $1.84$ & $8.4$  & --- & --- \\
AH(18, 6)  (cross-lingual)  & une\,$\leftarrow$\,card & $0.76$ & $3.5$  & --- & --- \\
AH(19, 1)  (cross-lingual)  & une\,$\leftarrow$\,card & $3.33$ & $15.2$ & --- & --- \\
MLP 20                      & une\,$\leftarrow$\,une  & $2.46$ & $11.2$ & --- & --- \\
MLP 22                      & une\,$\leftarrow$\,une  & $3.70$ & $16.9$ & --- & --- \\
MLP 23                      & une\,$\leftarrow$\,une  & $2.36$ & $10.7$ & --- & --- \\
MLP 24                      & une\,$\leftarrow$\,une  & $6.23$ & $28.4$ & $8.36$ & $48.2$ \\
AH(25, 7)                   & une\,$\leftarrow$\,une  & $1.26$ & $5.8$  & $8.99$ & $51.8$ \\
\midrule
Total                       &                          & $21.93$ &       & $17.35$ & \\
\bottomrule
\end{tabular}
\end{table}

%% file: checklist.tex
\section*{NeurIPS Paper Checklist}

\begin{enumerate}

\item {\bf Claims}
    \item[] Question: Do the main claims made in the abstract and introduction accurately reflect the paper's contributions and scope?
    \item[] Answer: \answerYes{}
    \item[] Justification: We are clear in the introduction what are our contributions, pointing out to where they are described in more detail (Sections~\ref{sec:improv-acc} and \ref{sec:uses})

\item {\bf Limitations}
    \item[] Question: Does the paper discuss the limitations of the work performed by the authors?
    \item[] Answer: \answerYes{}
    \item[] Justification: We have a limitations paragraph in Section~\ref{sec:conclusions}.

\item {\bf Theory assumptions and proofs}
    \item[] Question: For each theoretical result, does the paper provide the full set of assumptions and a complete (and correct) proof?
    \item[] Answer: \answerYes{}
    \item[] Justification: Appendix~\ref{app:unified-bilinear} states the conditions under which the unified bilinear form holds (with/without bias, with/without RoPE) and provides proofs that the transformed tokens reproduce the original attention score.

    \item {\bf Experimental result reproducibility}
    \item[] Question: Does the paper fully disclose all the information needed to reproduce the main experimental results of the paper to the extent that it affects the main claims and/or conclusions of the paper (regardless of whether the code and data are provided or not)?
    \item[] Answer:  \answerYes{}
    \item[] Justification: We provide extensive details in the Appendices~\ref{app:unified-bilinear}, \ref{app:new-acc}, \ref{app:finding-tau}, \ref{app:reproducing-previous-results}, \ref{app:interp-signals}, \ref{app:clustering}, \ref{app:ioi-task}, \ref{app:mioi-task}, and \ref{app:icl-translation} to show how to reproduce ACC++ and all the experiments.

\item {\bf Open access to data and code}
    \item[] Question: Does the paper provide open access to the data and code, with sufficient instructions to faithfully reproduce the main experimental results, as described in supplemental material?
    \item[] Answer: \answerYes{}
    \item[] Justification: We provide the code in the supplemental material.

\item {\bf Experimental setting/details}
    \item[] Question: Does the paper specify all the training and test details (e.g., data splits, hyperparameters, how they were chosen, type of optimizer) necessary to understand the results?
    \item[] Answer: \answerYes{}
    \item[] Justification: Appendix~\ref{app:finding-tau} describes how the hyperparameter $\tau$ of ACC++ is chosen. Appendices~\ref{app:reproducing-previous-results}, \ref{app:clustering}, \ref{app:ioi-task}, and \ref{app:mioi-task} specify the models (GPT-2 Small, Pythia-160M, Gemma-2 2B, mGPT), datasets (IOI, GT, GP, balanced IOI, multilingual IOI), and dataset sizes used for each experiment. Appendix~\ref{app:interp-signals} gives the full autointerpretation details (explainer/judge models, prompts, and scoring protocol).

\item {\bf Experiment statistical significance}
    \item[] Question: Does the paper report error bars suitably and correctly defined or other appropriate information about the statistical significance of the experiments?
    \item[] Answer: \answerYes{}
    \item[] Justification: When applicable, we report statistical significance of the experiments (see Figure~\ref{fig:multilingual-ioi-uriel} and Figure~\ref{fig:app-mioi-gemma-uriel}).

\item {\bf Experiments compute resources}
    \item[] Question: For each experiment, does the paper provide sufficient information on the computer resources (type of compute workers, memory, time of execution) needed to reproduce the experiments?
    \item[] Answer: \answerYes{}
    \item[] Justification: Appendix~\ref{app:compute} describes the hardware (local M4 Mac for ACC vs.\ ACC++ comparisons; 28-core CPU nodes for GPT-2/Pythia tracing; A100 40~GB GPU nodes for Gemma-2 2B tracing; 4$\times$L40S GPU nodes for the autointerpretation pipeline) and per-prompt / end-to-end timing for each set of experiments.
    
\item {\bf Code of ethics}
    \item[] Question: Does the research conducted in the paper conform, in every respect, with the NeurIPS Code of Ethics \url{https://neurips.cc/public/EthicsGuidelines}?
    \item[] Answer: \answerYes{} 
    \item[] Justification: Yes, the research conducted in the paper conforms, in every respect, with the NeurIPS Code of Ethics.

\item {\bf Broader impacts}
    \item[] Question: Does the paper discuss both potential positive societal impacts and negative societal impacts of the work performed?
    \item[] Answer: \answerYes{}
    \item[] Justification: Appendix~\ref{app:broader-impacts} discusses potential benefits for model auditing and debugging, and the risk of using interpretability findings to modify circuits involved in model safeguards.
    
\item {\bf Safeguards}
    \item[] Question: Does the paper describe safeguards that have been put in place for responsible release of data or models that have a high risk for misuse (e.g., pre-trained language models, image generators, or scraped datasets)?
    \item[] Answer: \answerNA{}
    \item[] Justification: This paper is about interpretability of language models and do not pose such risks.

\item {\bf Licenses for existing assets}
    \item[] Question: Are the creators or original owners of assets (e.g., code, data, models), used in the paper, properly credited and are the license and terms of use explicitly mentioned and properly respected?
    \item[] Answer: \answerYes{}
    \item[] Justification: Section~\ref{sec:improv-acc} and Appendices~\ref{app:reproducing-previous-results}, \ref{app:interp-signals}, and \ref{app:mioi-task} credit the original sources of all models (GPT-2, Pythia, Gemma-2 2B, mGPT) and datasets (IOI, GP, GT, The Pile via pile-10k, URIEL+) used in this work. Models and datasets are used under their respective licenses, and we follow the terms of use stated by their authors.

\item {\bf New assets}
    \item[] Question: Are new assets introduced in the paper well documented and is the documentation provided alongside the assets?
    \item[] Answer: \answerYes{}
    \item[] Justification: Our code is well documented (see README.md in the .zip file on the supplemental material).

\item {\bf Crowdsourcing and research with human subjects}
    \item[] Question: For crowdsourcing experiments and research with human subjects, does the paper include the full text of instructions given to participants and screenshots, if applicable, as well as details about compensation (if any)? 
    \item[] Answer: \answerNA{} 
    \item[] Justification: This paper does not involve crowdsourcing nor research with human subjects.

\item {\bf Institutional review board (IRB) approvals or equivalent for research with human subjects}
    \item[] Question: Does the paper describe potential risks incurred by study participants, whether such risks were disclosed to the subjects, and whether Institutional Review Board (IRB) approvals (or an equivalent approval/review based on the requirements of your country or institution) were obtained?
    \item[] Answer: \answerNA{}.
    \item[] Justification: This paper does not involve crowdsourcing nor research with human subjects.

\item {\bf Declaration of LLM usage}
    \item[] Question: Does the paper describe the usage of LLMs if it is an important, original, or non-standard component of the core methods in this research? Note that if the LLM is used only for writing, editing, or formatting purposes and does \emph{not} impact the core methodology, scientific rigor, or originality of the research, declaration is not required.
    \item[] Answer: \answerYes{}
    \item[] Justification: We explicitly mention how we use LLMs on the autointerpretation pipeline (see Section~\ref{sec:improv-acc}) and on the qualitative tracing figures (see Appendix~\ref{app:trace-examples}).

\end{enumerate}